\documentclass[10pt]{report}
\usepackage[utf8]{inputenc}
\usepackage[T1]{fontenc}
\usepackage{microtype}
\usepackage{amsmath}
\usepackage{graphicx}
\usepackage{geometry}
\usepackage{floatrow}
\usepackage{layout}
\usepackage{mathtools}
\usepackage{amssymb} 
\usepackage{multirow}
\usepackage{multicol}
\usepackage{caption}
\usepackage{subcaption}
\usepackage{authblk} 
\usepackage[square,numbers]{natbib} 
\usepackage{indentfirst}
\usepackage{tikz} 
\usetikzlibrary{trees} 
\usepackage{algorithm} 
\usepackage[noend]{algpseudocode}
\usepackage{tocloft} 
\usepackage{parskip}
\usepackage{titlesec} 
\titlespacing*{\section}{0pt}{5ex plus 1ex minus .2ex}{3ex plus .2ex}
\titlespacing*{\subsection}{0pt}{2.5ex plus 1ex minus .2ex}{1.5ex plus .2ex}
\titlespacing*{\subsubsection}{0pt}{1.5ex plus 1ex minus .2ex}{1ex plus .2ex}
\usepackage{wrapfig}
\usepackage{url}
\usepackage{amsmath, dsfont}
\usepackage{lscape}
\usepackage[modulo]{lineno} 
\usepackage{tabularx} 
\usepackage{hyperref} 
\hypersetup{
    colorlinks,
    linktoc=all, 
    citecolor=blue,
    filecolor=black,
    linkcolor=black,
    urlcolor=blue
}
\usepackage{booktabs} 
\usepackage{titlesec} 

 \fontencoding{T1}
  \fontfamily{garamond}
  \fontseries{m}
  \fontshape{it}
  \fontsize{12}{15}
  \selectfont

\usepackage{amsthm} 
\theoremstyle{definition}
\newtheorem{definition}{Definition}[section]
 
\usepackage{fancyhdr} 
\begin{document}

\title{Generating tabular datasets under differential privacy}
\author{Gianluca Truda}
\affil{Vrije Universiteit Amsterdam}
\date{\today}

\maketitle

\begin{abstract} 
Machine Learning (ML) is accelerating progress across fields and industries, but relies on accessible and high-quality training data. Some of the most important datasets are found in biomedical and financial domains in the form of spreadsheets and relational databases. But this tabular data is often sensitive in nature and made inaccessible by privacy regulations like GDPR, creating a major bottleneck to scientific research and real-world applications of Artificial Intelligence (AI).

The emerging field of synthetic data generation offers the potential to unlock this sensitive data, by training ML models to represent the statistical structure of the dataset and synthesise new samples, without compromising individual privacy. Unfortunately, generative models tend to memorise and regurgitate training data, which undermines the privacy goal. To remedy this, researchers have incorporated the mathematical framework of Differential Privacy (DP) into the training process of neural networks, limiting memorisation and enforcing provable privacy guarantees. But this creates a trade-off between the quality and privacy of the resulting data. 

Generative Adversarial Networks (GANs) are the current leading paradigm for synthesising tabular data under Differential Privacy, but suffer from unstable adversarial training and mode collapse. These effects are compounded by the added privacy constraints. The tabular data modality exacerbates this further, as mixed data types, non-Gaussian distributions, high-cardinality, and sparsity are all challenging for deep neural networks to represent and synthesise. Yet, it is precisely tabular data that is most likely to be sensitive and valuable to solving real-world problems.

In this work, we optimise the quality-privacy trade-off of generative models, with the aim of producing higher quality tabular datasets that can be safely released under provable privacy guarantees. We first implement novel end-to-end models that leverage attention mechanisms to learn reversible representations of tabular data. Whilst effective at learning dense embeddings, the increased model size rapidly depletes the privacy budget.

We also introduce TableDiffusion, the first differentially-private diffusion model for tabular data synthesis. Our approach leverages the stability of the emerging diffusion paradigm to achieve much higher data and privacy efficiencies than previous methods. In a comprehensive benchmarking experiment, with a suite of complementary evaluation metrics, we show that our diffusion model consistently  outperforms state-of-the-art GANs on real-world tabular datasets across privacy levels. TableDiffusion produces higher-fidelity synthetic datasets that capture more of the original sample diversity, whilst maintaining the same strict privacy guarantees. Our novel diffusion model also avoids the mode collapse problem of GANs and achieves state-of-the-art performance on privatised tabular data synthesis.
    
By implementing two variants of TableDiffusion, we could perform ablation to understand precisely which mechanisms resulted in improved performance. The diffusion paradigm appears to be vastly more data-efficient and privacy-efficient than the adversarial paradigm, due to the augmented re-use of each data batch and the smoother iterative training process. By implementing TableDiffusion to predict and subtract the added noise instead of directly denoising the data, we enabled it to bypass the challenges of reconstructing mixed-type tabular data and produce higher-quality synthetic datasets.

Overall, TableDiffusion is a powerful new tool for generating high-quality synthetic tabular data, even when adhering to strict privacy constraints. It is our hope that these contributions unlock new avenues for research and enable real-world progress in applying AI techniques on sensitive tabular data. 

\end{abstract}

\tableofcontents



\chapter{Introduction}

Machine learning (ML) has the ability to accelerate progress in every domain, with huge untapped potential in biomedical fields \cite{kadra2021well}. With recent advances in ML techniques and computing hardware, our capacity to extract valuable insight from raw data is only growing. But ML techniques rely on the availability of high-quality data, and this is usually the bottleneck in real-world applications, particularly in domains where data is sensitive, such as healthcare and consumer finance \cite{jordon2022synthetic}. 

Accessing this sensitive data is inefficient and cumbersome because of the ethical and regulatory restrictions that have been imposed to protect individual privacy. So although there are ample quantities of high-impact data available \cite{borisov2022deeptabular}, they are siloed away in databases that both internal and external researchers struggle to access \cite{jordon2020syntheticdatareview}. These regulatory hurdles are a major deterrent to researchers and companies involving themselves with analysing the sensitive data. These restrictions on sensitive data help protect individual privacy, but create major impediments to scientific progress \cite{jordon2022synthetic}. 

Unfortunately, naïvely anonymising data by stripping personally-identifiable information is woefully insufficient, as the sparsity of high-dimensionality data and the sophistication of ML make it trivial to re-identify individuals. For example, researchers were able to retrieve exact records for almost half the entire US population from public US Census statistics \cite{jordon2022synthetic} and re-identify thousands of anonymised Netflix viewers along with their watch history and ratings \cite{narayanan2008robust}. Ideally, we wish to leverage the full value of the sensitive data whilst still maintaining individual privacy. This goal is driving new lines of research in statistics, information theory, and machine learning \cite{jordon2020syntheticdatareview, charest2011can, jordon2022synthetic}. 

Producing synthetic datasets is one promising approach. Generative models like generative adversarial networks (GANs) and variational autoencoders (VAEs) can be used to synthesise new examples from the same probability distribution as their training dataset. The generative model can be trained on site, then either released directly to analysts or used to synthesise a dataset for them. But numerous studies have demonstrated how synthetic data can still leak private information because generative models often memorise training examples and regurgitate them, which undermines the privacy goals \cite{jordon2022synthetic, hayes2017logan-attack}.

Fortunately, we have tools like differential privacy \cite{dwork2014differentialprivacy} that allow us to work with sensitive data in privacy-preserving ways. Differential privacy (DP) places requirements on the mechanism generating the synthetic data, rather than on the data itself. DP requires that changing which samples are included in training the generative model does not affect the synthesised output so much that an outsider could infer any information about them, thus providing individual privacy whilst retaining high-level statistical structure. 

With the advent of differentially-private stochastic gradient descent (DP-SGD) \cite{song2013DP-SGD, abadi2016momentsaccounting}, it has become easier to train neural networks under differential privacy constraints -- clipping the gradients and injecting noise into them. This approach allows us to release the model, and any synthetic data it generates, with provable guarantees on individual privacy.

But there are trade-offs between quality and privacy \cite{jordon2022synthetic}. Adding noise destabilises the training of generative models and lowers the fidelity of the synthetic data, hampering its utility for analysis and machine learning applications. This creates a great need for algorithmic optimisation, as we strive to extract the most value from the data for any desired level of privacy constraints. Improved generative models yield higher quality data without further compromising privacy, allowing researchers to make better use of the wealth of sensitive datasets. One of the most promising new directions for efficient generative modelling is the rise of diffusion models, which learn the underlying distribution through iterative denoising \cite{ho2020denoising, sohl2015deep}. This makes them much more stable during training compared to GANs and VAEs \cite{yang2022diffusion}, which suggests immense potential for improving the quality-privacy trade-off in sensitive data settings.

However, the biggest challenge in optimising generative models under the quality-privacy trade-off lies in representing the raw data to the model. Much of the most valuable sensitive data, such as electronic health records (EHRs) and financial information, is tabular \cite{jordon2020syntheticdatareview, jordon2022synthetic, kadra2021well}. Tabular data is notoriously tricky to synthesise \cite{xu2019CTGAN} and is thus underserved, despite being ubiquitous as spreadsheets and relational databases. There is an unmet need for effective tools in this modality \cite{borisov2022deeptabular, shwartz2022tabular}. But, the properties of tabular data are particularly challenging to privacy-preserving synthesis, due to the mix of discrete and continuous types, non-Gaussian distributions, high-cardinality, and sparsity. 

Previous work on synthesising tabular data falls into three camps. Early work leaned on classifiers, such as inverted decision trees or vector machines, but struggled to balance classifier accuracy against data leakage risks \cite{jordon2020syntheticdatareview}, as DP-SGD relies on gradient-based algorithms and could not be used. The second wave of work applied VAEs and GANs (often with DP-SGD), but explicitly restricted the scope to single-type tabular data in highly-constrained formats. Whilst this boosted performance, the techniques are difficult to transfer across different datasets and contexts. Recently, more research has been done on mixed-type tabular data synthesis, with most researchers opting to use some combination of reversible data transformations alongside hybrid architectures and loss functions. Most of these approaches suffer from extreme implementation complexity, a loss of semantics during reverse transformation, and loss of correlational information across features of different types.  

Our work tackles these challenges of sensitive tabular data and improves the quality-privacy trade-off of synthetic data generators for this important modality.

\section{Problem statement and research questions}\label{sec:problem-and-research-Qs}
\paragraph{Problem statement} Improve the fidelity-privacy trade-off of generative models, with the goal of synthesising higher-quality tabular datasets without compromising individual privacy.

\paragraph{Research questions} We decompose the problem statement into the following research questions:
\begin{enumerate}
    \item \textbf{Increased data efficiency}: Can we make better use of the sensitive training data and thus produce a higher-fidelity synthetic dataset with the same privacy budget?
    \begin{enumerate}
        \item \textbf{Improved tabular representation}: Can we improve the way mixed-type tabular data is represented to the models and thus learn with higher-fidelity under the same privacy budget?
        \item \textbf{Sample augmentation}: Can we augment the sensitive data samples such that the models can learn more at each training step under the same privacy budget?
    \end{enumerate}
    \item \textbf{Increased training efficiency}: Can we implement generative models that produce high-fidelity data in fewer training steps and thus use less privacy budget?
\end{enumerate}

\section{Overview and main contributions}
We begin by providing the reader with extensive technical details and intuitive explanations of relevant background topics in Chapter \ref{ch:background}. We cover the challenges of applying deep learning to mixed-type tabular datasets, differential privacy and the techniques for utilising it to train neural networks, a comparison of leading generative modelling paradigms, and the approaches to evaluating synthetic data in the context of sensitive tabular data. Chapter \ref{ch:related-work} presents the findings of our extensive literature review on generative models for tabular synthesis, comparing and summarising the most notable related works. 

We implemented novel end-to-end tabular models that utilise attention mechanisms to learn better representations of mixed-type tabular data. Whilst these performed well as privatised autoencoders, their size and complexity made sampling new data too compute-intensive and was detrimental to privacy efficiency. We detail our designs and experiments in Chapter \ref{ch:end-to-end}.

To explore the potential of the diffusion paradigm, we developed the first differentially-private diffusion model for tabular data -- TableDiffusion. Chapter \ref{ch:tabular-DMs} details how we arrived at our novel design, along with the algorithms for reproducing two variants of our diffusion model. To evaluate our new diffusion models, we designed a comprehensive benchmarking experiment to compare them against leading models across multiple real-world datasets at varying privacy levels. We implemented four exemplary GAN-based models for the benchmark: 
\begin{enumerate}
    \item PATE-GAN (\citet{jordon2018PATE-GAN})
    \item DP-WGAN (\citet{frigerio2019DP-WGAN})
    \item DP-auto-GAN (\citet{tantipongpipat2019DP-auto-GAN})
    \item CTGAN (\citet{xu2019CTGAN})
\end{enumerate}

Chapter \ref{ch:benchmarking} details the selection criteria and implementation details of these benchmark models, along with our experimental design, which controlled for architectures, preprocessing schemes, privacy mechanisms, hyperparameter tuning protocols, training process, and stochastic variables. This allowed for a direct comparison of the high-level paradigms. By implementing two variants of our TableDiffusion model, we could perform ablation to understand which mechanisms resulted in improved performance. The results of the benchmarking are presented in Chapter \ref{ch:results-discussion}. We used a robust suite of metrics to quantitively evaluate the fidelity of the synthetic datasets and supported this with qualitative visualisations. 

Our key finding is that the diffusion paradigm appears to be vastly more data-efficient and privacy-efficient than the adversarial paradigm, due to the augmented re-use of each batch, the smoother gradient updates, and increased resilience to sparse representations. Our novel diffusion models produced higher-fidelity synthetic data than leading GAN-based approaches across privacy levels, avoiding mode collapse and achieving state-of-the-art performance. We also assessed the loss and privacy curves during model training, finding that the diffusion paradigm results in vastly more stable and privacy-efficient training. We highlight a few limitations of this work in Chapter \ref{ch:limits-and-future}, along with suggestions for future work, before presenting the final conclusions in Chapter \ref{ch:conclusions}.

\chapter{Background}\label{ch:background}

This section presents extensive technical details and intuitive explanations of relevant background topics for generating and evaluating synthetic tabular data under differential privacy.

\section{Mixed-type tabular data}\label{sec:mixed-type-tab-data}

Over the past decade, much of deep learning research and application has focussed on image data. Whilst high-dimensional and complex, images are comprised of regular and homogeneous feature types -- pixel values, arranged in grids, with identical ranges and similar distributions. However, much of the most useful and underutilised data in the world is in the form of mixed-type tabular data, from spreadsheets to relational databases \cite{shwartz2022tabular, borisov2022deeptabular}. This is particularly true in fields like healthcare and biomedicine \cite{kadra2021well, ulmer2020trust, somani2021deep, borisov2021robust}. 

\begin{figure}[ht]
    \centering
    \includegraphics[width=0.75\textwidth]{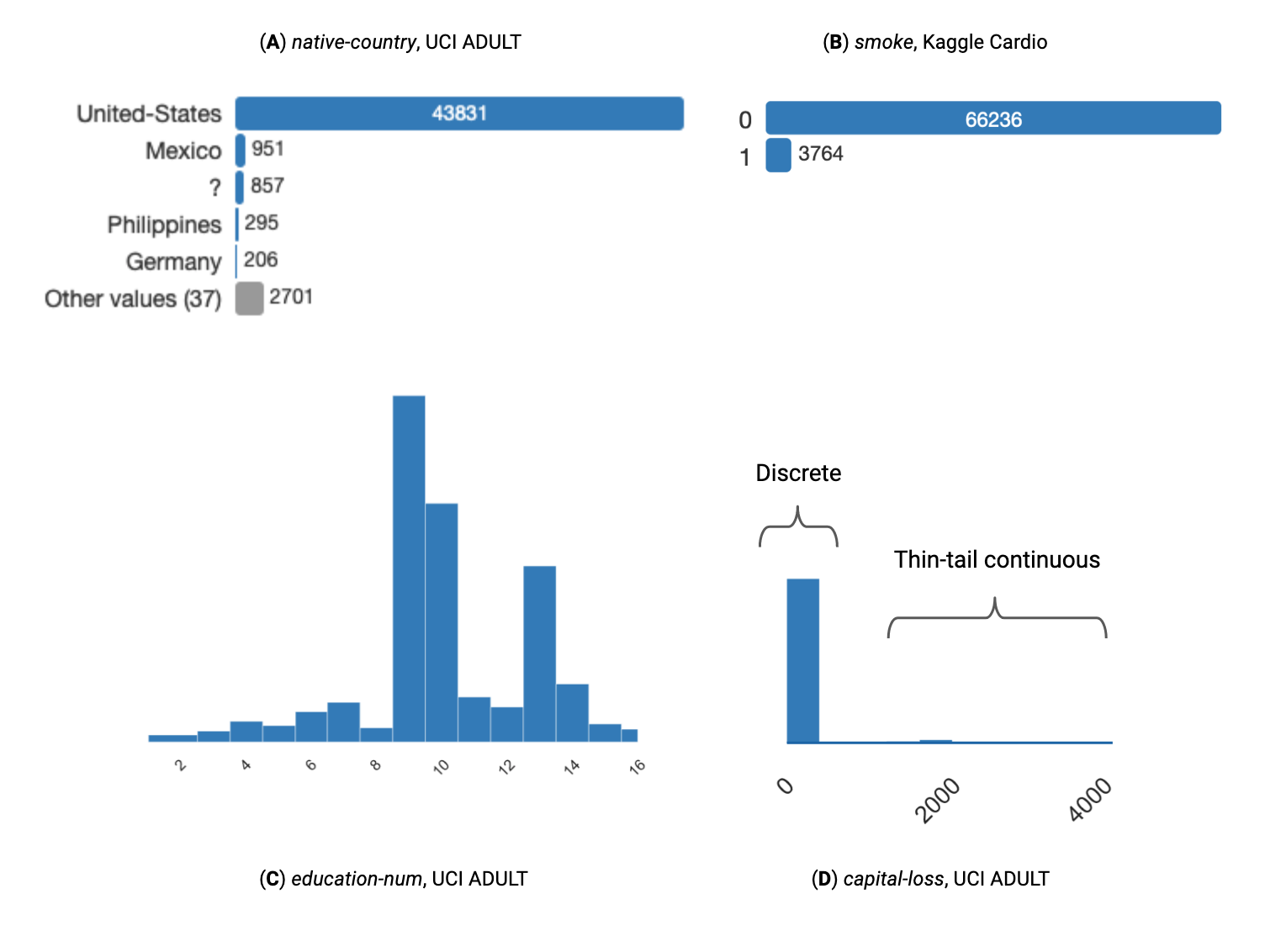}
    \caption{Examples of challenging tabular data features from real-world datasets used in this research (see Section \ref{sec:datasets}). (A) shows high cardinality. (A) and (B) show imbalanced categories. (C) shows a non-Gaussian, multi-modal distribution with long left tail (skewed). (D) shows a hybrid feature with many 0 values and some continuous values on a skewed long right tail.}
    \label{fig:tabular-data-challenges}
\end{figure}

In the context of sensitive tabular datasets, each of the $m$ rows of the table $\mathbf{X}$ correspond to one of $m$ real individuals and each of the $n$ columns represents some attribute measured, such as height, salary, or diagnosis. The goal of privatised synthesis is to produce a synthetic dataset $\hat{\mathbf{X}}$ that captures the joint distribution of the $n$ columns, without revealing enough information to personally re-identify any of the $m$ real individuals. This modality presents a number of challenges for deep learning applications, all of which are exacerbated when trying to synthesise new data using generative models \cite{jordon2022synthetic, xu2019CTGAN}. We illustrate some real-world examples in Figure \ref{fig:tabular-data-challenges}. 

\paragraph{Mixed discrete and continuous features} Tabular data from the real-world is a blend of different data types, with a variety of discrete and continuous columns. Neural networks are designed to work with normalised floating points, so extensive preprocessing is required for tabular data, with different techniques for each feature. But the choice of preprocessing steps biases the data towards some features over others, introduces sparsity, and loses correlation information between features, all of which negatively impacts the training process and lowers the fidelity of synthetic data \cite{xu2019CTGAN}. Importantly, the preprocessing steps must be \textit{reversible} in order to reconstruct tabular data from synthesised samples. 

\newpage

\paragraph{Hybrid features} Some columns of a table act as two or more features and are particularly challenging to represent. For example, a column in a spreadsheet that indicates salary might be empty, null, or zero for most instances, but then a continuous value with a long-tail distribution for others. Treating these hybrid features as a single feature makes it incredibly challenging for a neural network to model and results in post-processing artefacts. Converting the column into multiple features requires custom preprocessing for each dataset.

\paragraph{Non-Gaussian distributions} Unlike image data, where pixel values follow a Gaussian-like distribution that can be normalised within the range $[-1, 1]$ via a min-max transformation, tabular data often contains continuous values that do not follow a Gaussian distribution. Tabular data may exhibit a range of complex distributions. Multimodal distributions have multiple peaks or high-density regions in the data. Long-tailed distributions have far more outliers than the Gaussian distribution, while skewed distributions have a long tail on one side. Accurately capturing these distributions is crucial for generating synthetic data that truly represents the underlying data structure, but neural networks are optimised for Gaussian inputs and struggle to output non-Gaussian distributions, requiring practitioners to engineer custom pre- and post- processing pipelines \cite{borisov2022deeptabular}.

\paragraph{High cardinality and imbalanced categoricals} Certain discrete features in tabular data can take on a large number of unique values. High-cardinality categorical features pose a challenge due to the increased computational complexity. Effectively, each category becomes transformed into a feature. This also creates a bias towards high-cardinality features at the expense of other features in the data, as the model is not aware of the relative weightings of its input \cite{xu2019CTGAN}. In many real-world scenarios, certain classes in categorical data are represented far more than others. Generative models often fail to represent the less-frequent classes due to mode collapse. Moreover, highly-infrequent classes have a much higher privacy cost, as they are outliers that are much easier to identify. Often, these infrequent cases are highly important to some downstream task, such as identifying rare diseases, making for a challenging fidelity-privacy trade-off \cite{jordon2022synthetic}.

\paragraph{Sparsity} Many real-world tabular datasets have missing values or sparse features, where most of the entries are zero. In healthcare, for example, certain tests or procedures may only be performed for a small subset of patients. Generative models must be able to handle and reproduce such sparsity patterns. Moreover, most techniques for vectorising discrete features (e.g. one-hot encoding) introduce sparsity in the transformed data \cite{xu2019CTGAN}.

\subsection{Deep learning for tabular data}

The application of deep neural networks to tabular data, especially for tasks like inference or data generation, remains a formidable challenge. \citet{borisov2022deeptabular} categorise the research targeting deep learning for tabular data into three groups: data transformation techniques, robust regularisation models, and specialised architectures.

\begin{enumerate}
    \item \textbf{Data transformation techniques}: These methods focus on transforming both categorical and numerical data types into a shared representation, thus enabling deep neural network models to harness the underlying information more effectively. \textit{Single-dimensional} encodings transform each feature in isolation. This makes for simple, fast, and reversible transformations, but loses information about the relationships between features, biases the representation, and often introduces a high degree of sparsity. On the other hand, \textit{multi-dimensional} encodings attempt to transform all the features of an entire record. This can help preserve information between features and produce dense and high-fidelity representations, at the expense of added complexity. They also make the reverse transformations much more difficult to implement (or even impossible). Most work on synthesising tabular data thus makes use of single-dimensional encodings.
    \item \textbf{Regularisation models}: Two key challenges of deep learning on tabular data are pronounced nonlinearity and model complexity. To remedy this, numerous regularisation techniques have been explored, typically materialising as specialised loss functions. Whilst helpful in supervised settings, these are of little help in the challenge of reconstructing tabular data in our generative setting.
    \item \textbf{Specialised architectures}: Tabular data may necessitate bespoke deep neural network architectures. Two approaches have garnered significant attention. \textit{Hybrid models} seek to combine the strengths of classical machine learning techniques, such as decision trees, with the capabilities of neural networks. This often comes at the expense of not being fully differentiable. Concurrently, \textit{transformer-based models}, leveraging attention mechanisms, have risen to prominence. The advent of transformer architectures tailored for tabular data has spurred substantial research momentum. Exemplars of this trend are TabNet \cite{arik2021tabnet}, TabTransformer \cite{huang2020tabtransformer}, and SAINT \cite{somepalli2021saint}. In supervised settings, these methods show great promise and many are witnessing increased adoption \cite{borisov2022deeptabular}. However, in the generative paradigm, there is the added challenge of mapping synthetic data from the representation space back into mixed-type tables, which proves vastly more challenging than the original representation learning. There is currently a dearth of research on reversible tabular representation learning and there appear to be no successful works on end-to-end generative tabular networks to date. We explore the use of end-to-end attention-based networks for tabular data synthesis in Chapter \ref{ch:end-to-end}.
\end{enumerate}

\section{Differential privacy (DP)}

Differential privacy, introduced by \citet{dwork2014differentialprivacy}, provides a mathematically rigorous definition of privacy in the context of statistical databases. This framework offers strong privacy guarantees by ensuring that the addition or removal of a single database entry does not significantly change the outcome of any statistical query. This is achieved in practice by adding controlled statistical noise.

Consider a dataset $\mathbf{X}$ that consists of $m$ individuals. Two datasets $\mathbf{X}'$ and $\mathbf{X}''$ are considered neighbours if one can be obtained from the other by the addition or deletion of any one of the $m$ individuals. Differential privacy requires that an algorithm $\mathcal{M}$ produce similar outputs on neighbouring datasets. Intuitively, we want to ensure that our statistical algorithm, $\mathcal{M}$, when operating on a database, will yield similar results even if the data of a single individual is added or removed. This similarity ensures that no individual's data has a disproportionate impact on the result, thus preserving privacy. Formally, we use the notion of $(\epsilon, \delta)$-differential privacy:

\begin{definition}[$(\epsilon, \delta)$-Differential privacy] \label{def:differential_privacy}
For $\epsilon, \delta > 0$, an algorithm $\mathcal{M}$ is $(\epsilon, \delta)$-differentially private if, for any pair of neighbouring databases $\mathbf{X}'$, $\mathbf{X}''$, and any subset $S$ of possible outputs produced by $\mathcal{M}$,
$$
\operatorname{Pr}[\mathcal{M}(\mathbf{X}') \in S] \leq e^{\epsilon} \cdot \operatorname{Pr}[\mathcal{M}(\mathbf{X}'') \in S] + \delta
$$
\end{definition}

In this definition, $e^\epsilon$ can thus be thought of as an amplification factor that allows a slight difference in probabilities, with the degree of difference controlled by $\epsilon$. The smaller $\epsilon$ is, the closer to 1 this factor is, and the less difference we allow between the probabilities for the two databases. Thus, $\epsilon$ is the parameter controlling the privacy level, with lower values correspond to stronger privacy guarantees. In real-world applications, $\epsilon$ is typically targeted at values between $0.5$ and $10.0$, with larger values being acceptable when the number of individuals $m$ is larger and the data is less sensitive, such as in a national census. The parameter $\delta$ is usually fixed at $10^{-5}$ and is a relaxation used to make $\epsilon$-DP attainable in practice \cite{dwork2014differentialprivacy}.

Intuitively, if $\mathcal{M}$ is differentially private, it is very unlikely that some third party comparing $\mathcal{M}(\mathbf{X}')$ with $\mathcal{M}(\mathbf{X}'')$ could tell if any of the $m$ individuals differed between $\mathbf{X}'$ and $\mathbf{X}''$. This translates to the privacy of all $m$ individuals being preserved. Effectively, $\mathcal{M}$ gives all individuals plausible deniability as to whether their data was included. For an intuitive analogy, consider a fruit salad recipe. If the recipe is differentially private, then adding or removing any single piece of fruit should not alter the overall taste of the salad enough to identify the piece of fruit changed.

\subsection{Properties of DP}\label{ssec:properties-of-dp}
Differential privacy has a number of useful mathematical properties \cite{dwork2014differentialprivacy}. The two most important in our context are:

\paragraph{Composability} If you combine $k$ $(\epsilon, \delta)$-differentially private algorithms, the composed algorithm is at least $(k\epsilon , k\delta)$- differentially private. This is invaluable in iterative algorithms like the gradient descent approach used to train neural networks, as we can account for the privacy expended across multiple steps. We explore this in detail in Section \ref{sec:DPSGD}. 
\paragraph{Post-processing guarantee} If an algorithm $\mathcal{M}$ is $(\epsilon, \delta)$-differentially-private, then any function $f(\mathcal{M})$ is at least $(\epsilon, \delta)$-differentially-private. This means that if we train a differentially-private generative model, then any data we synthesise with that model will also be differentially-private to at least the same extent, allowing us to release the model and the synthetic data with strong guarantees.

\subsection{Sensitivity}\label{sec:sensitivity}
In the context of differential privacy, an important concept is that of \textit{sensitivity}. The sensitivity of a function quantifies the maximum difference in the function's outputs when computed on two neighbouring datasets. Formally, for a function $f$ that maps datasets to $\mathbb{R}^d$, the sensitivity $S_f$ is defined as follows:

\begin{equation} \label{eq:sensitivity}
S_f := \max_{\mathbf{X}', \mathbf{X}''} ||f(\mathbf{X}') - f(\mathbf{X}'')||_2
\end{equation}
where the maximum is over all pairs of datasets $\mathbf{X}'$ and $\mathbf{X}''$ that differ by at most one element (i.e. neighbours).

Intuitively, sensitivity measures the potential impact of a single individual's data on the function's output. A function with low sensitivity has a stronger privacy guarantee because the contribution of each individual to the output is small. For example, consider a simple counting query, such as the number of individuals in a database who have a certain attribute. In this case, adding or removing a single individual can change the count by at most one, so the sensitivity of the counting query is one. As we shall see in Section \ref{sec:DPSGD}, the calculation of sensitivity plays a crucial role in applying differential privacy to real-world algorithms by determining the amount of noise that needs to be added to ensure a particular level of privacy.

\subsection{Rényi differential privacy (RDP)} 
In the past decade, theoretical improvements have been made to the original DP formulation that afford desirable properties for real-world applications. One of these is Rényi DP, which is now the default in most DP libraries \cite{opacus}. Rényi differential privacy (RDP) \cite{mironov2017RenyiDP} is a relaxation of DP that allows for tighter estimates of privacy bounds when composing heterogeneous mechanisms, whilst maintaining the post-processing guarantee. Formally,

\begin{definition}[$(\alpha, \epsilon)$-RDP]
An algorithm $\mathcal{M}$ is $(\alpha, \epsilon)$-RDP if, for all neighbouring databases $\mathbf{X}', \mathbf{X}''$:
$$
D_\alpha \left( \mathcal{M}(\mathbf{X}') \mid \mid \mathcal{M}(\mathbf{X}'') \right) \leq \epsilon
$$
where $D_{\alpha}(P \| Q):=\frac{1}{\alpha-1} \log \mathbb{E}_{x \sim \mathbf{X}}\left(\frac{P(x)}{Q(x)}\right)^{\alpha}$ is the Rényi divergence of order $\alpha$ between distributions $P$ and $Q$.
\end{definition}

The essence of RDP lies in its use of Rényi divergence, a measure of dissimilarity between two probability distributions, instead of the probability ratio used in the standard DP. By using this measure, RDP allows us to handle multiple rounds of privacy-preserving noise addition with more precision, leading to better privacy accounting. Rényi DP can be converted back to $(\epsilon, \delta)$-differential privacy after composing multiple DP algorithms, allowing for better intuitions and direct comparison with established techniques \cite{mironov2017RenyiDP}. RDP has become the leading approach for applying differential privacy to neural networks \cite{opacus} as it allows for the most precise estimation of privacy expenditure when composing the iterative steps of gradient descent and heterogeneous neural modules.

\newpage
\section{Differentially-private stochastic gradient descent (DP-SGD)}\label{sec:DPSGD}

Gradient descent is a fundamental optimisation algorithm in machine learning that is utilised for training a variety of models, including neural networks. The aim is to minimise some loss function $\mathcal{L}$, which quantifies the error between the model predictions and the true data. The model's parameters (or weights) are iteratively updated by taking steps in the direction of steepest descent in the loss function. In a traditional gradient descent step, the weights $\boldsymbol{\theta}$ of the model are updated according to the gradient of the weights with respect to the loss function, $\nabla_{\boldsymbol{\theta}} \mathcal{L}$, weighted by some learning rate $\eta$:
\begin{equation}
    \boldsymbol{\theta} \gets \boldsymbol{\theta} - \eta \cdot \nabla_{\boldsymbol{\theta}} \mathcal{L} 
\end{equation}

To apply the guarantees of differential privacy to training neural networks, \citet{song2013DP-SGD} introduced differentially-private stochastic gradient descent (DP-SGD). At each training step, the gradients are calculated as normal, but modified before the model weights are updated. Firstly, the $L_2$ norm of the gradients is clipped to some constant $C \geq 1$. Secondly, we sample noise from the standard multivariate Gaussian (scaled by noise multiplier $\sigma$ and the clipping constant $C$) and add this to the clipped gradients. Formally, our weight update formula becomes:  
\begin{equation}\label{eq:dp-sgd}
    \boldsymbol{\theta} \gets \boldsymbol{\theta} - \eta \cdot \operatorname{CLIP}\left(\nabla_\theta \mathcal{L}, C\right) + \mathcal{N}\left(\boldsymbol{0}, C^2 \sigma^2 \boldsymbol{I}\right)
\end{equation}
where $\operatorname{CLIP}\left(\nabla_\theta \mathcal{L}, C\right) := \nabla_\theta \mathcal{L} / \operatorname{max}(1, \frac{1}{C} ||\nabla_\theta \mathcal{L}||_2)$

The clipping procedure ensures that the contribution of each individual training example to the overall gradient computation is bounded, helping to protect the privacy of each data point. The additional noise injection process further enhances the privacy protection by ensuring that the updated model parameters do not reveal too much information about any individual training example. The combined effect of gradient clipping and noise addition forms the basis of differential privacy guarantees in DP-SGD. For performance and robustness, the clipping, noising, and gradient updates are usually applied at the end of each batch. Because the privacy is enforced on the gradients themselves, superior optimisers (like Adam) can be used instead of simple SGD with a simple substitution, whilst maintaining the same guarantees. In modern implementations, Laplacian noise is used instead of Gaussian noise and Rényi DP is used in lieu of traditional DP, both providing better performance and tighter privacy bound estimation \cite{opacus, torfi2020RDPCGAN}.

\subsection{Privacy accounting}
DP-SGD gives us the method for privatising model training steps, but we need to accumulate the privacy cost over the course of multiple training steps and monitor when we have reached the privacy budget. This is known as \textit{privacy accounting}.  

We saw in equation \ref{eq:dp-sgd} that DP-SGD adds Gaussian noise $\mathcal{N}\left(\boldsymbol{0}, C^2 \sigma^2 \boldsymbol{I}\right)$ to the clipped gradients, with $C$ being the clipping constant (which we set to 1 in this work), and $\sigma$ being some noise multiplier we must specify. We also know from equation \ref{eq:sensitivity} that we can quantify the sensitivity $S_f$ of some function $f$ (in this case, our gradient update). Theorem 3.22 of \citet{dwork2014differentialprivacy} shows that a single application of this Gaussian mechanism to function $f$ of sensitivity $S_f$ satisfies $(\epsilon, \delta)$-DP if $\delta \leq \frac{4}{5} \exp(-(\sigma \epsilon)^2 / 2)$ and $\epsilon < 1$. Thus, by choosing $\sigma = \sqrt{2\log{\frac{1.25}{\delta}}}/\epsilon$ for the Gaussian noise mechanism, each gradient update step becomes $O(q\epsilon, q\delta)$-DP, where $q$ is the batch sampling ratio and $\epsilon \leq 1$.

Next, the composability property of differential privacy allows us to implement an ``accountant'' that accumulates this cost over gradient update steps as the training progresses, thus tracking the total privacy loss. Before each training step, we use the accountant to check our current total privacy loss, and only continue if it is sufficiently below our privacy budget (i.e. the target $\epsilon$ value). 

Directly applying the composition theorem of DP gives the tightest privacy bounds of DP-SGD as $O(q\epsilon\sqrt{T\log{\frac{1}{\delta}}}, Tq\delta)$-DP, for $T$ training steps and batch sampling ratio $q$. But \citet{abadi2016momentsaccounting} introduced the Moments Accountant, which achieves a much tighter privacy bound of $O(q\epsilon\sqrt{T}, \delta)$-DP by more accurately estimating the privacy loss composition by tracking detailed information (higher moments) of the privacy loss. 

Armed with the Moments Accountant, we can fix the value of $\delta$ and target some privacy budget $\epsilon$ by specifying the noise multiplier $\sigma$, the batch sampling probability $q$ (each epoch consists of $1/q$ batches), the clipping constant $C$, and the number of epochs $E$ (so the number of training steps is $T=E/q$), all of which can be established before we begin training under DP-SGD.

\section{Generative models}

Generative models are a category of machine learning model designed to learn the true underlying data distribution of the training set, enabling us to sample new data points from the same distribution. The three types of generative models discussed in this section are Variational Autoencoders (VAEs), Generative Adversarial Networks (GANs), and Diffusion Models (DMs).

\begin{figure}[ht]
    \centering
    \includegraphics[width=0.95\textwidth]{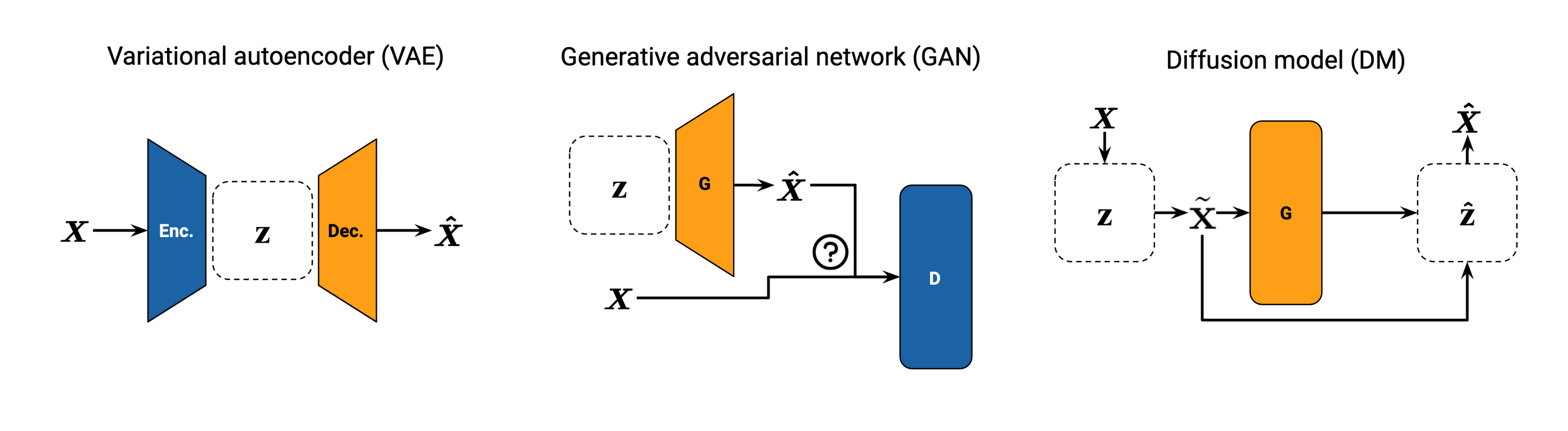}
    \caption{Illustrations of the three types of generative model relevant to tabular data synthesis. Coloured blocks represent Encoder (\texttt{Enc}), Decoder (\texttt{Dec}), Generator (\texttt{G}), and Discriminator (\texttt{D}) networks. A real dataset $\mathbf{X}$ is used in training the models to produce a synthetic dataset $\hat{\mathbf{X}}$. The latent space from which the models sample is denoted $\mathbf{z}$. Diffusion models make use of noised data $\tilde{\mathbf{X}}$.}
    \label{fig:generative-models-compared}
\end{figure}

\subsection{Variational autoencoders (VAEs)}

Variational Autoencoders (VAEs) \cite{kingma2013auto} are a type of generative model that learn to encode input data $\mathbf{x}$ into a defined latent space $\boldsymbol{z}$. After training the VAE, we can sample from the latent distribution $p(\mathbf{z})$ and decode the samples into new data $\hat{\mathbf{x}}$. 

VAEs consist of two neural networks: an encoder network $q_{\boldsymbol{\phi}}$ parameterised by weights ${\boldsymbol{\phi}}$ and a decoder network $p_{\boldsymbol{\theta}}$ parameterised by weights ${\boldsymbol{\theta}}$. The objective function maximised by VAEs, often called the Evidence Lower BOund (ELBO), is given by:

\begin{equation}
\mathbb{E}_{q_{\boldsymbol{\phi}}(\boldsymbol{z}|\mathbf{x})}\left[\log p_{\boldsymbol{\theta}}(\mathbf{x}|\boldsymbol{z})\right] - D_{\text{KL}}\left(q_{\boldsymbol{\phi}}(\boldsymbol{z}|\mathbf{x})||p(\boldsymbol{z})\right)
\end{equation}

where $D_{\text{KL}}$ denotes the Kullback-Leibler divergence. The first term encourages the reconstructed data $\hat{\mathbf{x}}$ to resemble the distribution of the original data $\mathbf{x}$, while the second term regularises the latent space $\boldsymbol{z}$ to follow a predetermined distribution (the pior), typically chosen to be a standard multivariate Gaussian.

The reparameterisation trick is employed to make the model differentiable, where the sampling operation is replaced by a deterministic function of the Gaussian parameters ($\boldsymbol{\mu}$ and $\boldsymbol{\sigma}$) and a random noise variable $\boldsymbol{\varepsilon}$. This allows backpropagation through the model, and hence end-to-end training of the weights. Samples from the latent distribution are therefore computed as $\boldsymbol{z} = \boldsymbol{\mu}_{\boldsymbol{\phi}}(\mathbf{x}) + \boldsymbol{\sigma}_{\boldsymbol{\phi}}(\mathbf{x}) \odot \boldsymbol{\varepsilon}$, where $\boldsymbol{\varepsilon} \sim \mathcal{N}(0, \boldsymbol{I})$.

\subsection{Generative adversarial networks (GANs)}
Generative Adversarial Networks (GANs) \cite{goodfellow2020generative} are comprised of two components: a generator network $G_{\boldsymbol{\theta}}$ (parameterised by weights $\boldsymbol{\theta}$), and a discriminator network $D_{\boldsymbol{\phi}}$ (parameterised by weights $\boldsymbol{\phi}$). The generator network takes as input random noise $\boldsymbol{z} \sim \mathcal{N}(\boldsymbol{0}, \boldsymbol{I})$ and outputs synthetic data. The discriminator network takes as input either real data from the original dataset $\mathbf{x}$ or synthetic data from the generator $\hat{\mathbf{x}} \gets G_{\boldsymbol{\theta}}(\boldsymbol{z})$, and outputs a probability of the input data being real. The objective function of GANs is defined as:

\begin{equation}
\min_{\boldsymbol{\theta}} \max_{\boldsymbol{\phi}} \mathbb{E}_{\mathbf{x}}[\log D_{\boldsymbol{\phi}}(\mathbf{x})] + \mathbb{E}_{\boldsymbol{z}}[\log(1 - D_{\boldsymbol{\phi}}(G_{\boldsymbol{\theta}}(\boldsymbol{z})))]
\end{equation}

In this min-max game, the generator seeks to generate data that the discriminator cannot distinguish from real data, while the discriminator aims to correctly classify real and synthetic data. This interplay optimises the generator to produce data closely resembling the original data distribution.

While GANs can produce high-quality samples, they are notoriously difficult to train due to the adversarial setup, which can lead to unstable dynamics during training, where the generator and discriminator do not reach an equilibrium. Additionally, GANs can suffer from the mode collapse problem, where the generator outputs a limited diversity of samples. To tackle these issues, Wasserstein GANs (WGANs) \cite{arjovsky2017wasserstein} were proposed, which employ the Wasserstein distance as a more stable objective function. A gradient penalty (GP) term is often added to the loss function to enforce the Lipschitz constraint and further stabilise training \cite{gulrajani2017improved}. The gradient penalty ensures that the gradients of the discriminator with respect to its input do not exceed a certain threshold, effectively regularising the discriminator and keeping the gradient norms under control. The gradient penalty term is defined as:

\begin{equation}
\lambda \left(\| \nabla_{\tilde{\mathbf{x}}} D_{\boldsymbol{\phi}}(\tilde{\mathbf{x}}) \|_2 - 1\right)^2
\end{equation}
where $\lambda$ is a weighting constant, and $\nabla_{\tilde{\mathbf{x}}}$ denotes the gradient with respect to a batch of sampled data $\tilde{\mathbf{x}}$.

Conditional GANs (CGANs) \cite{gauthier2014conditional} extend the capabilities of basic GANs by introducing conditional information to guide the generation process. Instead of generating data from random noise alone, CGANs utilise both noise and a conditional variable \( \boldsymbol{c} \) to produce data that not only resembles the real distribution but also adheres to specific conditions. Formally, the generator network \( G_{\boldsymbol{\theta}} \) of a CGAN takes as input both the random noise \( \boldsymbol{z} \sim \mathcal{N}(\boldsymbol{0}, \boldsymbol{I}) \) and the conditional variable \( \boldsymbol{c} \) to produce synthetic data \( \hat{\mathbf{x}} \gets G_{\boldsymbol{\theta}}(\boldsymbol{z}, \boldsymbol{c}) \). Similarly, the discriminator network \( D_{\boldsymbol{\phi}} \) is modified to accept both data (real or synthetic) and the conditional variable to produce the probability of the input data being real, \( D_{\boldsymbol{\phi}}(\mathbf{x}, \boldsymbol{c}) \). During training, this conditional formulation can be leveraged to upsample modes that are infrequently represented in the training data, which can help alleviate the mode-collapse issue. When synthesising data, CGANs have the added benefit of allowing conditional sampling of certain subsets of the data, which has many uses in synthetic data augmentation \cite{jordon2022synthetic}. Whilst quite common in GAN architectures, this conditional formulation can also be easily applied to other generative models where necessary.

\subsection{Diffusion models (DMs)}

Diffusion Models \cite{sohl2015deep, ho2020denoising} are a more recent type of implicit generative model that smoothly perturb data by adding noise, then reverse this process to generate new data. The original data $\mathbf{x}$ is gradually corrupted through a diffusion process over a sequence of steps $t \in [1, T]$. At each step $t$, Gaussian noise $\boldsymbol{z} \sim \mathcal{N}(\boldsymbol{0}, \boldsymbol{I})$ is scaled by a dynamic value $\beta_t$ and added to the real data. A single neural network is trained to separate the noise from the data at each step of this process. 

This training is performed by having the model predict one of three potential objectives: the original source data from noised data (denoising), the original source noise from the noised data (noise prediction), or some score function of noised data at any arbitrary noise level \cite{luo2022understanding}.

During sampling, synthetic data $\hat{\mathbf{x}}$ is generated by running the diffusion process in reverse. We first sample pure noise from a Gaussian distribution and then iteratively apply the model to subtract the noise until we reach the target distribution. This mechanism offers a tractable means to learn and sample from complex data distributions without making any assumptions about the true underlying distribution. In this work, we develop two novel DMs that are specifically tailored to mixed-type tabular data, one of the denoising variety and one of the noise prediction variety. We thus provide a more detailed explanation of diffusion models in Chapter \ref{ch:tabular-DMs} before describing our own implementations.

\section{Evaluating synthetic data}
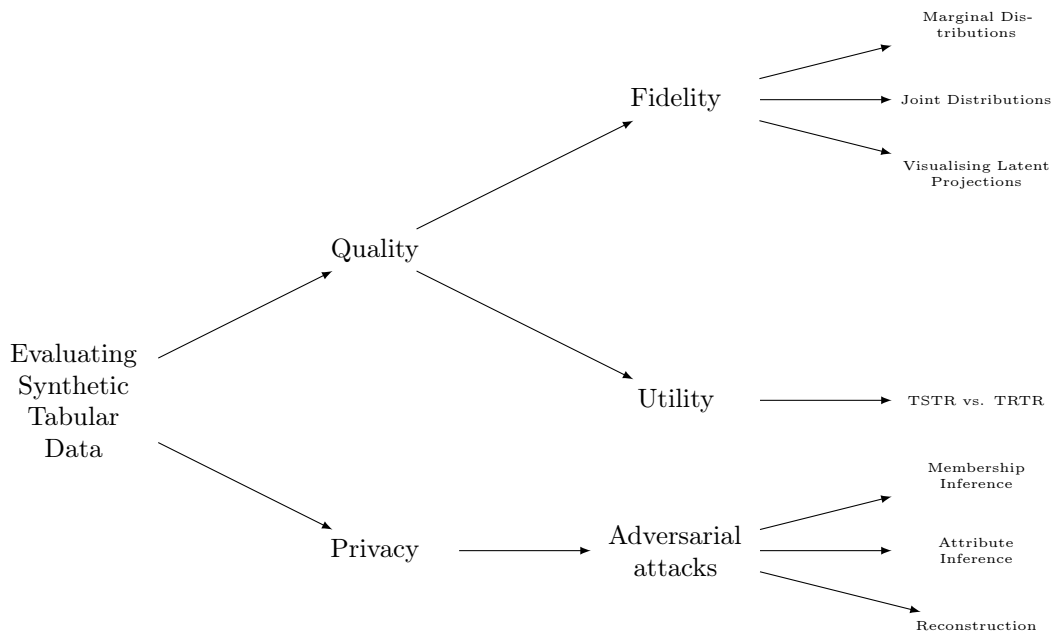
\begin{figure}[ht]
\centering
\begin{tikzpicture}[
    grow'=right,
    level distance=4cm,
    sibling distance=4cm,
    every node/.style={text width=2cm, align=center },
    edge from parent/.style={draw, -latex},
    level 3/.style={font=\tiny, text width=4cm, align=center, sibling distance=1cm, draw=none}
    ]

    \node {Evaluating Synthetic Tabular Data}
    child { node {Quality} 
        child { node {Fidelity} 
            child { node {Marginal Distributions} }
            child { node {Joint Distributions} }
            child { node {Visualising Latent Projections} }
        }
        child { node {Utility} 
            child { node {TSTR vs. TRTR} }
        }
    }
    child { node {Privacy}
        child { node {Adversarial attacks} 
            child { node {Membership Inference} }
            child { node {Attribute Inference} }
            child { node {Reconstruction} }
        }
    };

\end{tikzpicture}
\caption{A hierarchy for the evaluation of synthetic tabular data.}
\label{fig:eval_synthetic_data}
\end{figure}

The ideal synthetic data should substitute perfectly for the real data in any use-case, maintaining equivalent \textit{utility} and \textit{fidelity} while ensuring \textit{privacy}. These three properties of synthetic data are interrelated and often involve trade-offs, particularly between privacy and the other two aspects \cite{jordon2020syntheticdatareview, jordon2022synthetic}. However, quantifying these properties presents significant challenges which we briefly explore in this section. We illustrate the hierarchy of synthetic data evaluation techniques discussed in Figure \ref{fig:eval_synthetic_data}.

\paragraph{Utility versus Fidelity}
Utility and fidelity are two facets of synthetic data quality. The utility of synthetic data refers to its effectiveness for specific tasks, such as training a model or conducting a particular analysis. The fidelity of synthetic data, on the other hand, is a measure of its similarity to the real data from a statistical perspective. These two attributes are often conflated, but they are not synonymous nor perfectly correlated. In certain scenarios, it is possible to reduce fidelity while retaining utility, which could create opportunities for enhancing privacy. For this work, we focus on fidelity, as it is more quantifiable, application-agnostic, and generally correlates with utility \cite{jordon2022synthetic}.

\paragraph{Inception-based metrics}
The inception score (IS) and the Frechet Inception Distance (FID) are popular metrics for assessing the quality of synthetic images. IS uses the inception model, a deep convolutional neural network, to classify the images and computes the score based on the predictability and diversity of the classifications. FID, on the other hand, calculates the statistical distance between features extracted from the real and synthetic images using the inception model, with a lower FID indicating more similar and thus better-quality synthetic data. These metrics have proven effective in the image domain because they capitalise on properties specific to images, particularly the availability of pre-trained models like the Inception network that are accustomed to the distributions and characteristics of photographic data. However, they are unsuitable for evaluating synthetic tabular data because they rely on principles and models that are uniquely tailored to images. The field of synthetic tabular data lacks such standardised and reliable metrics, which impedes its progress. The development of equivalent measures for synthetic tabular data is a significant research challenge \cite{jordon2020syntheticdatareview}.

\subsection{Evaluating fidelity}

The \textit{fidelity} of synthetic data refers to how closely it matches the statistical distribution of the original data. This property is particularly challenging to quantify due to the curse of dimensionality, especially in the context of tabular data with mixed types and sparse distributions. Existing methods for evaluating fidelity fall into three main categories: marginal distributions, joint distributions, and visualisations of latent projections. The aim of this work was to improve the fidelity-privacy trade-off of generative models, with the goal of synthesising higher-quality tabular datasets. We therefore focus on assessing quality via a broad set of complementary fidelity metrics, that are also a good proxy for utility on arbitrary downstream tasks. We formalise the metrics and their properties in detail in Section \ref{sec:data-fidelity-metrics}, but give a high-level overview here. 

\paragraph{Marginal distributions} Comparing the real and synthetic data on a feature-by-feature basis is one of the simplest and most common measures of fidelity. This can be done visually, by plotting histograms of each feature, or analytically, by using statistical tests or distance measures. Synthetic data with high marginal fidelity is particularly useful for privately computing downstream statistics like averages and ranges. However, marginal fidelity is insufficient on its own, as it does not consider how the features interact or correlate. A trivial example is a synthetic dataset of heights and weights. It would be possible to perfectly match the distributions of each feature, but have nonsensical samples (e.g. where someone that is 2 metres tall only weighs 45 kilograms). For performing downstream correlation analysis or machine learning tasks, it is essential to also consider joint distribution fidelity.

\paragraph{Joint distributions} Comparing the real and synthetic data across all the features at the same time is challenging, particularly in tabular data \cite{jordon2022synthetic}. Some approaches try to approximate this by considering $k$-way marginal distributions for various subsets of $k$ features and then aggregating over them. Other approaches transform the data into some homogenous latent space and then use distance metrics (e.g. $L_1$, Euclidean, etc.). Many researchers also use variants of KL-divergence that are tailored to mixed-type data, but this always requires some weighting and aggregation over the different data types in the features, which introduces biases. Recently, some research \cite{alaa2022faithful} has extended the concepts of unsupervised precision and recall to efficiently compare the real and synthetic data. As with the distance metrics, this requires the datasets to be transformed into a homogenous latent space, which introduces biases based on the preprocessing methods used. 

\paragraph{Visualising latent projections} In the domain of images, one can simply look at synthetic data to evaluate its quality. This has enabled the image modality to pave the way in deep generative models. To try leveraging human perception in a similar fashion, some works have employed two-dimensional visualisations (e.g. heatmaps) on projections of the tabular data, produced via principal component analysis (PCA) or other dimensionality-reduction techniques. Whilst these are necessarily lossy representations of the joint distributions, they are extremely useful for quickly assessing synthetic data quality and detecting failures such as mode collapse using the human eye.

\subsection{Evaluating utility}
\textit{Utility} is measured in the context of specific downstream tasks. However, we typically do not have all the details of these tasks when evaluating generative models and their synthetic data. As a loose proxy measurement, one common approach is the Train on Synthetic Test on Real (TSTR) paradigm, where models are trained on synthetic data and then evaluated on real data for a collection of supervised learning tasks. The performance of these evaluation models is often compared to that of models trained on the real data -- Train on Real Test on Real (TRTR) -- to calculate a utility ratio. However, estimating utility in this way is heavily biased by the selection of supervised learning tasks and evaluation models \cite{jordon2022synthetic}. Fortunately, a strong correlation usually exists between the utility and fidelity of synthetic data. High-fidelity data, which captures as much information as possible about the original distribution (within privacy constraints), is likely to have high utility, though the inverse is not always true \cite{jordon2022synthetic}.

\subsection{Evaluating privacy}
Prior to DP-SGD adoption, researchers used multivariate distance metrics, such as Euclidean Distance to the Closest Record, to compare the real and synthetic data for memorised examples \cite{park2018table-GAN}. Indeed, this was how many of the failure modes of synthetic data were first detected. But these metrics fail to account for the effects of regional density on sensitivity (see Section \ref{sec:sensitivity}) and are largely irrelevant when training under differential privacy. But, whilst DP-SGD allows for the training of generative models with formal privacy guarantees, validating the privacy is vital in two specific cases: when modifying the core DP-SGD algorithm, or when designing generative models for real-world applications, where implementation differences could potentially lead to sensitive data leaks. In these cases, it is important to explicitly validate the theory by conducting empirical privacy tests \cite{jordon2022synthetic}. This is especially true when the model is being released as opposed to just the synthetic data. To validate the privacy of a model, researchers must adopt an adversarial approach and can implement three kinds of attack, all based on training a supervised model. Membership inference attacks \cite{shokri2017membership} aim to predict whether a specific individual was part of the training data, given the synthetic data. Attribute inference attacks \cite{kosinski2013private} aim to recover attributes of a specific individual, given the synthetic data and some outside knowledge of their other attributes. Reconstruction attacks \cite{dinur2003revealing} aim to extract entire records from the original dataset based on the synthetic output of the model.

\subsection{Ongoing research}
Given the complexities and challenges of measuring privacy, utility, and fidelity, evaluating synthetic data requires adopting a variety of approaches. This domain is still an active area of research \cite{jordon2020syntheticdatareview, jordon2022synthetic}, with no one-size-fits-all solution currently available. In this work, we favour fidelity metrics for their clear measurability, task-agnostic applicability, and general correlation with utility.

\chapter{Related work}\label{ch:related-work}
This chapter examines the most notable prior works on tabular data synthesis and differentially-private generative models.  

\section{VAE approaches}
Whilst there has been some work \cite{acs2018DP-VAE, xu2019CTGAN, abay2018DP-SYN, chen2018DP-AuGM_DP-VaeGM} applying (variational) autoencoders to tabular dataset synthesis, the popularity and performance of GANs on image data has drawn most of the research focus to GAN-based synthesis \cite{jordon2022synthetic}. A number of authors \cite{torfi2020CorGAN, torfi2020RDPCGAN, chen2018DP-AuGM_DP-VaeGM} have utilised autoencoders as part of their GAN-based methods in an attempt to better represent the data, thus helping to stabilise GAN training.

VAEs are prescribed models, whereby we must explicitly specify a latent distribution and penalise the model for drifting from that distribution during training. This affords approximate likelihood estimation, stable training, and fast sampling, but limits the extent to which the model can learn the true distribution of the data \cite{BondTaylor2021DeepGM}. GANs, on the other hand, are implicit models that learn the data distribution internally. This prevents us from being able to perform likelihood estimation and compression, but these are not important in the context of data synthesis. Whilst the adversarial learning paradigm does make training far less stable \cite{BondTaylor2021DeepGM}, an abundance of techniques have been developed to mitigate the issue. Overall, this led to GANs winning over VAEs and other generative models \cite{BondTaylor2021DeepGM, jordon2022synthetic}. 

Whilst there is some work \cite{acs2018DP-VAE, xu2019CTGAN, abay2018DP-SYN, chen2018DP-AuGM_DP-VaeGM} highlighting good performance of VAEs for differentially-private dataset synthesis, this has often been done in the context of images or single-type structured data. In the case of TVAE \cite{xu2019CTGAN}, which worked well on mixed-type tabular data and outperformed the CTGAN model \cite{xu2019CTGAN} in some cases, the authors still recommend GANs over VAEs for being easier to privatise and adapt to different datasets. Indeed, the state-of-the-art mixed-type tabular generators are currently GAN-based \cite{xu2019CTGAN, jordon2022synthetic}. For these reasons, this work focuses on GAN-based models that have demonstrated the best performance on tabular data synthesis.

\newpage
\section{GAN approaches}\label{sec:related-gans}
An overview of the key information for GAN-based approaches is presented in Table \ref{tab:related-work}.

\begin{table}[ht]
\centering
\caption{Overview of GAN methods from related work. Models benchmarked in this study indicated in bold.}
\label{tab:related-work}
\begin{footnotesize} 
\newcolumntype{Z}{>{\centering\arraybackslash}p{1.8cm}} 
\newcolumntype{Y}{>{\centering\arraybackslash}p{3.3cm}} 
\begin{tabularx}{\textwidth}{ZZZZZY} 
\hline
\textbf{Model} & \textbf{Architecture} & \textbf{Privacy} & \textbf{Comparison} & \textbf{Datasets} & \textbf{Evaluation} \\
\hline
MedGAN \cite{choi2017medGAN} & Hybrid AE and GAN & -- & Ablation only & Discrete (binary and integer) EHRs & Fidelity (marginal comparison, expert qualitative evaluation). Utility (TSTR). \\
\hline
    TableGAN \cite{park2018table-GAN} & DCGAN and auxiliary classifier & -- & Baseline DCGAN\cite{radford2015DCGAN} & 4 real tabular datasets & Privacy (Euclidean distance to closest record, membership attacks). Fidelity (marginal comparison). Utility (TSTR). \\
\hline
DPGAN \cite{xie2018DPGAN} & (2D convolutional) WGAN & DP-SGD & -- & Images (MNIST) and discrete (binary) EHRs & Fidelity (dimension-wise probability). Utility (TSTR). \\
\hline
\textbf{PATE-GAN} \cite{jordon2018PATE-GAN} & GAN (multiple discriminators) & PATE & DPGAN\cite{xie2018DPGAN} and baseline GAN & 6 datasets of binary and continuous data & Utility (TSTR, TRTR, TSTS with ranking agreement). \\
\hline
DP-AC-GAN \cite{beaulieu2019DP-AC-GAN} & Auxiliary classifier GAN with 2D convolutions & DP-SGD & Unprivatised AC-GAN & 2 EHR datasets (one mixed-type data) & Fidelity (feature correlation). Utility (TSTR, expert review). \\
\hline
\textbf{CTGAN} \cite{xu2019CTGAN} & Conditional WGAN (with gradient penalties) & -- & Bayesian methods, TableGAN\cite{park2018table-GAN}, MedGAN\cite{choi2017medGAN}, VeeGAN\cite{srivastava2017veegan}, TVAE\cite{xu2019CTGAN}, ablation & 15 mixed-type datasets (7 simulated, 8 real) & Fidelity (simulated likelihood fitness), Utility (TSTR). \\
\hline
\textbf{DP-WGAN} \cite{frigerio2019DP-WGAN} & WGAN (with clipping decay) & DP-SGD & Baseline GAN and ablation & 3 non-mixed tabular datasets (1 simulated, 2 real) & Utility (TSTR). Privacy (membership inference attacks). \\
\hline
RDP-GAN \cite{torfi2020RDPCGAN} & (Convolutional) WGAN within AE & RDP-SGD & DPGAN\cite{xie2018DPGAN}, PATE-GAN\cite{jordon2018PATE-GAN}, ablation & 6 real mixed-type datasets & Utility (TSTR). \\
\hline
\textbf{DP-auto-GAN} \cite{tantipongpipat2019DP-auto-GAN} & WGAN within AE & RDP-SGD & DP-WGAN\cite{frigerio2019DP-WGAN}, DP-VAE\cite{acs2018DP-VAE}, DP-SYN\cite{abay2018DP-SYN} & 2 real EHR datasets (one mixed-type) & Fidelity (marginal distributions, KL-divergence, Jenson-Shannon distance, PCA scatterplots). Utility (TSTR).  \\
\hline
\end{tabularx}
\end{footnotesize} 
\end{table}

\newpage
\paragraph{MedGAN} \citet{choi2017medGAN} created MedGAN, which was one of the earliest approaches in this domain. The model was a hybrid autoencoder and GAN that was able to produce good synthetic electronic health records (EHRs). However, the model did not incorporate any privacy preserving techniques and made no privacy guarantees. Moreover, the model was only able to work with discrete data types, mainly binary categoricals and integer counts. The model was not benchmarked against other techniques, but an ablation study was done to evaluate the effects of pretraining the outer autoencoder and performing batch averaging, with fidelity being assessed by marginal distribution comparison and expert qualitative evaluation, and utility being assessed by TSTR evaluation on each feature, scoring with the F1 metric.

\paragraph{TableGAN} \citet{park2018table-GAN} produced TableGAN to accommodate mixed-type data. They also made use of a more sophisticated architecture, with a primary GAN (based on DCGAN \cite{radford2015DCGAN}) and an auxiliary classifier module. They were able to produce good synthetic data in a variety of domains and evaluated privacy leakage using both Euclidean distance to closest records and membership inference attacks. However, they still made no use of privacy-preserving mechanisms to train their model, meaning there were no formal guarantees. Moreover, they accommodated mixed-type data by simply treating categoricals as integers, which is unprincipled as it implies ordering and arithmetic relationships that are not present in categorical data. TableGAN was benchmarked against DCGAN \cite{radford2015DCGAN} on 4 real tabular datasets, with fidelity being evaluated through statistical comparisons of marginals and utility being evaluated with the TSTR compared to TRTR for each feature, using F1 scores and mean relative error for discrete and continuous features respectively.

\paragraph{DPGAN} \citet{xie2018DPGAN} presented DPGAN, one of the first generative models trained under differential privacy constraints. They used a 2D-convolutional WGAN architecture for image synthesis and a simple multilayer WGAN for synthesising binary electronic health records (EHRs). However, they did not accommodate mixed-type data, compare to other approaches, or use an ablation study to evaluate which implementation choices affected their results. Instead, they made use of marginal comparisons (for fidelity) and TSTR comparison with area under the curve (for utility) to evaluate DPGAN performance at varying privacy budgets, demonstrating a quality-privacy trade-off.

\paragraph{PATE-GAN} \citet{jordon2018PATE-GAN} presented a model called PATE-GAN, which refines the Private Aggregation of Teacher Ensembles (PATE) \cite{papernot2016PATE} framework and applies it to GANs. The authors were able to show that their formulation of PATE-based training of GANs can be mapped back to $(\epsilon, \delta)$-DP for consistency and comparison with other approaches. Indeed, they were able to show that using PATE allowed them to outperform DPGAN and a baseline GAN. The approach made use of multiple teacher discriminators trained on disjoint partitions of the data, which are noisily aggregated to train the student discriminator. The PATE-GAN model was evaluated against earlier approaches on 6 datasets of binary and continuous data, 4 of which were electronic health records (EHRs). They evaluated utility by comparing TRTR, TSTR, and TSTS (train on synthetic test on synthetic) and over a range of machine learning algorithms and assessing the relative rankings. Whilst PATE-GAN was performant compared to DCGAN \cite{radford2015DCGAN} and the baseline GAN, determining the optimal number of teachers and the noise level to apply is challenging and requires more iterations than DP-SGD. Moreover, training multiple teacher models imposes a much higher computational cost during training, compared to models trained under DP-SGD, making the technique less helpful in real-world use-cases, where computational resources may be severely constrained.

\newpage
\paragraph{DP-AC-GAN} \citet{beaulieu2019DP-AC-GAN} produced DP-AC-GAN, an auxiliary classifier GAN that made use of 2D convolutions to more efficiently model features that were ordered sequentially. The model was trained under DP-SGD and compared to an unprivatised version on two electronic health record datasets. Fidelity was assessed via marginal comparison using correlation tests and expert clinician review. Utility was assessed via TSTR with AUROC (area under receiver operating curve). Whilst fairly performant on the specific datasets chosen, the approach was highly tailored to the niche of temporally-ordered health records and was not benchmarked against other models.

\paragraph{CTGAN} \citet{xu2019CTGAN} introduced CTGAN, a conditional generative adversarial network explicitly designed for mixed-type tabular data generation. The model first makes use of novel model-based data transformations to better handle non-Gaussian and multimodal distributions. CTGAN uses two-layer, fully-connected networks in both the generator and discriminator to capture the correlations between features. Wasserstein loss with gradient penalties is used to train the GAN with higher stability. The authors made use of packing (as per the PacGAN \cite{lin2018pacgan} framework) to further prevent mode collapse by modifying the discriminator to make decisions based on multiple samples from the same class, either real or artificially generated. In comprehensive benchmarking, the CTGAN model showed superior performance in generating realistic synthetic data in comparison to Bayesian methods, prior GAN-based approaches (TableGAN \cite{park2018table-GAN}, MedGAN \cite{choi2017medGAN}, VeeGAN \cite{srivastava2017veegan}), and a VAE method, TVAE, implemented by the same authors. The authors also used simulated datasets (with known distributions) to assess fidelity in terms of likelihood fitness. On the real datasets, utility was evaluated using the TSTR paradigm, with $R^2$ and F1 metrics for continuous and discrete features respectively. Unfortunately, both the GAN itself and the data transformation steps are difficult to privatise using DP-SGD, as the gradient penalties are incompatible with DP-SGD weight clipping and the preprocessing steps are not gradient based. 

Whilst the model does not implement differential privacy or make any formal privacy guarantees, it represents state-of-the-art performance in mixed-type tabular data synthesis. It is also available as an easy-to-use Python library that automatically infers the correct hyperparameters given the dataset. This makes CTGAN a great benchmark of ideal GAN performance on mixed-type tabular data when no privacy constrains are applied to training. Because CTGAN is used as the main reference model in this work, we discuss how it works in greater detail.

\begin{figure}[ht]
    \centering
    \includegraphics[width=0.90\textwidth]{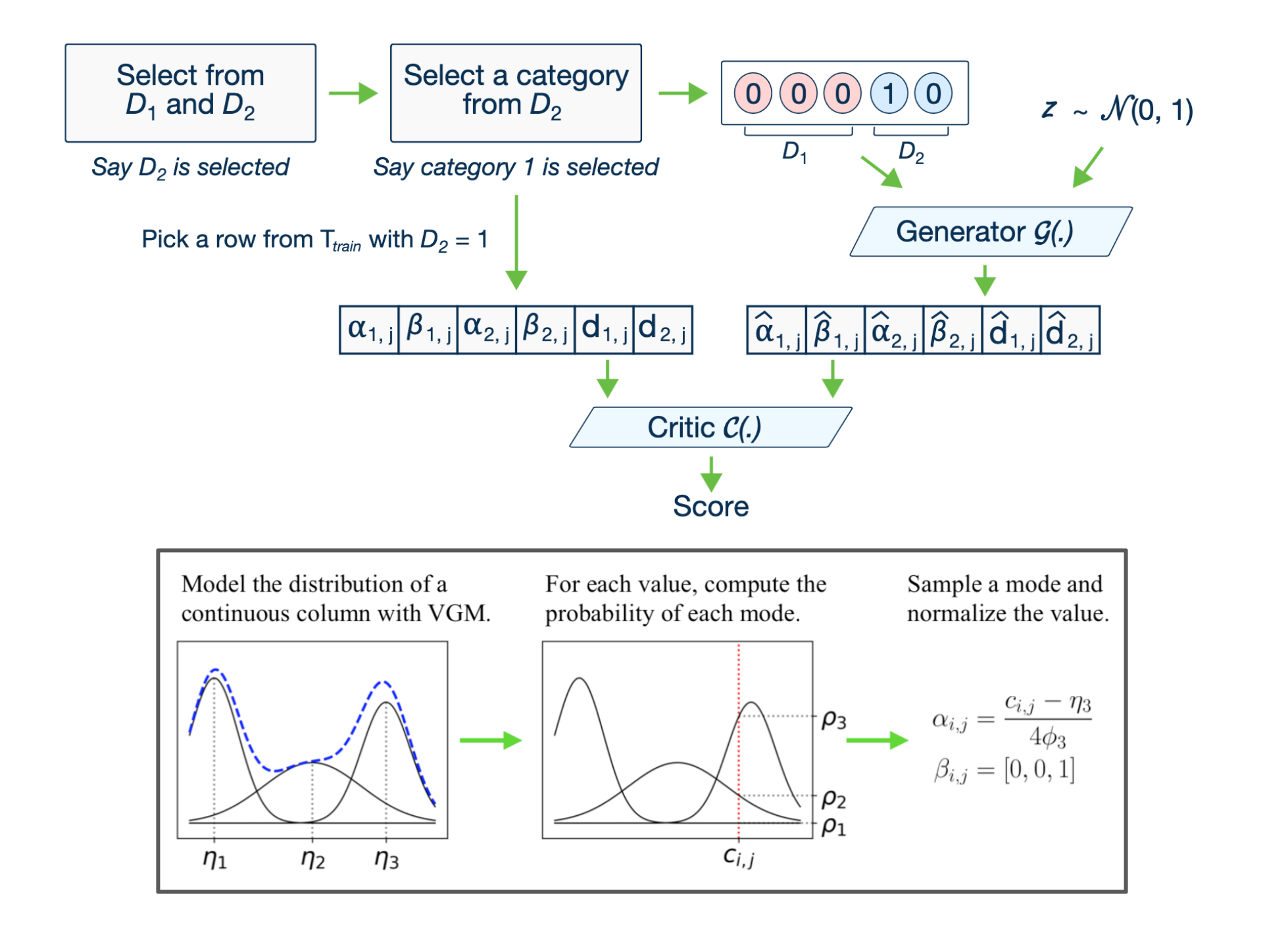}
    \caption{Illustrations of CTGAN conditional sampling (top) and mode-specific normalisation (bottom) from the CTGAN paper by \citet{xu2019CTGAN}.}
    \label{fig:ctgan-illustrated}
\end{figure}

Mode-specific normalisation (Figure \ref{fig:ctgan-illustrated}) works by fitting a variational Gaussian mixture model to each continuous feature to estimate the number of modes ($k$) and represent them with a $k$-way weighted mixture of Gaussians. This is then represented as a concatenation of one-hot vectors (indicating which of the modes) and scalars (representing the normalised values of the mode). This is concatenated with the one-hot representation of the discrete (categorical) features. Next, a conditional GAN is trained with a train-by-sampling approach that better handles imbalanced discrete columns and avoids mode collapse (Figure \ref{fig:ctgan-illustrated}). At each step, a category is selected from one of the discrete features. The one-hot vector indicating this is fed to the generator as the condition along with the standard Gaussian noise $\boldsymbol{z}$. The generator is penalised for producing synthetic examples that do not come from the sampled condition using the mean cross-entropy between the synthetic categories and the sampled conditions in each batch. This forces the generator to adhere to the condition and reduces mode collapse. The discriminator is given real data sampled from the condition and synthetic data from the generator (with the same condition).

\newpage
\paragraph{DP-WGAN} \citet{frigerio2019DP-WGAN} proposed DP-WGAN, which trains a modified Wasserstein GAN under DP-SGD. The authors varied the architecture between an LSTM or an MLP for timeseries or tabular data respectively. They introduced an optimisation called clipping decay that is both compatible with DP-SGD and aims to help stabilise GAN training similarly to gradient penalties. However, the authors did not make special efforts to accommodate arbitrary mixed-typed tabular data, focussing instead on improvements to WGAN training that apply across data modalities. They compared DP-WGAN (with and without clipping decay) to a baseline GAN on three single-type tabular datasets (1 simulated, 2 real), evaluating utility with the TSTR approach measuring classification accuracy and AUROC. They also evaluated privacy with membership inference attacks. Overall, the authors concluded that clipping decay helped stabilise DP-WGAN and resulted in better data utility.

\paragraph{RDP-GAN} \citet{torfi2020RDPCGAN} presented RDP-GAN, one of the first mixed-type tabular GANs trained under Rényi differential privacy using a modified privacy accountant. In recent years, it has become standard practice to use Rényi DP instead of ordinary DP due to the tighter privacy bound estimation. This is then mapped back to $(\epsilon, \delta)$-DP for consistency and comparison. The architecture involved wrapping a WGAN within a pretrained autoencoder, with the aim of having the autoencoder learn a helpful representation of the data that simplifies and stabilises the GAN training. The drawback is that this requires multiple optimisers to share the privacy budget and needs careful hyperparameter tuning for each different dataset. RDP-GAN also made use of one-dimensional convolutions to try capture correlations between adjacent features more effectively. Though it appeared to improve performance, this was likely due to more parameter-efficient training, and the use of convolutions on tabular data makes little sense from first principles. RDP-GAN was compared with DPGAN \cite{xie2018DPGAN}, PATE-GAN \cite{jordon2018PATE-GAN}, and variants of itself (in an ablation study) on 6 real-world mixed-type tabular datasets, using the TSTR paradigm to measure AUROC and AUPRC (area under precision-recall curve). Whilst RDP-GAN outperformed DPGAN and PATE-GAN at all privacy levels, this was most likely due to its exclusive use of Rényi DP, which is far more efficient at composition than the original DP formulation \cite{mironov2017RenyiDP} and is now the recommended approach for all models \cite{opacus}. 

\paragraph{DP-auto-GAN} \citet{tantipongpipat2019DP-auto-GAN} took a similar approach to RDP-GAN by making use of both Rényi DP and by wrapping a Wasserstein GAN in an autoencoder for better representation learning. The authors focused on two electronic health record datasets (one being mixed-type) and compared DP-auto-GAN with DP-WGAN \cite{frigerio2019DP-WGAN}, DP-VAE \cite{acs2018DP-VAE}, and DP-SYN \cite{abay2018DP-SYN}. Fidelity was assessed on marginal distributions, KL-divergence, Jenson-Shannon distance, and PCA latent projections, whilst utility was assessed with AUROC under the TSTR paradigm. Whilst DP-auto-GAN handily outperformed the other models, this is likely exacerbated by its exclusive use of Rényi DP, as in the case of RDP-GAN above.

\chapter{Learning tabular representations with end-to-end attention models}\label{ch:end-to-end}
\section{Motivation}
One of the major limitations of previous works on tabular dataset synthesis is that neural networks struggle to generate heterogeneous, mixed-type, non-Gaussian data found in tables \cite{borisov2022deeptabular, xu2019CTGAN}. This requires a reversible transformation of the tabular datasets into a continuous and homogeneous representation, that is typically implemented with pre- and post- processing. Unfortunately, the majority of techniques rely on independently transforming each feature and thus do not take the join distribution into account. They also require domain knowledge to be hard-coded into the data transformation step. Moreover, the pre-processing of categorical data results in problems of sparsity and varying implied importance depending on the cardinality of the categorical features. For example, a feature like ``country of birth'' can have hundreds of possible values, many of which occur rarely. Conversely, a standard ``sex'' feature is typically binary and often more uniform. In the transformed representation, the ``country of birth'' would result in dozens or hundreds of sparse features for the model to consider, whereas ``sex'' would have only one or two. This mix of sparse and dense features and the explosion in the feature space due to the transformation further exacerbates the challenges of synthesising tabular data. 

One approach to remedy this is to have the model learn its own representation of the features to replace the pre-processing step. This allows the model to learn a dense and proportional representation that takes all features into account simultaneously and best enables the model to capture the underlying joint distribution. Indeed, there has been some success in representation learning for tabular data \cite{borisov2022deeptabular}. However, the problem of synthesising tabular data requires the data transformation to be reversible so that the model can produce synthetic data in the original tabular feature space. Unfortunately, it is much more challenging for neural networks to learn to reverse the representations and there has been almost no research in this direction \cite{borisov2022deeptabular}. We therefore investigated the use of attention-based neural networks for reversible, end-to-end representation learning for tabular data, with the hope of improving tabular synthesis performance.

\section{Background}

\subsection{Attention mechanisms}
Attention encourages the model to focus selectively on disparate segments of the input data when generating an output. It was originally introduced for sequence-to-sequence models, and significantly bolstered the efficacy of machine translation systems by enabling the model to dynamically modulate the emphasis placed on individual input elements based on their relevance to the task. This was further refined with the introduction of the Transformer model \cite{vaswani2017attention}. Transformers deviate from the conventional sequential processing paradigm of prior sequential models like Recurrent Neural Networks (RNNs), and instead opts for a wholly parallelisable methodology known as the self-attention (scaled dot-product) mechanism. 

For each input vector, the self-attention mechanism permits interactions with every other vector in the sequence, irrespective of their relative positions. This interaction is modelled mathematically as:

\begin{equation}
\text{Attention}(\mathbf{Q}, \mathbf{K}, \mathbf{V}) = \text{softmax}\left(\frac{\mathbf{QK}^T}{\sqrt{d_k}}\right) \mathbf{V}
\end{equation}

where \(\mathbf{Q}\) is a matrix of queries, \(\mathbf{K}\) is a matrix of keys, \(\mathbf{V}\) is a matrix of values, and \(d_k\) is the dimensionality of the queries and keys.

Each element in the sequence generates a query \(\mathbf{Q}\), a key \(\mathbf{K}\), and a value \(\mathbf{V}\) by multiplying the input with learned weight matrices. The dot product between the query of one element and the key of another produces a relevance score. The scores are normalised before applying a softmax function to yield a distribution of positive values that sum to 1. The values $\mathbf{V}$ are scaled by this corresponding softmax output. The Transformer model applies this self-attention mechanism $h$ times in parallel across all inputs, culminating in a construct known as multi-head attention.

Since their introduction, attention mechanisms have been applied to most branches of deep learning, dramatically increasing model performance across domains and forming the backbone of state-of-the-art models like GPT-4 and Stable Diffusion.

\subsection{Attention mechanisms for tabular data}
Three notable works to apply attention mechanisms to tabular data are TabNet \cite{arik2021tabnet}, TabTransformer \cite{huang2020tabtransformer}, and SAINT \cite{somepalli2021saint}. Whilst none of these are designed for the generative paradigm, and thus all focus on only the forward representation, they serve as a solid foundation for building reversible tabular representations. TabNet operates much like decision trees, but leverages attention to prioritise salient features at each decision step. TabTransformer, on the other hand, applies self-attention layers to embed each row of the categorical features and then concatenates them with the continuous features.

SAINT (self-attention and inter-sample attention transformer) builds on TabTransformer in two major ways: (1) It projects the continuous and categorical features into the same embedding space, then passes the embeddings through transformer blocks that apply (2) both self-attention (row-wise) and inter-sample attention (column-wise) to produce the final embeddings. In the context of tabular records of individuals, this allows the model to incorporate information across the features of each individual (row) and across all individuals for a feature (column), which affords a better representation of the joint feature distributions across individuals. \citet{somepalli2021saint} show that SAINT's improved representations enable it to outperform XGBoost as well as other neural network approaches on supervised tabular tasks.

\section{Implementation}
Building on the state-of-the-art SAINT architecture, we implemented an end-to-end attention model for synthesising tabular data. This was performed iteratively as a series of modules.

\subsection{Attention-AE: attention-based tabular autoencoder}

First, a tabular encoder module based on the SAINT architecture was used to transform the original $m \times n$ tabular dataset $\mathbf{X}$ into an embedded dataset $T(\mathbf{X})$. We used a 4-head, single-layer transformer with self-attention across the features and inter-sample attention across the samples in each input batch, producing a 32-dimensional embedding.

\begin{figure}[ht]
    \centering
    \includegraphics[width=0.95\textwidth]{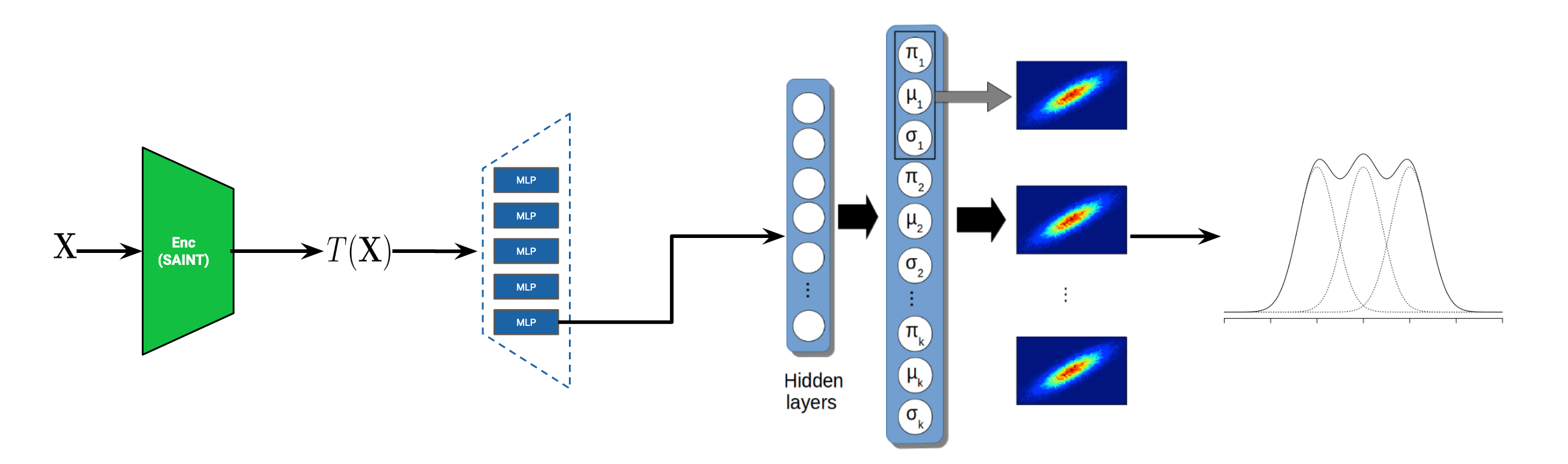}
    \caption{Illustration of the attention-AE with mixture density networks to reconstruct multimodal continuous features.}
    \label{fig:saint-mdn}
\end{figure}

Second, a fully-connected decoder module with $n$ multi-layer perceptrons (MLPs), one for each of the $n$ original features of the dataset, took in the 32-dimensional embedding and reconstructed the tabular dataset $\hat{\mathbf{X}}$. For features that were originally discrete, the MLPs had one output for each class in their feature, with a softmax activation function applied. The argmax of the activation was used to determine which class to reconstruct. For features that were originally continuous, we tried two approaches: using the raw activations, or using mixture-density networks (MDNs). This design is illustrated in Figure \ref{fig:saint-mdn}. 

Mixture Density Networks (MDNs) are a method for modelling complex multi-modal probability distributions with neural networks. Instead of predicting a single output value, MDNs predict the parameters of a mixture model, capturing a probabilistic distribution over possible outputs \cite{bishop2006pattern}. Formally, given an input vector \(\mathbf{x}\) and a target vector $\mathbf{y}$, an MDN models the conditional density function \(p(\mathbf{y}|\mathbf{x})\) as a mixture of \(k\) Gaussian distributions, each characterised by its mean \(\mu_i\), standard deviation \(\sigma_i\), and a mixing coefficient \(\pi_i\), where \(i = 1, \dots, k\):

\begin{equation}
p(\mathbf{y}|\mathbf{x}) = \sum_{i=1}^{k} \pi_i(\mathbf{x}) \mathcal{N}(\mathbf{y}|\boldsymbol{\mu}_i(\mathbf{x}), \sigma_i^2(\mathbf{x}))
\end{equation}

Here, \(\mathcal{N}\) denotes the Gaussian distribution. The coefficients \(\pi_i\) are non-negative and sum to one, ensuring that \(p(\mathbf{y}|\mathbf{x})\) is a valid probability distribution. In the context of reconstructing continuous features for tabular data, MDNs offer the advantage of being able to capture and reproduce the multi-modality of real-world distributions. In our implementation, each of the MDNs had $3 \times k$ nodes in the final layer, predicting the parameter tuples $(\mu_i, \sigma_i, \pi_i)$ that parameterise the $i$-th distribution of a $k$-way Gaussian mixture approximating the original continuous feature, where $\mu$ and $\sigma$ are the mean and standard deviation and $\pi$ is the weighting in the mixture.

Combining the encoder and decoder modules produced a transformer-based autoencoder (attention-AE). The attenion-AE was trained by minimising reconstruction loss, the weighted sum of cross-entropy loss on the categorical features and root mean squared error (RMSE) loss on the continuous features.
\begin{equation}
\mathcal{L}_{\text{recon.}} = \alpha \frac{-1}{c} \sum_{i=1}^{c} x_{i} \log(\hat{x}_{i}) + \beta \sqrt{\frac{1}{d} \sum_{j=1}^{d} (\hat{x}_{j} - x_{j})^2}
\end{equation}

where $c$ is the number of categorical features, $d$ is the number of continuous features, and $\alpha$ and $\beta$ are the weights for the cross-entropy and RMSE losses respectively. These weights can be tuned based on the importance of each feature type, or to balance the magnitudes of the two losses during training.

\subsection{Attention-VAE and Attention-GAN}

\begin{figure}[ht]
    \centering
    \includegraphics[width=0.95\textwidth]{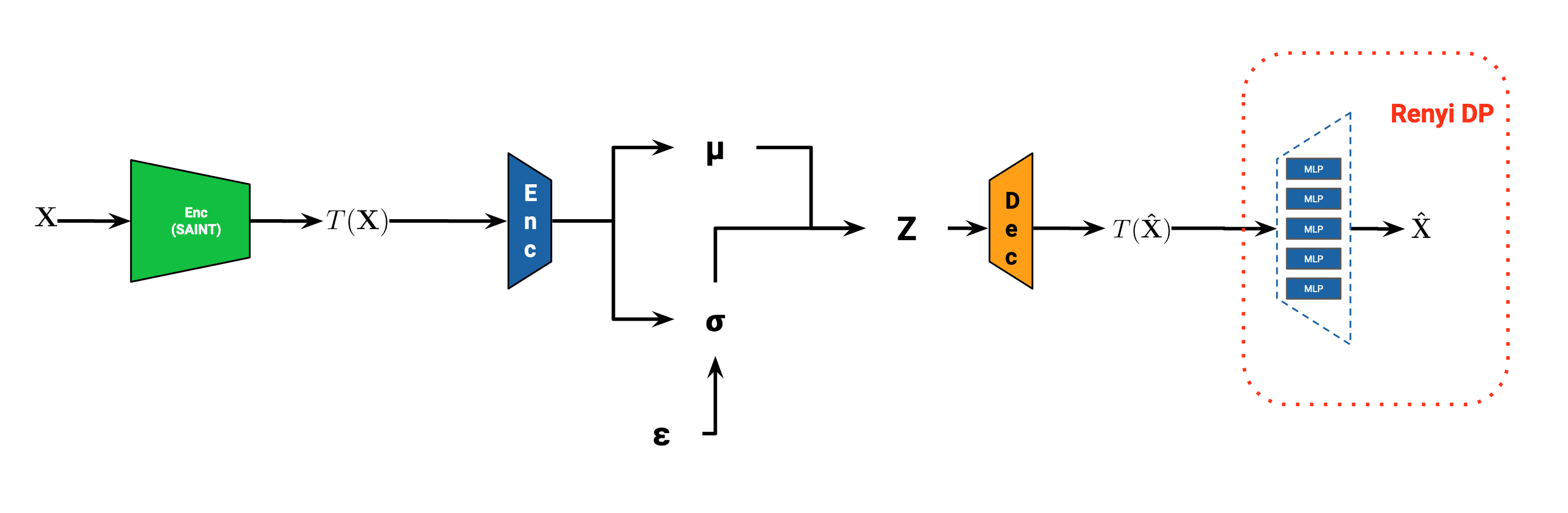}
    \caption{Illustration of attention-VAE implementation.}
    \label{fig:attention-vae}
\end{figure}

Finally, variational autoencoder (VAE) and generative adversarial network (GAN) inner modules were added between the encoder and decoder of the attention-AE to allow for sampling of synthetic data. For the attention-VAE (illustrated in Figure \ref{fig:attention-vae}), the SAINT-based tabular encoder fed the embedded data into a multi-layer encoder that produced the multivariate mean $\boldsymbol{\mu}$ and standard deviation $\boldsymbol{\sigma}$ for a multivariate Gaussian $\boldsymbol{z}$ that was fed to a multi-layer inner decoder, which attempted to reproduce the embedded data. The reconstructed embedding was fed to the $n$ MLPs of the original attention-AE decoder to reproduce the tabular data. 

The attention-VAE was trained to minimise the weighted sum of the reconstruction loss and the regularisation term (computed as the KL-divergence between the $\boldsymbol{z}$ distribution and a standard multivariate Gaussian). For the attention-GAN, a generator module and discriminator module were adversarially trained on the embedded space using Wasserstein loss. This required the outer attention encoder and decoder to first be pre-trained using reconstruction loss and then frozen for training the inner GAN modules.

\newpage
\section{Tuning and evaluation}
First, the hyperparameters of the attention-AE were tuned using Bayesian hyperparameter search in Optuna \cite{akiba2019optuna}. We tuned the embedding dimension, number of attention heads ($h$), MDN mixture size ($k$), loss weighting constants (for the continuous and categorical loss terms), and learning rate, all to minimise reconstruction loss. Next, the attention-VAE and attention-GAN variants were tuned in a similar fashion to find the optimal learning rates and hidden dimension sizes. Finally, the attention-VAE and attention-GAN were repeatedly trained and used to synthesise synthetic versions of two different datasets. The results were compared to those of CTGAN in terms of estimated joint distribution fidelity and marginal distribution reconstruction with a number of performance metrics. To preserve the flow of reading, we present the results in Appendix \ref{ch:appendix-attn}. 

Whilst the attention-AE performed well at embedding and then reconstructing tabular data in an end-to-end fashion (see Figures \ref{fig:exp_220221_121750_results_pmse}, \ref{fig:kaggle_cardio_marginals_AttentionAE_Synthesiser_1.0_1}, and \ref{fig:uci_adult_marginals_AttentionAE_Synthesiser_1.0_3}), the inner VAE and GAN modules were not able to make use of the dense representation to produce good synthetic samples. The Wasserstein GAN was able to converge if the attention-AE was pre-trained, but the results were poor compared to CTGAN, and the VAE variant failed to converge reliably, resulting in poor performance (see Tables \ref{tab:attn_metrics_kaggle_cardio} and \ref{tab:attn_metrics_uci_adult}).

Furthermore, the complexity and scale of the attention-AE (both from the massive number of parameters needed for the attention mechanisms and the large number of MLPs needed to reconstruct the original tabular features) made the approach untenable for the differentially-private version of training. The rapid depletion of the privacy budget and the added instability of training with DP-SGD massively affected the quality of the synthetic data and varying the privacy levels required intricate rebalancing of the hyperparameters for each dataset, which is too impractical for real-world use. Whilst we made some initial progress towards end-to-end tabular representation learning, there remains much work to be done to fully realise the potential of this approach. We hope future work can build on our findings.

\chapter{Novel diffusion models for tabular data}\label{ch:tabular-DMs}
We propose a differentially-private diffusion model, \textit{TableDiffusion}, specifically designed for tabular data, which we believe to be the first of its kind\footnote{Although there is some recent work applying DP-SGD to diffusion models for \textit{image} synthesis \cite{dockhorn2022differentially}, we believe our TableDiffusion model to be the first differentially-private diffusion model for \textit{tabular} data. We are aware of only one other attempt to apply diffusion models to tabular data synthesis, in a very recent paper \cite{kotelnikov2023tabddpm}. However, the authors did not incorporate differential privacy guarantees and only implemented a direct denoising model. Our TableDiffusion model is implemented in two variants -- one denoising, and one that directly predicts noise, which bypasses the challenges of reconstructing tabular data -- and is trained under DP-SGD, making it more relevant for real world applications.}. The training process operates under differentially-private stochastic gradient descent (DP-SGD), providing provable privacy guarantees, whilst allowing arbitrary quantities of data to be synthesised from the same distribution as the training data.

\section{Motivation}

As we have seen, generative adversarial networks (GANs) generate data through a competitive framework involving a generator and a discriminator network. They are known to be challenging to optimise due to their instability and are susceptible to mode collapse \cite{kodali2017convergence}, especially during differentially private stochastic gradient descent (DP-SGD) training \cite{xu2019CTGAN, frigerio2019DP-WGAN, torfi2020RDPCGAN, tantipongpipat2019DP-auto-GAN}. 

Diffusion models (DMs), on the other hand, are an emerging paradigm in generative modelling. DMs are a family of probabilistic generative models that progressively destruct data by injecting noise, then learn to reverse this process for sample generation. They have come to dominate image synthesis applications and are recognised for their high quality, sample diversity, and robust training objective \cite{croitoru2023diffusion, yang2022diffusion}. 

Unlike traditional generative models (GANs and VAEs) that learn the sampling function end-to-end, DMs define the sampling function through a sequential denoising process, which divides the task into many smaller and simpler steps, resulting in smoother and more stable training and simpler neural networks. But diffusion models remain unexplored in the domain of tabular data \cite{yang2022diffusion}.

\newpage
Notably, DMs can be implemented to predict the added Gaussian noise instead of directly denoising data. Outputting standard Gaussian noise is much easier for neural networks than outputting sparse, heterogeneous data with varying non-Gaussian distributions \cite{borisov2022deeptabular}. This, along with the training robustness and parameter efficiency of diffusion models, makes them an ideal candidate for differentially-private tabular dataset synthesis.

\section{Background}

Current research on diffusion models is mostly based on three predominant formulations \cite{yang2022diffusion}: denoising diffusion probabilistic models (DDPMs), score-based generative models, and stochastic differential equations. It has been shown that the training objectives of DDPMs and score-based DMs are mathematically equivalent, and DMs based on stochastic differential equations are a generalisation of the other formulations to the case of infinite time steps or noise levels \cite{yang2022diffusion}.

We base our tabular diffusion models on the widely-validated and flexible formulation of DDPMs \cite{ho2020denoising, sohl2015deep}, which is based on two Markov chains: a forward chain that perturbs data to noise, and a reverse chain that converts noise back to data. We begin with a simplified overview of how such DMs work to provide strong intuitions for the reader.

\paragraph{Forward process} In the forward process, we start with a batch of real data $\mathbf{x}$ and iteratively add increasing amounts of noise according to a scheduler. The simplest version of this would be sampling standard Gaussian noise $\boldsymbol{\xi} \sim \mathcal{N}(\boldsymbol{0}, \boldsymbol{I})$ and scaling it by $\sqrt{\beta_t}$ according to a linear schedule $\beta_t = \frac{t}{T}$, where $T$ is the total number of time steps and $t$ is the current step. In this case, the forward diffusion process can be expressed as:

\begin{equation}
\mathbf{x}_{t+1} = \mathbf{x}_{t} + \boldsymbol{z}_t
\end{equation}

where $\mathbf{x}_t$ is the batch of data at time step $t$ and $\boldsymbol{z}_t \sim \mathcal{N}(\boldsymbol{0}, \boldsymbol{I})\sqrt{\beta_t}$ is scaled Gaussian noise.

\paragraph{Reverse process} In the reverse process, we aim to denoise the data by going from the final time step $\mathbf{x}_T$, which is pure noise, back to the initial step $\mathbf{x}_0$, which approximates the real batch of data $\mathbf{x}$. We achieve this by training the parameters $\boldsymbol{\theta}$ of a neural network $M_{\boldsymbol{\theta}}$. But we have three choices for designing the model's objective function: predict the original data from the noised data (denoising), predict the added noise from the noised data (noise prediction), or predicting some score function of the noised data \cite{luo2022understanding}. In our work, we implement both denoising and noise-predicting DMs. The latter is more instructive, as it is unique to the diffusion paradigm. The model learns to predict the noise $\boldsymbol{z}_t$ added at each step:

\begin{equation}
\boldsymbol{z}_t \approx M_{\boldsymbol{\theta}}(\mathbf{x}_t)
\end{equation}

Given a prediction for $\boldsymbol{z}_t$, we can compute a prediction for $\mathbf{x}_{t-1}$:

\begin{equation}
\mathbf{x}_{t-1} \gets \mathbf{x}_t - M_{\boldsymbol{\theta}}(\mathbf{x}_t)
\end{equation}

By training the neural network in this way, we essentially learn to reverse the diffusion process and recover the original data from the noisy data. After training, the neural network can be used to generate new samples by applying the reverse process to a sample of pure noise.

This kind of diffusion model can be thought of as a hierarchical Markovian VAE, but with a fixed encoder. Specifically, the forward process functions as the unparameterised encoder (a linear Gaussian model). The reverse process corresponds to the parameterised decoder, which is shared across multiple decoding steps. The latent variables within the decoder are all the same size as the sample data. And optimising the diffusion model can be seen as training an infinitely-deep hierarchical VAE \cite{yang2022diffusion}.

\section{Applying DMs to tabular data}
The majority of work on diffusion models has been done in the context of image data \cite{yang2022diffusion}. In those contexts, autoencoders are often used to learn a compressed latent representation of the images and diffusion is performed in the latent space. In applying the diffusion paradigm to sensitive tabular data, we depart from prior work in a few notable ways. 

\begin{figure}[H]
    \centering
    \begin{subfigure}[b]{0.99\textwidth}
        \centering
        \includegraphics[width=\textwidth]{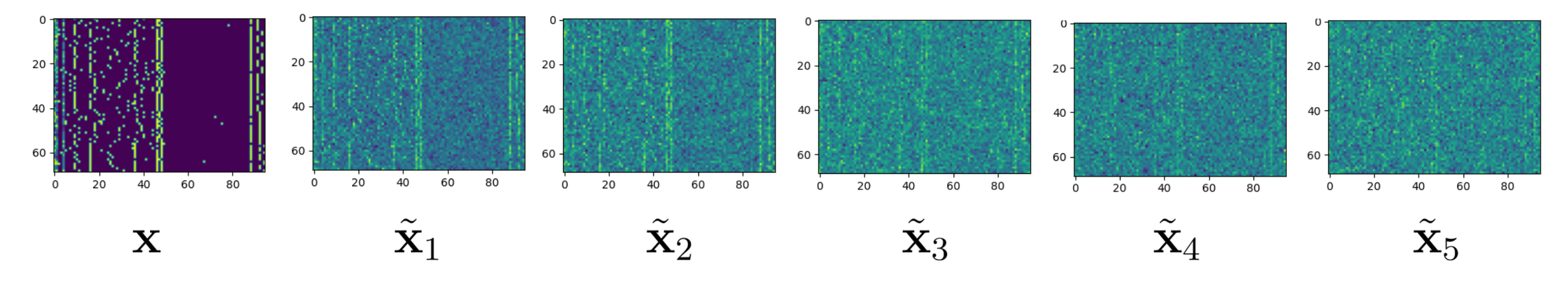}
        \caption{Steps for forward diffusion process for a batch of tabular data.}
        \label{fig:diffusion_forward}
    \end{subfigure}
    
    \begin{subfigure}[b]{0.99\textwidth}
        \centering
        \includegraphics[width=\textwidth]{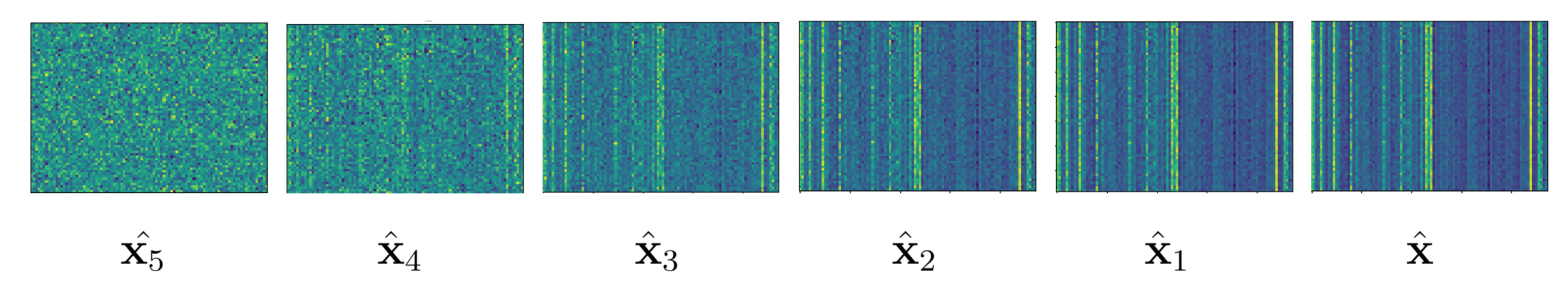}
        \caption{Steps for reverse diffusion process (sampling from model) for a batch of tabular data.}
        \label{fig:diffusion_reverse}
    \end{subfigure}
    
    \caption{Forward and reverse diffusion processes for a batch of tabular data. Each image shows a heatmap of the values in a $b \times n$ batch of data, where $b$ is the (variable) batch size and $n$ is the number of features in the transformed representation. Each row thus represents an individual and each column a latent feature. Note the sparsity of the latent features (most are zero) due to the one-hot encoding of categoricals.}
    \label{fig:diffusion}
\end{figure}

Firstly, instead of compressing an image to a latent space using an encoder network, we use a custom preprocessing approach \footnote{In future work, it would be preferable to mimic the image modality and use an end-to-end autoencoder to learn a robust representation of the tabular data. We explored this approach in Chapter \ref{ch:end-to-end}, but found that considerable work is required to make this approach viable under differential privacy constraints. For the purposes of this work, we opt to use the more conventional pre- and post- processing approach to transform mixed-type tabular data into a latent space for the diffusion model.} to convert the mixed-type tabular data into a homogeneous vector representation and perform diffusion in that space. We can visualise the forward (Figure \ref{fig:diffusion_forward}) and reverse (Figure \ref{fig:diffusion_reverse}) processes in this space, noting the sparsity of features.

\newpage
Secondly, we simplify both diffusion processes by using only a few diffusion steps and by omitting the addition of noise between diffusion steps in the reverse process. These modifications adapt the diffusion paradigm to the lower-dimensional manifolds of tabular data and performed better in our initial experiments, compared to the standard diffusion approaches for the image modality \cite{ho2020denoising}. Many aspects of standard diffusion implementations are unnecessary when training under DP-SGD, as the gradient clipping and noising serves to regularise the model and prevent overfitting \cite{abadi2016momentsaccounting}.

The cost of sampling increases proportionally with the number of diffusion steps, so many researchers have focused on developing approximations and schemes to speed the sampling process up in the compute-heavy modality of images \cite{yang2022diffusion}. But tabular data is lower dimensional and our diffusion models are much smaller and more efficient (because they are optimised for the privacy-preserving context), so full sampling is fast. This allows us to simplify away many of these sophisticated techniques for our context.

The noise at each diffusion step $t$ is scaled by a variable $\beta_t$, which we draw from a cosine schedule:

\begin{equation}
    \beta_t = \frac{1 - \cos\left(\frac{\pi t}{T}\right)}{2},
\end{equation}
where $T$ is the total number of diffusion steps and $t \in \{1, 2, \ldots, T\}$ is the current step. The noise schedule, through the scaling variable $\beta_t$, determines the magnitude of noise that is added at each step. At each step $t$, the noised data $\tilde{\mathbf{x}}_t$ is generated by adding Gaussian noise $\boldsymbol{\xi} \sim \mathcal{N}(\boldsymbol{0}, \boldsymbol{I})$ to the real data $\mathbf{x}$. The noise is scaled by $\beta_t$ as determined by the noise schedule:

\begin{equation}
    \tilde{\mathbf{x}}_t = \mathbf{x} + \sqrt{\beta_t} \cdot \boldsymbol{\xi},
\end{equation}

Compared to earlier diffusion model work by \citet{ho2020denoising} that used a linear noise schedule, \citet{nichol2021improved} demonstrated that adding noise more gradually through a cosine schedule reduced the number of forward process steps $T$ needed by an order of magnitude, whilst achieving similar sample quality.  

We implemented two variants of the model, one which predicts the added noise and one which predicts the denoised data. This allowed for an ablation analysis of which diffusion properties affect synthetic data fidelity.

\newpage
\section{Noise predictor variant}
\label{subsubsec:noise_predictor}
The noise predictor variant (\texttt{TableDiffusion}) is trained to estimate the noise added to the data at step $t$ of the diffusion process. The loss function for this model is defined as the Mean Squared Error (MSE) between the predicted noise $\hat{\boldsymbol{z}}_t$ and the actual noise $\boldsymbol{z}_t$.

For a preprocessed dataset $\mathbf{X}$ with $N$ data points, and a neural network M parameterised by weights $\boldsymbol{\theta}$, we can describe the $(\epsilon_{\text{target}}, \delta)$-DP training process with target batch size $B$ as follows:
\begin{algorithm}[H]
\caption{Training noise-predicting tabular diffusion model under DP}\label{alg:training-diffusion-model}
\begin{algorithmic}[1]
\Procedure{Train}{$\mathbf{X}$, $\boldsymbol{\theta}$, $T$, epochs, $\epsilon_{\text{target}}$}
    \For{epoch in epochs} \Comment{Iterate through each epoch}
        \For{batch $\mathbf{x}_b \sim \text{Poisson}(|\mathbf{X}|, B/N)$} \Comment{Sample mini-batch from dataset}
            \For{step $t \in \{1, \ldots, T\}$} \Comment{Iterate through diffusion steps}
                \State Get $(\epsilon, \delta)$ spent from RDP accountant
                \Comment{Check privacy budget}
                \If{$\epsilon \geq \epsilon_{\text{target}}$} 
                    \Return $\text{M}_{\boldsymbol{\theta}}$
                \EndIf
                \State $\beta_t \gets \frac{1}{2}(1 - \cos\frac{\pi t}{T})$ \Comment{Compute scale parameter from schedule}
                \State $\boldsymbol{z}_t \sim \mathcal{N}(\boldsymbol{0}, \boldsymbol{I})\sqrt{\beta_t}$ \Comment{Sample and scale Gaussian noise}
                \State $\tilde{\mathbf{x}}_{b,t} \gets \mathbf{x}_b + \boldsymbol{z}_t$ \Comment{Add scaled noise to data}
                \State $\hat{\boldsymbol{z}}_t \gets \text{M}_{\boldsymbol{\theta}}(\tilde{\mathbf{x}}_{b,t})$ \Comment{Model predicts noise}
                \State $\mathcal{L}_t \gets \left\| \hat{\boldsymbol{z}}_{t} - \boldsymbol{z}_t \right\|^2$ \Comment{Compute step loss with MSE}
                \EndFor
            \State $\mathcal{L} \gets \frac{1}{T}\sum_{t \in T}{\mathcal{L}_t}$ \Comment{Average loss over diffusion steps}
            \State $\tilde{\nabla}_{\boldsymbol{\theta}} \mathcal{L} \gets \operatorname{CLIP}\left(\nabla_\theta \mathcal{L}, C\right) + \mathcal{N}\left(\boldsymbol{0}, C^2 \sigma^2 \boldsymbol{I}\right)$ \Comment{DP-SGD: Clip and noise gradients}
            \State $\boldsymbol{\theta} \gets \boldsymbol{\theta} - \eta \cdot \text{Adam}(\tilde{\nabla}_{\boldsymbol{\theta}} \mathcal{L})$ \Comment{Update model weights}
        \EndFor
    \EndFor
    \Return $\text{M}_{\boldsymbol{\theta}}$ \Comment{Return trained model}
\EndProcedure
\end{algorithmic}
\end{algorithm}

We can see that this algorithm is broadly similar to training a supervised neural network. We observe the same nested loops over epochs and batches of data (lines 2-3). Instead of sampling fixed-sized batches of data, we use Poisson sampling to get batches approximately $B$ in size (line 3). This sampling-based technique consumes the privacy budget slower than the deterministic batching commonly seen in gradient descent \cite{opacus}. In the diffusion model, we also have an additional inner loop (line 4) over the steps $t \in \{1, \ldots, T\}$ in the forward diffusion process. This means that each batch of data is used $T$ times, augmented by varying levels of added noise. Unlike GAN-based and VAE-based models, this effectively gives the diffusion model $T$ times more data to learn from, without affecting the privacy budget. For each diffusion step $t$, the data is noised, the model predicts the noise, and the loss between the predicted and true noise is calculated (lines 9-11). At the end of the diffusion loop, the losses for each step are aggregated (line 12). This is the point at which DP-SGD is implemented, clipping the gradients of the aggregated loss and adding further noise (line 13). Finally, the privatised gradients are used to update the model parameters with the Adam optimiser (line 14). This is repeated for each batch sample and each epoch, until either the epoch limit (line 2) or the target privacy value $\epsilon_{\text{target}}$ is reached (line 6).

\newpage
Once trained, we can sample synthetic data $\hat{\mathbf{x}}$ by sampling pure noise from a standard multivariate Gaussian of the same shape as the transformed data, then iteratively feeding it through the model in the reverse diffusion process, subtracting the (scaled) predicted noise at each step $t$.

\begin{algorithm}[H]
\caption{Sampling synthetic data from noise-predicting tabular diffusion model}
\begin{algorithmic}[1]
\Procedure{Synthesise}{$\text{M}_{\boldsymbol{\theta}}$, $T$}
    \State $\hat{\mathbf{x}}_{T} \sim \mathcal{N}(\boldsymbol{0}, \boldsymbol{I})$ \Comment{Initialise process from Gaussian noise}
    \For{$t = T$ \textbf{downto} $1$} \Comment{Reverse diffusion process}
        \State $\beta_t \gets \frac{1}{2}(1 - \cos\frac{\pi t}{T})$ \Comment{Compute scale parameter from schedule}
        \State $\hat{\boldsymbol{z}}_t \gets \text{M}_{\boldsymbol{\theta}}(\hat{\mathbf{x}}_{t})$ \Comment{Predict noise}
        \State $\hat{\mathbf{x}}_{t-1} \gets \hat{\mathbf{x}}_{t} - \hat{\boldsymbol{z}}_t \sqrt{\beta_t}$ \Comment{Estimate previous diffusion step}
    \EndFor
    \Return $\hat{\mathbf{x}}_{0}$ \Comment{Return synthetic data}
\EndProcedure
\end{algorithmic}
\end{algorithm}

\section{Mixed-type denoiser variant}
\label{subsubsec:mixed_denoiser}
The mixed-type denoiser variant (\texttt{TableDiffusionDenoiser}) is trained to reconstruct the original data $\mathbf{x}$ from the noised data $\tilde{\mathbf{x}}$. The loss function for this model is a combination of Mean Squared Error (MSE) for numerical features and normalised Kullback-Leibler (KL) divergence for categorical features:

\begin{equation}
    \mathcal{L}_{\text{denoising}} = \text{MSE}(\mathbf{\hat{x}}_t, \mathbf{x}) + \text{KLD}_{\text{norm.}}(\mathbf{\hat{x}}_t, \mathbf{x}) 
\end{equation}

where $\text{MSE}$ is the Mean Squared Error loss for numerical features, calculated similarly to the MSE formula in the diffusion noise predictor model and $\text{KLD}_{\text{norm.}}$ is the average Kullback-Leibler divergence loss over all categorical features, defined as:
\begin{equation}
     \text{KLD}_{\text{norm.}}(\mathbf{\hat{x}}_t, \mathbf{x}) = \frac{1}{K} \sum_{c \in C} D_{\text{KL}} (\hat{P}_c || P_c)
\end{equation}

where $K=\sum_{c \in C}{|c|}$ is the total number of categories across all categorical features, $\hat{P}_c$ is the probability distribution predicted by the model for the categorical feature $c$, and $P_c$ is the true probability distribution of $c$.

Once trained, we can sample synthetic data $\mathbf{\hat{x}}$ by performing a reverse diffusion process similarly to the noise predictor variant. However, in this case, the model estimates the \textit{original} data at each step (denoising) instead of the noise itself:

\begin{equation}
\mathbf{\hat{x}}_{t-1} \gets \text{M}_{\boldsymbol{\theta}}(\mathbf{\hat{x}}_{t}),
\end{equation}
where $\text{M}_{\boldsymbol{\theta}}$ is the trained denoiser model.

\chapter{Benchmarking GANs and diffusion models}\label{ch:benchmarking}

This section describes the experimental setup for benchmarking state-of-the-art models against our novel tabular diffusion models and evaluating the synthetic dataset fidelity using a robust suite of metrics and visualisations.

\section{Implementing prior work}

For a comprehensive comparison, we implemented three leading GAN-based models from the literature (see Section \ref{sec:related-gans}), namely PATE-GAN \cite{jordon2018PATE-GAN}, DP-WGAN \cite{frigerio2019DP-WGAN}, and DP-auto-GAN \cite{tantipongpipat2019DP-auto-GAN}, based on the methodologies described by the authors and any source code they made available. PATE-GAN uses a variant of the PATE mechanism \cite{papernot2016PATE} instead of DP-SGD for privacy guarantees and has been shown to outperform older GAN implementations like DPGAN \cite{jordon2018PATE-GAN}. DP-WGAN is a good baseline GAN method, with only the enhancement of Wasserstein loss\footnote{Although the DP-WGAN authors found that clipping decay improved performance in their study, we omit this technique from the implementation used in our benchmarking, as we used a fixed clipping constant ($C=1$) and privacy engine for all models (except PATE-GAN) to make for a fair comparison of the overarching generative paradigms.}. DP-auto-GAN wraps a Wasserstein GAN in an autoencoder for better latent space construction and is known to outperform VAE-based approaches \cite{tantipongpipat2019DP-auto-GAN}. 

For a baseline of state-of-the-art \textit{unprivatised} performance (i.e. no differential privacy constraints), we used the official CTGAN \cite{xu2019CTGAN} library. CTGAN is one of the state-of-the-art approaches for synthesising tabular data without privacy guarantees. It makes use of extensive model-based preprocessing and conditional sampling to enhance the fidelity and diversity of the synthesised data, but is unsuitable for sensitive data and is difficult to privatise. We thus used CTGAN as a best-case baseline of performance against which to evaluate the privatised models. With this collection of techniques, we have a good variety of exemplary and popular models with a range of different privacy schemes, GAN modifications, and architectural augmentations, allowing for a nuanced and representative comparison to our novel tabular diffusion models. 

\newpage
The goal of this work was to evaluate different high-level approaches to differentially-private tabular data synthesis. To this end, the models used the same pre-processing methods, privacy accountant, and optimiser where possible, with some exceptions due to the properties of the models. In the original works, the models all base the underlying network on similar WGAN architectures. We therefore make use of shared neural architectures for the closest comparison between the approaches in our benchmark, using hyperparameter optimisation to ensure all models had their optimal setup.

\section{Model architectures}
We make use of the same multilayer residual networks used in the implementation of the state-of-the-art CTGAN \cite{xu2019CTGAN} model to implement the core of each benchmark model, as well as for our novel tabular diffusion models. This architecture has been shown to perform exceptionally well on mixed-type tabular datasets.

\paragraph{Generator}
The Generator module was the core of all models, as it was used by all the GAN-based methods and as the sole network in both variants of the diffusion model. It consists of two 128-dimensional residual layers and a final linear layer to scale to the target dimension. Each residual layer consists of a fully-connected layer followed by a GroupNorm and ReLU activation. Group normalisation was used instead of batch normalisation for better privacy preservation \cite{opacus}. The output of each residual block was concatenated with the input to the residual block. 

\paragraph{Discriminator}
The Discriminator module of the GANs was built upon a three-layer feedforward neural network with $0.2$-LeakyReLU activation functions and $p=0.5$ dropout layers for regularisation. The output passes through a sigmoid activation function, producing a value between 0 and 1, indicating the probability of the input data being real. LeakyReLU was chosen for its ability to handle the vanishing gradient problem, while dropout helps to prevent overfitting.

\paragraph{Encoder and Decoder}
The architecture for the wrapper autoencoder was inspired by the high-performing TVAE model from \citet{xu2019CTGAN}, the same authors as CTGAN. The Encoder was specifically used in the DP-auto-GAN model for encoding the real data into a latent space representation to be used by the GAN. The Encoder takes the preprocessed data as input and passes it through several fully-connected layers with ReLU activations. Like the Encoder, the Decoder was also used exclusively in the DP-auto-GAN model. It takes the latent space representation as input and outputs the reconstructed data. Similar to the Encoder, it consisted of fully-connected layers with ReLU activations.

\section{Datasets} \label{sec:datasets}
For ease of comparison with previous works, we use two popular open-source mixed-type tabular datasets in our experiments. Both are known to be challenging instances of real-world, mixed-type tabular data that exhibit all the challenges of the modality (see Section \ref{sec:mixed-type-tab-data}).

\textbf{UCI Adult} \cite{misc_adult_2} is derived from the 1994 U.S. Census database and includes information from approximately 48,000 working adults. It comprises nine categorical and five continuous features. The dataset has been used in numerous related studies \cite{tantipongpipat2019DP-auto-GAN, chen2018DP-AuGM_DP-VaeGM, xu2019CTGAN, frigerio2019DP-WGAN} and is renowned for its complexity, with several challenges that make it an invaluable benchmark. These challenges include highly non-Gaussian continuous distributions and a combination of high-and-low cardinality categorical features exhibiting non-uniform distributions. It also reflects a real-world scenario, given its Census origin, which makes it a suitable dataset for assessing the capability of models in handling mixed-type data in the real world. Similar to previous work \cite{tantipongpipat2019DP-auto-GAN, chen2018DP-AuGM_DP-VaeGM}, we exclude the `education' and `fnlwgt' features to maintain consistency in preprocessing steps.

\textbf{Kaggle Cardio} is an open-source dataset pertaining to cardiovascular disease patients, available on Kaggle and featured in previous works in this domain \cite{torfi2020RDPCGAN}. It contains 70,000 records and 12 features, with 5 being continuous and 7 being categorical (5 of which are binary). The continuous features are a mix of near-Gaussian and extreme power-law distributions. We selected this dataset because of its highly multi-modal nature, with dozens of disjoint clusters of individuals that prove challenging for generative models to capture without mode collapsing \cite{torfi2020thesis}.

\section{Data preprocessing}
Neural networks operate best on normally-distributed floating-point numbers, making preprocessing of skewed, mixed-type tabular data essential. Similar to \citet{tantipongpipat2019DP-auto-GAN}, we preprocessed each feature separately, using the Scikit-learn \cite{scikit-learn} preprocessing module, with \texttt{LabelBinarizer} for categorical features and \texttt{MinMaxScaler} for continuous features. The same processing objects were used to reverse transform the synthesised data back into the same tabular features. Hardcoded rounding and clipping was used to ensure values could be cast back to their original data-types. All models shared the same pre- and post- processing to allow for direct comparison of the results. In the case of the CTGAN benchmark and the two variants of the diffusion model, the raw tabular data (as a dataframe object) was passed into the models in lieu of preprocessed data. This was because these models use bespoke preprocessing to actively handle continuous and categorical features instead of operating on a fully-continuous transformed space. 

\section{Hyperparameter tuning}
Our implementations of PATE-GAN, DP-auto-GAN, DP-WGAN, and both diffusion model variants (TableDiffusion and TableDiffusionDenoiser) were subjected to hyperparameter tuning using the Optuna library \cite{akiba2019optuna} to ensure that each model was in an optimal (but generalisable) configuration for the mixed-type datasets used in the benchmarking. 

Each model was given 50 Optuna trials to find the best hyperparameter combinations. Each trial trained two instances of the model, one for each dataset, for 5 epochs and a privacy budget of $\epsilon=5.0$, $\delta=10^{-5}$. The models were then evaluated in terms of marginal distance and AUPRC (for $\alpha$-precision and $\beta$-recall) metrics (see Section \ref{sec:data-fidelity-metrics}). Each trial thus had 4 score values, which were used in a multi-object optimisation with the NSGA-II algorithm as the sampler. All learning rates (for generators, autoencoders, and discriminators) were tuned on the interval $[10^{-6}, 10^{-1}]$ and batch sizes $B \in [2^6, 2^7, \ldots, 2^{11}]$. Any other essential hyperparameters (such as the number of teachers in PATE-GAN or the pretraining fraction in DP-auto-GAN) were also tuned for their respective models. Wherever authors had provided suggested hyperparameter values in their papers or code, these were used as queued trials in Optuna. For the purposes of benchmarking techniques in this study, we used the entirety of each dataset and did not include the privacy budgets of the hyperparameter optimisation in the final privacy accounting \footnote{In sensitive real-world deployments, the hyperparameter tuning should be done under the same privacy budget as the final model training. Because of the post-processing guarantee of differential privacy (see Section \ref{ssec:properties-of-dp}), the hyperparameter tuning runs can then be safely excluded from the final privacy accounting.}. 

\newpage
The Pareto-optimal hyperparameter configurations from the Optuna trials were selected for each model and used in the final comparison runs. The same hyperparameters were used across datasets and privacy budgets, meaning there was very low risk that the configurations were overfit to the specific datasets. Regardless, all models were tuned with an identical protocol, enabling direct comparison of the underlying techniques used.

\section{Privacy accounting}
Except for PATE-GAN (which uses the PATE \cite{papernot2016PATE} paradigm to enforce differential privacy) and the unprivatised CTGAN baseline, all models used the Opacus library \cite{opacus} to manage privacy accounting. Opacus allows for Poisson sampling of training batches, automatic Rényi-DP-SGD, and automatic conversion between $(\alpha, \epsilon)$-RDP and $(\epsilon, \delta)$-DP bounds when using PyTorch models. Rényi Differential Privacy was used for tighter privacy bound estimates during training and the gradient clipping factor $C$ was set to 1 for all models, as per recommendations \cite{opacus}.

In the case of GANs, only the Discriminator module is trained under differential privacy, because information only flows between the sensitive data and the Generator module via the Discriminator. The post-processing property of differential privacy guarantees that the Generator will match the privacy level of the Discriminator. In the case of DP-auto-GAN, the Decoder module of the autoencoder was also trained under DP-SGD, summing the $\epsilon$-values for the Decoder and Discriminator to get the total $\epsilon$ used by the model, as per the composability property of DP. Similarly to the GAN case, the post-processing property means the Encoder module inherits the privacy guarantees of the Decoder.

\section{Experiment configuration}
All models were trained from initialisation and sampled 10 times (using different, but shared, random seeds for each run) for each dataset and privacy budgets $\epsilon \in [0.5, 1, 5, \ldots, 500, 1000]$, with $\delta=10^{-5}$. Each repeat, we also trained the unprivatised CTGAN baseline for 30 epochs on each dataset and recorded the samples. This resulted in 800 synthetic datasets, 400 for UCI Adult and 400 for Kaggle Cardio, to be evaluated against the real datasets for fidelity and diversity. As the goal of the benchmarking experiments was to compare the techniques, the entirety of each dataset was used for training, with synthetic data being evaluated against the entire real datasets.

\section{Data fidelity metrics}\label{sec:data-fidelity-metrics}
In this section, we detail the evaluation metrics used to assess the fidelity of the generated synthetic data. To quantify \textit{joint}-distribution fidelity, we used probabilistic mean squared error (pMSE) ratio, and the $\alpha$-precision and $\beta$-recall metrics proposed by \citet{alaa2022faithful}. To quantify \textit{marginal}-distribution fidelity, we used two statistical tests aggregated across features to compute an overall distance score. Collectively, these metrics provide a comprehensive evaluation of the fidelity of synthetic data on both marginal and joint distributions.

\newpage
\subsection{Probabilistic mean squared error (pMSE) ratio}

The pMSE ratio was proposed by \citet{snoke2018general} and implemented by \citet{arnold2020really}. It is a model-based measure of joint-distribution fidelity that quantifies the similarity between real and synthetic datasets by training a logistic regression discriminator to distinguish between them. Mathematically, the pMSE ratio is defined as:
\begin{equation}
\text{pMSE} = \frac{\text{Observed Utility}}{\text{Expected Utility}}.
\end{equation}

The Observed Utility is computed as:
\begin{equation}
\text{Observed Utility} = \frac{1}{n_1 + n_2} \sum_{i=1}^{n_1 + n_2} \left( s_i - \frac{n_2}{n_1 + n_2} \right)^2,
\end{equation}
where \( n_1 \) is the size of the original dataset, \( n_2 \) is the size of the synthetic dataset, and \( s_i \) represents the predicted probability for each data point \( i \) belonging to the synthetic dataset, as output by the discriminator. The Expected Utility is given by:
\begin{equation}
\text{Expected Utility} = p \times (1 - p) \times \frac{d}{n_1 + n_2},
\end{equation}
where \( d \) is the number of features in the dataset and \( p = \frac{n_1}{n_1 + n_2} \) is the proportion of real data points.

The intuition behind this metric is that a well-trained discriminator would accurately distinguish between real and synthetic datasets if they are notably different. However, a high-quality synthetic dataset would confound the discriminator, leading to poor performance. Consequently, the closer the pMSE ratio is to zero, the better the synthetic data mirrors the original data in terms of general structure and feature distributions. 

\subsection{One-way marginal distribution metrics}

To assess the fidelity of synthetic data in terms of the marginal distributions, we use two frequentist statistical tests, which we can then aggregate into a single distance score. For discrete features, the similarity between the synthetic and real datasets is measured using the inverted chi-squared test \cite{plackett1983karl}. Let \( f_{\text{obs}} \) and \( f_{\text{exp}} \) denote the observed frequencies in the synthetic dataset and the expected frequencies from the real dataset, respectively, for each categorical value. The test statistic is computed as:
\begin{equation}
\chi^2 = \sum \frac{(f_{\text{obs}} - f_{\text{exp}})^2}{f_{\text{exp}}}.
\end{equation}
The p-value derived from this test statistic provides a measure of the divergence between the datasets. Given that a p-value close to 1 indicates that the datasets likely derive from the same distribution, the similarity measure is taken as \( 1 - \text{p-value} \), ensuring a result within the range of $[0.0, 1.0]$, which can be interpreted as a distance measure.

For continuous features, the Kolmogorov-Smirnov (KS) distance \cite{massey1951kolmogorov} is used. The KS test gauges the maximum difference between the cumulative distribution functions (CDFs) of the two datasets. Formally, if \( F_{\text{real}}(x) \) and \( F_{\text{synth}}(x) \) represent the CDFs of the real and synthetic datasets, respectively, the KS statistic is:
\begin{equation}
D = \max_x |F_{\text{real}}(x) - F_{\text{synth}}(x)|.
\end{equation}
A smaller value of \( D \) indicates greater similarity between the two datasets.

\newpage
Both the inverted chi-squared test and the KS distance are appropriate for one-way marginal distribution comparison due to their sensitivity to changes in location, dispersion, and shape of the distribution. We adapted the implementations of KS distance and inverted-chi-squared from the Synthetic Data Vault Project \cite{SDMetrics}, combining them over the features into a single marginal distance metric by taking the mean. This gives us a distance measure on the interval $[0.0, 1.0]$ that captures the average distance between the real marginal distributions and the synthetic ones, for a given dataset, with scores closer to zero indicating higher-fidelity synthetic data.

\subsection{$\alpha$-precision and $\beta$-recall}
\begin{figure}[ht]
    \centering
    \includegraphics[width=0.95\linewidth]{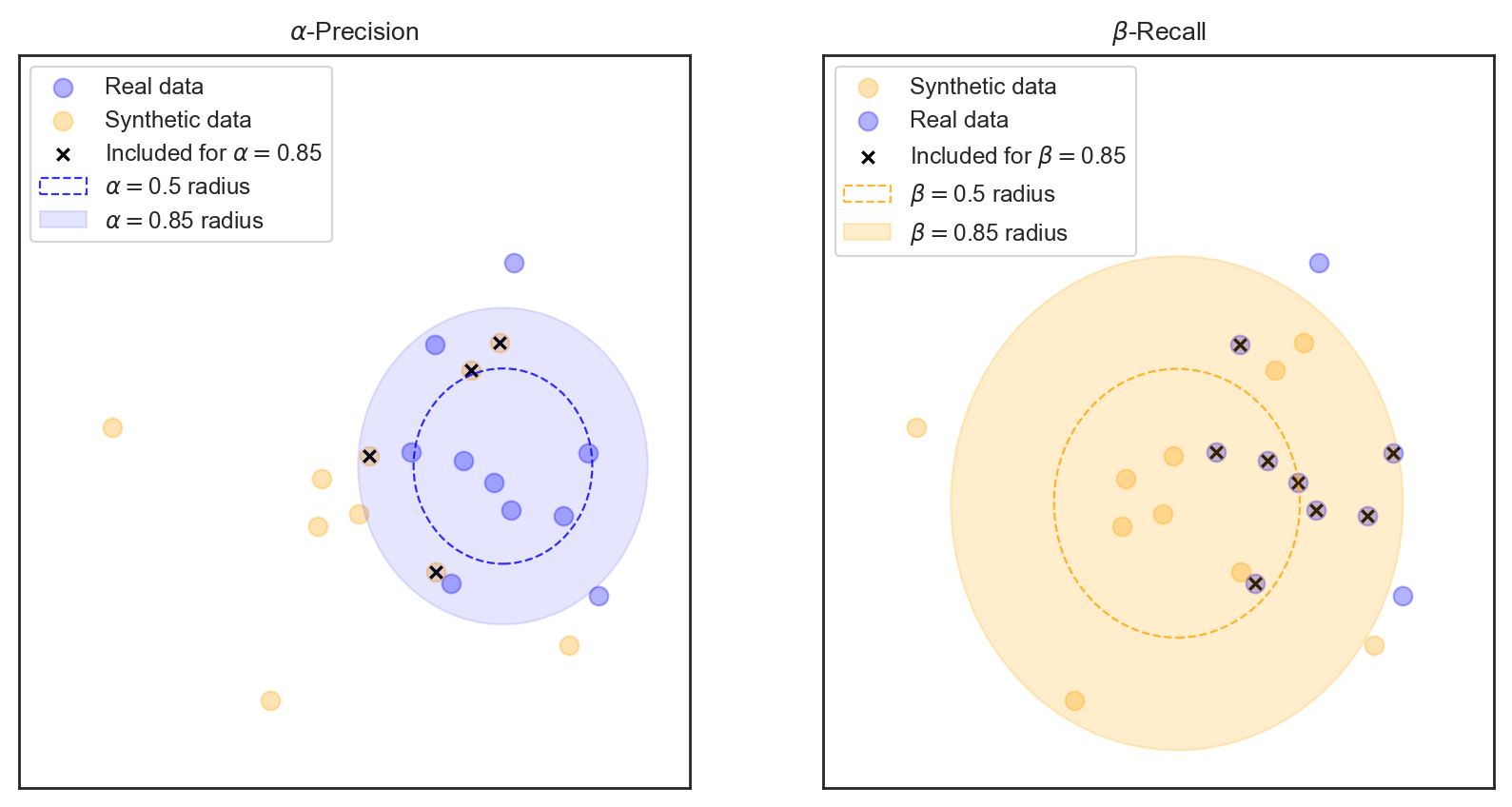}
    \caption{2D illustration of $\alpha$-precision and $\beta$-recall metrics being calculated on toy synthetic data.}
    \label{fig:alpha_beta_illustration}
\end{figure}

To further evaluate the fidelity of the synthetic data, we used the rigorous $\alpha$-Precision and $\beta$-Recall metrics from \citet{alaa2022faithful}. For a visual intuition of these metrics, see Figure \ref{fig:alpha_beta_illustration}. $\alpha$-Precision is the fraction of synthetic samples that resemble the most typical fraction of real samples, whereas $\beta$-Recall is the fraction of real samples covered by the most typical fraction of synthetic samples. The metrics are first calculated on a per-sample basis, then aggregated. 

$\alpha$-Precision ($P_{\alpha}$) measures synthetic data fidelity by quantifying the probability that a synthetic sample falls within the $\alpha$-support of the real distribution. Formally,
\begin{equation}
    P_\alpha \triangleq \mathbb{P}\left(\hat{\mathbf{x}} \in \mathcal{S}_{\text{real}}^\alpha\right), \text { for } \alpha \in[0,1]
\end{equation}
where $\hat{\mathbf{x}}$ is a synthetic data point and $\mathcal{S}_{\text{real}}^\alpha$ is the $\alpha$-support of the real data distribution -- the smallest volume (the most compact set) that has a cumulative probability mass of at least $\alpha$. We can get a visual intuition for this by considering the left panel of Figure \ref{fig:alpha_beta_illustration}, where the synthetic data points that fall within the $(\alpha=0.85)$-support of all the real data points are indicated with an \textbf{x}. 

Intuitively, we can think of the $\alpha$-support as a series of hyperspheres centred around the mean of the real data points, with their size increasing with larger values of $\alpha$. $\alpha$-Precision is computed as the proportion of synthetic points inside the hypersphere with radius $\alpha$. Higher values indicate that synthetic data points are more concentrated around regions where real data points are densely packed, meaning the generated samples resemble real samples, thus indicating high joint distribution fidelity.

$\beta$-Recall ($R_\beta$) measures synthetic data fidelity by assessing the coverage of the real data by the synthetic data. It uses the same concept of hyperspheres as in $\alpha$-Precision but inverts the calculation to quantify the real data points that fall within the $\beta$-support of the synthetic data. Formally,

\begin{equation}
    R_\beta \triangleq \mathbb{P}\left(\mathbf{x} \in \mathcal{S}_{\text{synth}}^\beta\right), \text { for } \beta \in[0,1]
\end{equation}
where $\mathbf{x}$ is a real data point and $\mathcal{S}_{\text{synth}}^\beta$ is the $\beta$-support of the synthetic data distribution -- the smallest volume (the most compact set) that has a cumulative probability mass of at least $\beta$. Again, we can get a visual intuition for this by considering the right panel of Figure \ref{fig:alpha_beta_illustration}, where the real data points that fall within the $(\beta=0.85)$-support of all the synthetic data points are indicated with an \textbf{x}. High $\beta$-Recall for a given $\beta$ implies that the generated samples are diverse enough to cover the variability of the real data, and the model should be able to generate a wide variety of high-fidelity samples.

Aggregating over all the synthetic data points for each $\alpha$ allows us to construct the precision curve, whilst aggregating over all the real data points for each $\beta$ allows us to construct the recall curve. Following the recommendations of \citet{alaa2022faithful}, we take the integral over all $\alpha$ and $\beta$ values to reduce the curves to scalar values that capture the overall realism and diversity of the synthetic values on the interval $[0.0, 1.0]$. Multiplying these together gives as the area under the precision-recall curve (AUPRC), which represents data fidelity in a single metric on the interval $[0.0, 1.0]$. High-fidelity synthetic datasets have AUPRC values closer to 1, whilst low-fidelity synthetic datasets have AUPRC values closer to 0.

\chapter{Results and discussion} \label{ch:results-discussion}

\begin{figure}[ht]
    \centering
    \includegraphics[width=0.99\textwidth]{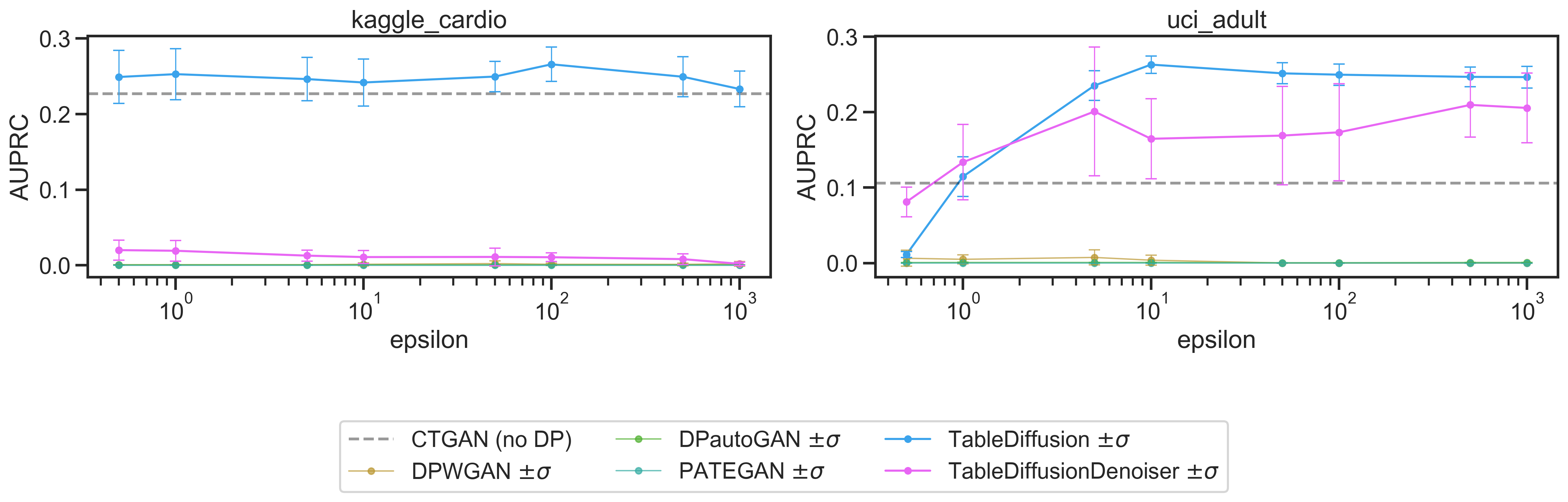}
    \caption{Comparison of area under $\alpha$-precision and $\beta$-recall curves (higher is better) of models across datasets and privacy budgets ($\epsilon \in [0.5, 10^{3}]$, $\delta=10^{-5}$). Points show the mean performance over 10 runs and error bars show $\pm 1\sigma$ of standard deviation. Dashed line shows mean CTGAN baseline score over 10 runs with no privatisation.}
    \label{fig:exp_230530_222958_metrics_2}
\end{figure}

Overall, the diffusion models consistently outperformed the GAN-based methods in terms of fidelity and robustness across datasets and privacy levels (Figure \ref{fig:exp_230530_222958_metrics_2}). The noise-predicting variant, TableDiffusion, excelled and often outperformed the unprivatised CTGAN benchmark, even under the strictest privacy constraints (Tables \ref{tab:metrics_kaggle_cardio_1.0} and \ref{tab:metrics_uci_adult_1.0}). We assessed the loss curves (Figure \ref{fig:loss_curves}) and privacy curves (Figure \ref{fig:privacy_curves}) during model training, finding that the diffusion paradigm results in vastly more stable and privacy-efficient training. The diffusion models also avoided common issues seen in the GAN models, such as mode collapse, and accurately sampled rare classes at the correct frequencies, all contributing to synthetic datasets that replicate the structure of the original data in both its joint and marginal distributions. We confirm this visually by comparing the feature histograms (Figures \ref{fig:marginal_comparison_kaggle_cardio} and \ref{fig:marginal_comparison_uci_adult}) and PCA projections (Figure \ref{fig:projection_heatmaps}) of the synthetic and real datasets, showing nearly identical reconstructions with the TableDiffusion model. The rest of the chapter explores and discusses these results in more detail.

\section{Performance on joint distributions}
\begin{figure}[H]
    \centering
    \includegraphics[width=0.9\textwidth]{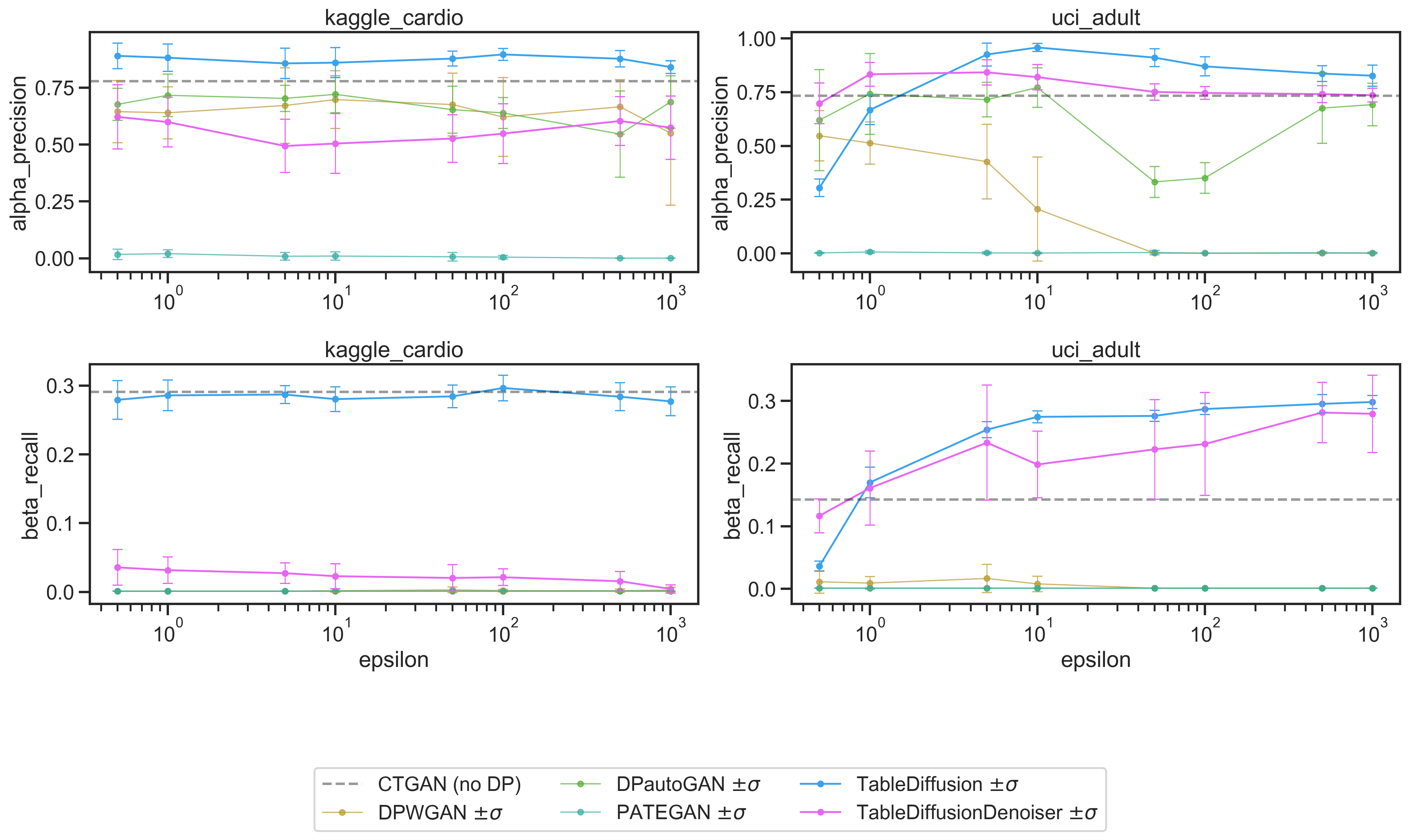}
    \caption{Comparison of $\alpha$-precision and $\beta$-recall scores (higher is better) of models across datasets and privacy budgets ($\epsilon \in [0.5, 10^{3}]$, $\delta=10^{-5}$). Points show the mean performance over 10 runs and error bars show $\pm 1\sigma$ of standard deviation. Dashed line shows mean CTGAN baseline score over 10 runs with no privatisation.}
    \label{fig:exp_230530_222958_metrics_1}
\end{figure}

\begin{figure}[H]
    \centering
    \includegraphics[width=0.9\textwidth]{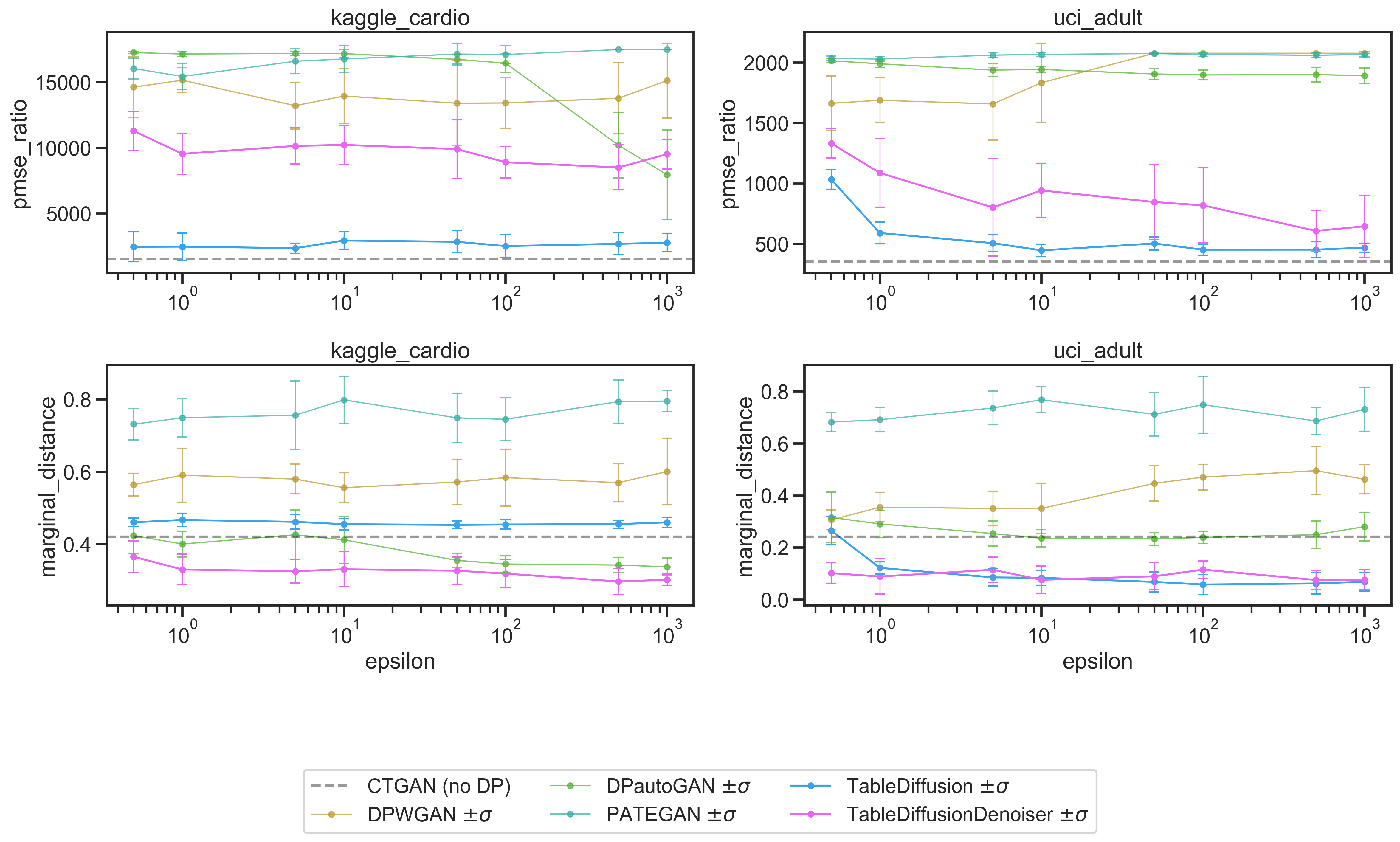}
    \caption{Comparison of pMSE ratio and marginal distance scores (lower is better) of models across datasets and privacy budgets ($\epsilon \in [0.5, 10^{3}]$, $\delta=10^{-5}$). Points show the mean performance over 10 runs and error bars show $\pm 1\sigma$ of standard deviation. Dashed line shows mean CTGAN baseline score over 10 runs with no privatisation.}
    \label{fig:exp_230530_222958_metrics_0}
\end{figure}

Across datasets and privacy levels, the two variants of diffusion model consistently outperformed the GAN-based models on both $\alpha$-precision and $\beta$-recall metrics (Figure \ref{fig:exp_230530_222958_metrics_1}), indicating superior fidelity and diversity of synthetic data. Moreover, the logistic-regression discriminator found the diffusion-synthesised data far more difficult to discriminate from the real data than the GAN-synthesised data (see top of Figure \ref{fig:exp_230530_222958_metrics_0}), indicating a superior reconstruction of the distribution with diffusion models. In particular, the noise-predicting variant was almost always the top performer across different metrics. In most cases, it outperformed even the CTGAN baseline (which had no privacy constraints), even under strict privacy budgets (Tables \ref{tab:metrics_kaggle_cardio_1.0} and \ref{tab:metrics_uci_adult_1.0}). 

\begin{table}[ht]
\centering
\caption{Mean performance (over 10 runs) on Kaggle Cardio under (1.0,$10^{-5}$)-DP. Best results in bold.}
\label{tab:metrics_kaggle_cardio_1.0}
\begin{tabular}{lllllll}
\hline
                 Model & $\epsilon$ &          pMSE &             MD & $\alpha$-Precision & $\beta$-Recall &          AUPRC \\
\hline
                DP-WGAN &          1 &         15164 &          0.591 &              0.639 &          0.001 &          0.001 \\
             DP-auto-GAN &          1 &         17162 &          0.401 &              0.716 &          0.001 &          0.001 \\
               PATE-GAN &          1 &         15448 &          0.749 &              0.021 &          0.001 &          0.000 \\
        TableDiffusion &          1 & \textbf{2461} &          0.467 &     \textbf{0.881} & \textbf{0.286} & \textbf{0.253} \\
TableDiffusionDenoiser &          1 &          9543 & \textbf{0.331} &              0.599 &          0.032 &          0.019 \\
                 
\hline
		CTGAN (benchmark) &         -- &          1518 &          0.421 &              0.778 &          0.291 &          0.227 \\
\hline
\end{tabular}
\end{table}

\begin{table}[ht]
\centering
\caption{Mean performance (over 10 runs) on UCI Adult under (1.0,$10^{-5}$)-DP. Best results in bold.}
\label{tab:metrics_uci_adult_1.0}
\begin{tabular}{lllllll}
\hline
                 Model & $\epsilon$ &         pMSE &             MD & $\alpha$-Precision & $\beta$-Recall &          AUPRC \\
\hline
                DP-WGAN &          1 &         1689 &          0.355 &              0.513 &          0.009 &          0.005 \\
             DP-auto-GAN &          1 &         1991 &          0.291 &              0.742 &          0.001 &          0.001 \\
               PATE-GAN &          1 &         2031 &          0.692 &              0.007 &          0.001 &          0.000 \\
        TableDiffusion &          1 & \textbf{590} &          0.122 &              0.667 & \textbf{0.170} &          0.115 \\
TableDiffusionDenoiser &          1 &         1088 & \textbf{0.089} &     \textbf{0.833} &          0.161 & \textbf{0.134} \\
                 
\hline
		CTGAN (benchmark) &         -- &          353 &          0.242 &              0.733 &          0.143 &          0.106 \\
\hline
\end{tabular}
\end{table}

To assess the significance of the results, we performed the Friedman test to compare the result distributions (for 10 runs) across all the models by one-way repeated measures analysis of variance by ranks. For both datasets, all models had significantly different performance on each metric (Friedman statistic $>35$, $p<10^{-6}$). In post-hoc analysis, the Wilcoxon signed-rank test was used to assess the significance of each privatised GAN model's performance compared with the diffusion models. At $(1.0, 10^{-5})$-DP, the TableDiffusion model significantly outperformed the GAN-based models across all metrics ($p<0.01$), except DP-auto-GAN in the $\alpha$-precision metric on the UCI Adult dataset ($p=0.32$), and TableDiffusionDenoiser outperformed the GAN-based models across all metrics ($p<0.01$), with three exceptions \footnote{DP-WGAN in $\alpha$-precision on Kaggle Cardio ($p=0.38$), DP-auto-GAN in MD on Kaggle Cardio ($p=0.02$), and DP-auto-GAN in $\alpha$-precision on UCI Adult ($p=0.56$).}.

Notably, the diffusion-based models (particularly the noise-predictor) also varied much less than the GAN based models in their performance across training runs. We can see this in the much tighter error bars for TableDiffusion compared to the GAN-based models in Figures \ref{fig:exp_230530_222958_metrics_2}, \ref{fig:exp_230530_222958_metrics_1}, and \ref{fig:exp_230530_222958_metrics_0}. The greater consistency of the diffusion model results (across datasets, privacy-levels, and fidelity metrics) suggests that diffusion models would be more reliable in real-world applications.


\section{Performance on marginal distributions}
Over all marginal distributions, the diffusion models dramatically outperformed most GAN-based models and matched or outperformed even the unprivatised CTGAN baseline (Figure \ref{fig:exp_230530_222958_metrics_0}). The DP-WGAN model also performed quite well in terms of the aggregated marginal scores. Interestingly, both diffusion-based and GAN-based models appeared insensitive to the privacy constraints on marginal distribution performance, showing similar scores at both very low ($\epsilon=0.5$) and very high ($\epsilon=1000$) privacy budgets. 

\begin{figure}[H]
  \centering
  \begin{subfigure}[b]{0.99\textwidth}
    \centering
    \includegraphics[width=\textwidth]{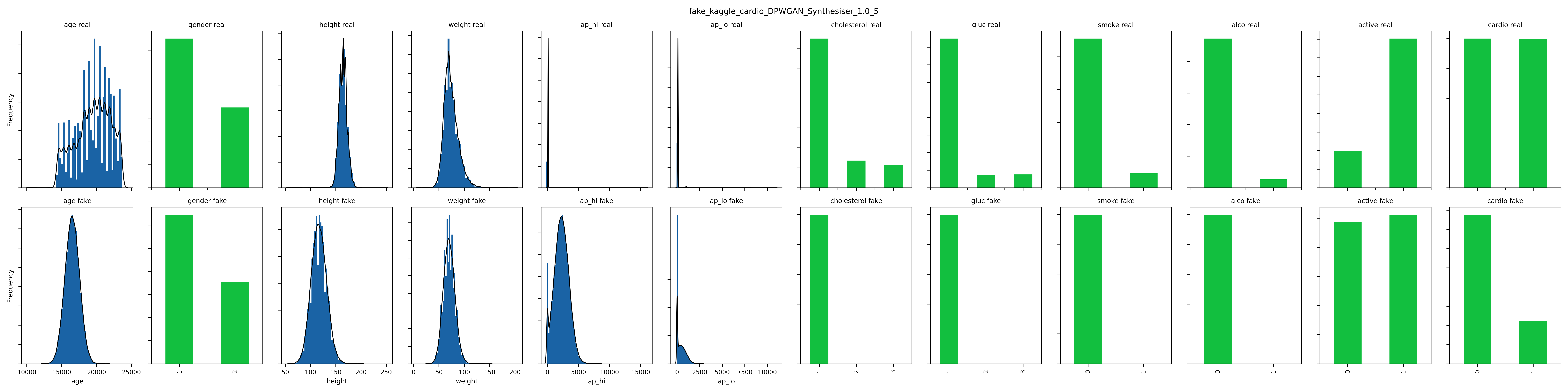}
    \caption{DP-WGAN ($\epsilon=1.0$)}
    \label{fig:kaggle_cardio_marginals_DPWGAN_Synthesiser_1.0_5}
  \end{subfigure}
  \hfill
  \begin{subfigure}[b]{0.99\textwidth}
    \centering
    \includegraphics[width=\textwidth]{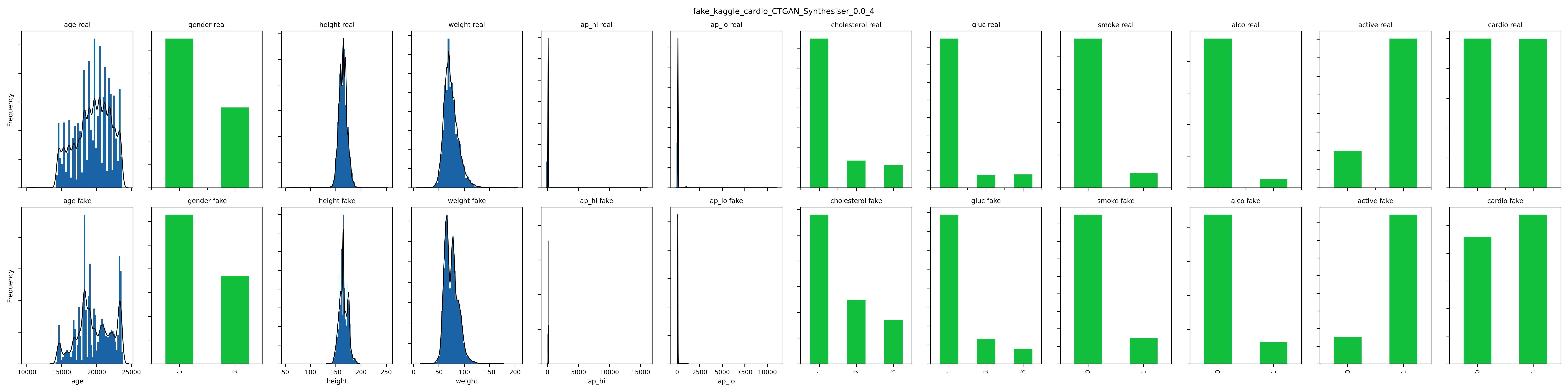}
    \caption{CTGAN benchmark (unprivatised)}
    \label{fig:kaggle_cardio_marginals_CTGAN_Synthesiser_0.0_4}
  \end{subfigure}
  \hfill
  \begin{subfigure}[b]{0.99\textwidth}
    \centering
    \includegraphics[width=\textwidth]{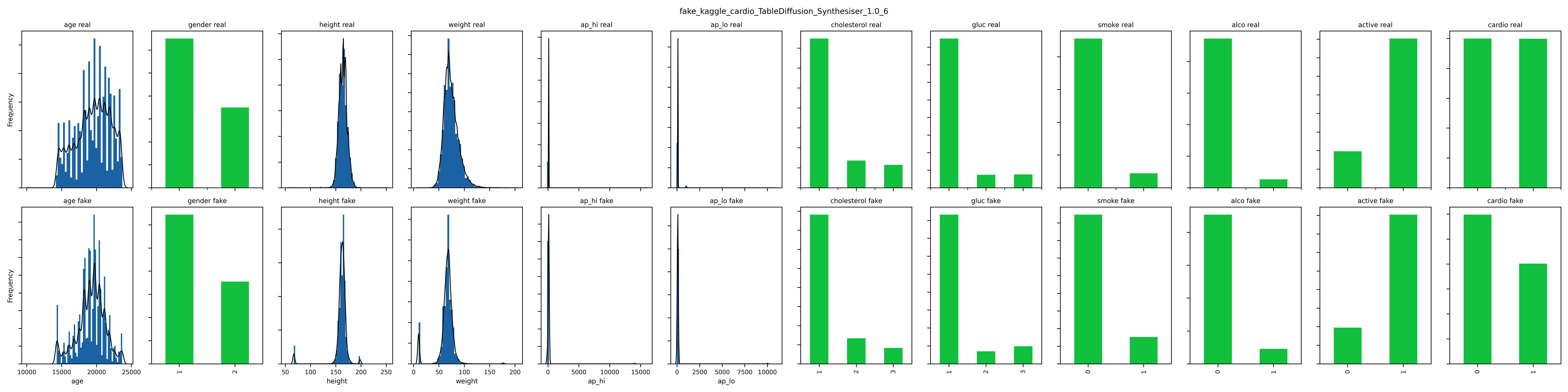}
    \caption{TableDiffusion ($\epsilon=1.0$)}
    \label{fig:kaggle_cardio_marginals_TableDiffusion_Synthesiser_1.0_6}
  \end{subfigure}
  \caption{Comparison of real (top row) and synthetic (bottom row) marginal distributions of DP-WGAN (top), unprivatised CTGAN benchmark (middle), and TableDiffusion (bottom) models on the Kaggle Cardio dataset. Categorical features are coloured green and continuous features are coloured blue, with both bins and overlaid kernel density estimates (black) for clarity. For easy comparison, all subfigures are histograms sharing the same vertical axes ($y \in [0,1]$). Real and synthetic features in the same column share the same horizontal axes.}
  \label{fig:marginal_comparison_kaggle_cardio}
\end{figure}

\begin{figure}[H]
  \centering
  \begin{subfigure}[b]{0.99\textwidth}
    \centering
    \includegraphics[width=\textwidth]{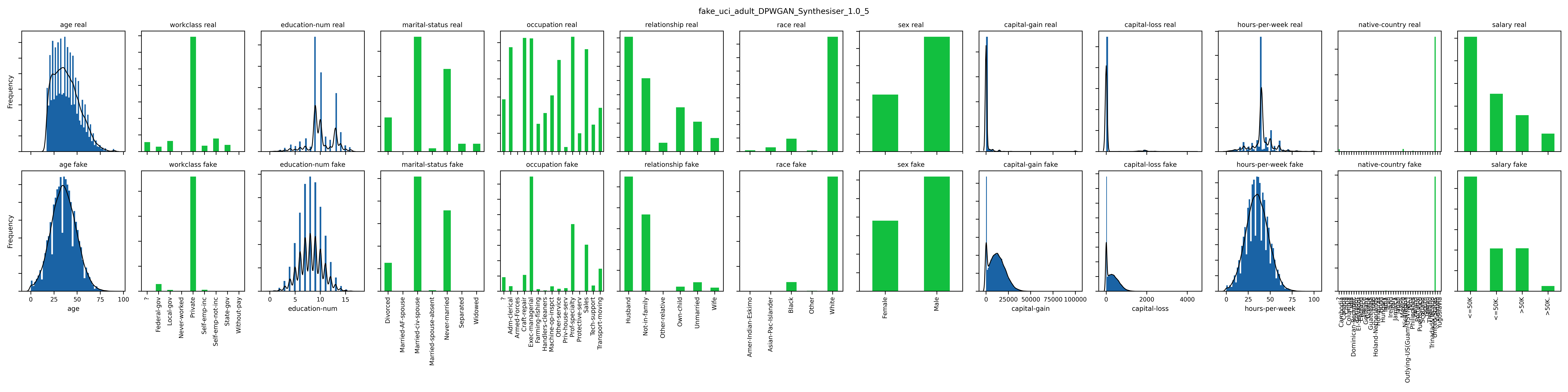}
    \caption{DP-WGAN ($\epsilon=1.0$)}
    \label{fig:uci_adult_marginals_DPWGAN_Synthesiser_1.0_5}
  \end{subfigure}
  \hfill
  \begin{subfigure}[b]{0.99\textwidth}
    \centering
    \includegraphics[width=\textwidth]{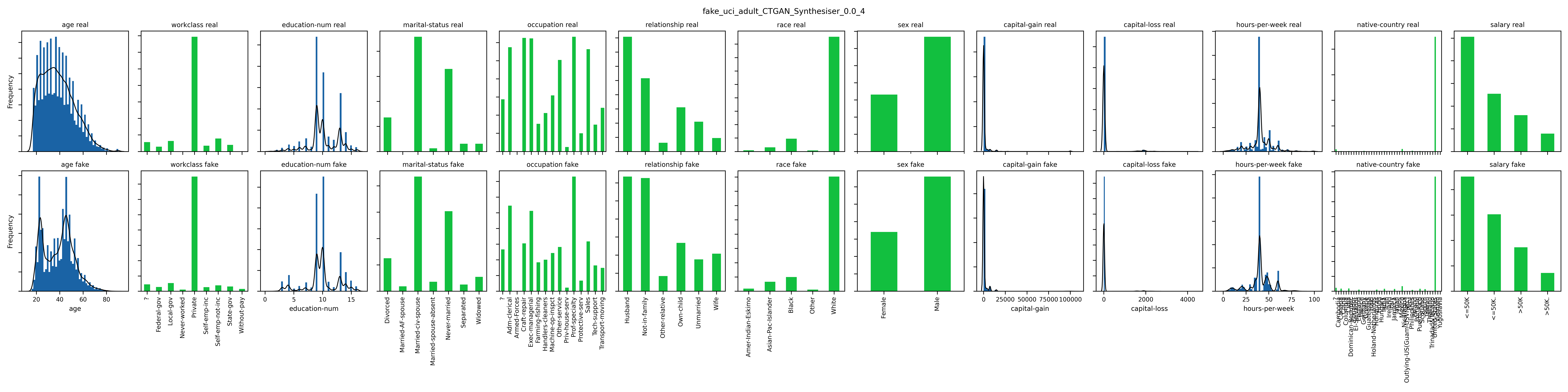}
    \caption{CTGAN benchmark (unprivatised)}
    \label{fig:uci_adult_marginals_CTGAN_Synthesiser_0.0_4}
  \end{subfigure}
  \hfill
  \begin{subfigure}[b]{0.99\textwidth}
    \centering
    \includegraphics[width=\textwidth]{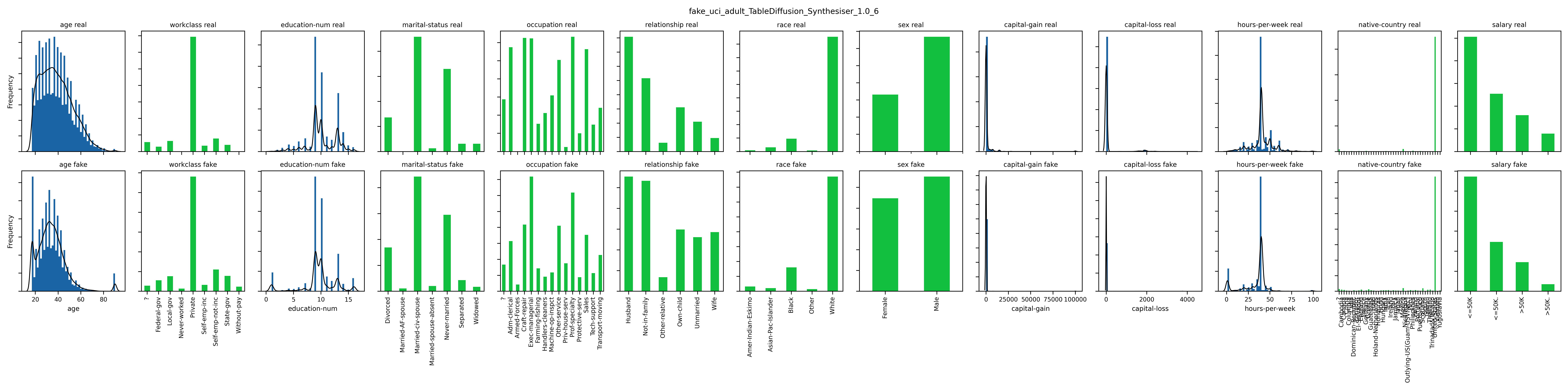}
    \caption{TableDiffusion ($\epsilon=1.0$)}
    \label{fig:uci_adult_marginals_TableDiffusion_Synthesiser_1.0_6}
  \end{subfigure}
  \caption{Comparison of real (top row) and synthetic (bottom row) marginal distributions of DP-WGAN (top), unprivatised CTGAN benchmark (middle), and TableDiffusion (bottom) models on the UCI Adult dataset. Categorical features are coloured green and continuous features are coloured blue, with both bins and overlaid kernel density estimates (black) for clarity. For easy comparison, all subfigures are histograms sharing the same vertical axes ($y \in [0,1]$). Real and synthetic features in the same column share the same horizontal axes.}
  \label{fig:marginal_comparison_uci_adult}
\end{figure}

When examining the individual marginal distributions, we can see that diffusion models produce excellent reproductions of the original data distributions, despite strict privacy budgets (Figures \ref{fig:marginal_comparison_kaggle_cardio} and \ref{fig:marginal_comparison_uci_adult}). 

\newpage
Compared to GAN-based models, like DP-WGAN, the diffusion models appear much less likely to suffer mode collapse or oversampling on categorical features and are much more robust to non-Gaussian distributions for continuous features. Even compared to the unprivatised CTGAN benchmark, which makes extensive use of pre-processing and conditional sampling and was trained for six times the number of epochs, the diffusion models produce comparable or better synthetic distributions in almost all cases. This is remarkable, considering they do not perform conditional sampling or make use of any explicit mode analyses. In the case of categorical features, the diffusion models were not only less prone to mode collapse, but are notably better than CTGAN at sampling rare and non-modal classes at the correct frequencies. 

\newpage
One failure mode of the diffusion models occurs in multi-modal, long-tailed continuous distributions, where unrealistic or unrepresentative outlier values were sometimes invented. This appears to happen in a tiny minority of samples and most commonly found in the denoising variant and at strict privacy budgets. Given the extremity of the values and the shapes of the distributions, it is likely that this was an artefact of either numerical instability in the model output or of the post-processing method used to transform the ranges. It would make sense that the denoising variant was more prone to these issues, as synthesising non-Gaussian distributions is far more challenging for neural networks than predicting Gaussian noise \cite{borisov2022deeptabular}, concurring with the reconstruction issues of end-to-end attention-based models (see Chapter \ref{ch:end-to-end}).

\section{Stability and privacy consumption}
\begin{figure}[H]
    \centering
    \includegraphics[width=0.95\textwidth]{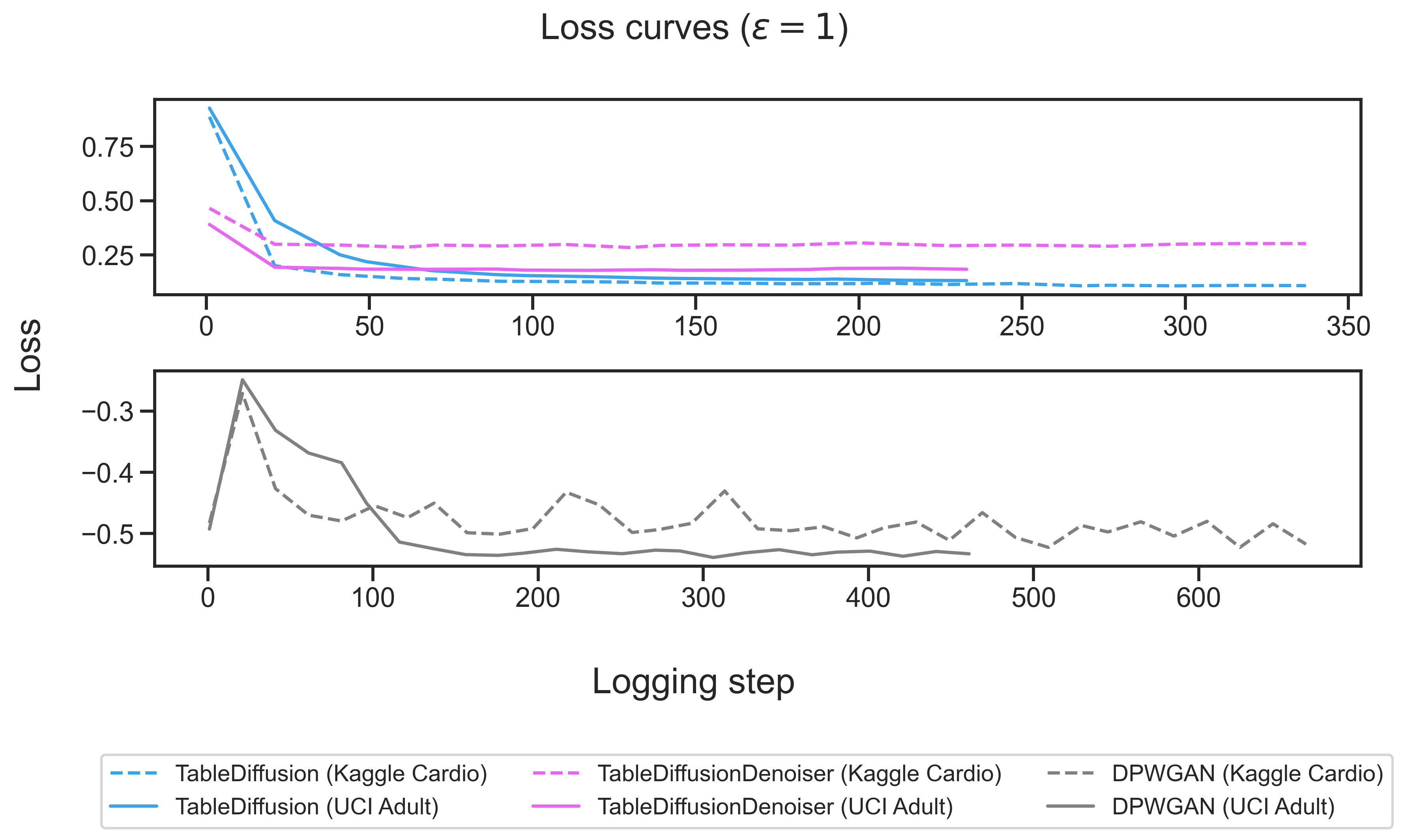}
    \caption{Comparison of the training loss during training of our TableDiffusion models and DP-WGAN (Generator loss) for a target of $\epsilon=1.0$. Note the different scales on the horizontal axes, which show that the diffusion models also trained faster.}
    \label{fig:loss_curves}
\end{figure}

\begin{figure}[H]
    \centering
    \includegraphics[width=0.95\textwidth]{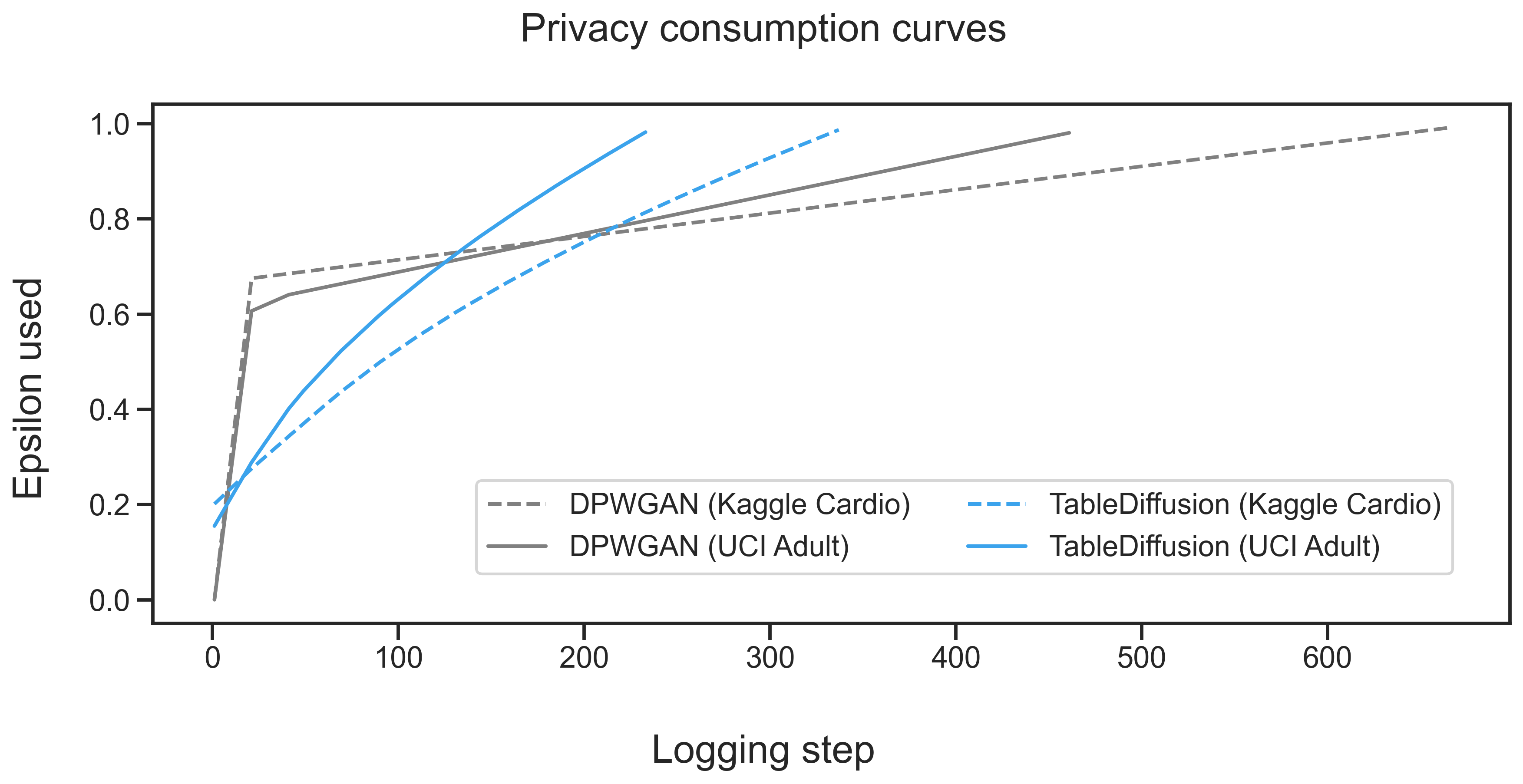}
    \caption{Comparison of the privacy consumption during training of our TableDiffusion model and DP-WGAN for a target of $\epsilon=1.0$. Note: Both variants of the diffusion model have nearly identical curves, which would overlap in the figure, so we omit the denoising variant for visual clarity.}
    \label{fig:privacy_curves}
\end{figure}

Not only were the diffusion models more consistent in their results, they converged faster and more reliably than the GAN-based models, showing smoother loss curves that consistently plateau earlier (Figure \ref{fig:loss_curves}). This more stable and convergent training process, along with making repeated use of each batch of data (with varying augmentation by the diffusion noise), combined to give the diffusion models vastly improved data efficiency and training efficiency. This is illustrated by smoother and more consistent privacy curves (Figure \ref{fig:privacy_curves}).


\section{Visual analysis of 2D projections}
\begin{figure}[H]
\centering

\begin{minipage}[t]{.40\linewidth}
\raggedleft 
\subfloat{\includegraphics[trim=0.3\linewidth{} 0.25\linewidth{} 0.25\linewidth{} 0.12\linewidth{},clip,width=0.6\linewidth]{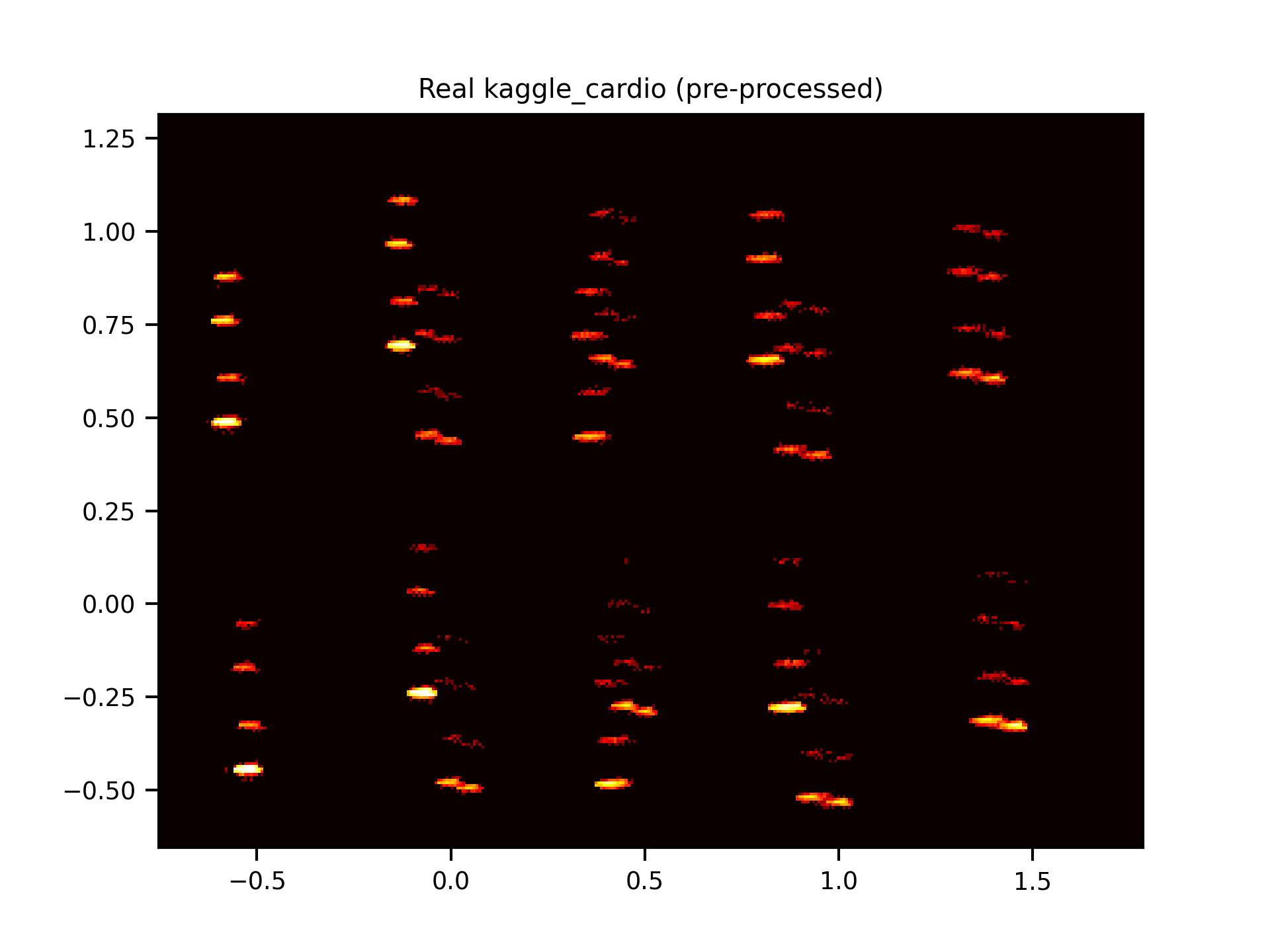}
\label{fig:kaggle_cardio_pca_real}}

\vspace{-1ex}

\subfloat{\includegraphics[trim=0.3\linewidth{} 0.25\linewidth{} 0.25\linewidth{} 0.12\linewidth{},clip,width=0.6\linewidth]{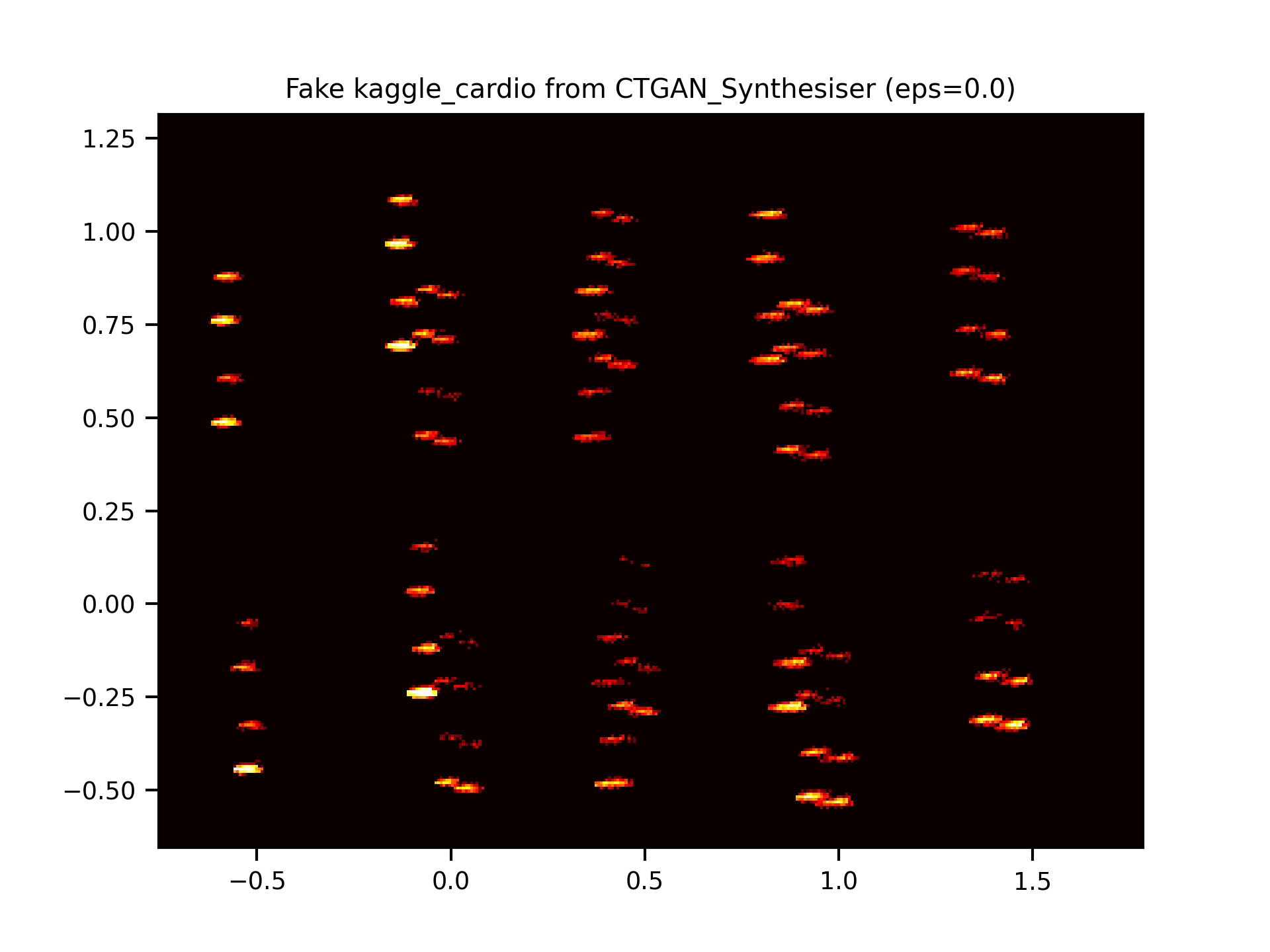}
\label{fig:kaggle_cardio_pca_CTGAN_Synthesiser_0.0_1}}

\vspace{-1ex}

\subfloat{\includegraphics[trim=0.3\linewidth{} 0.25\linewidth{} 0.25\linewidth{} 0.12\linewidth{},clip,width=0.6\linewidth]{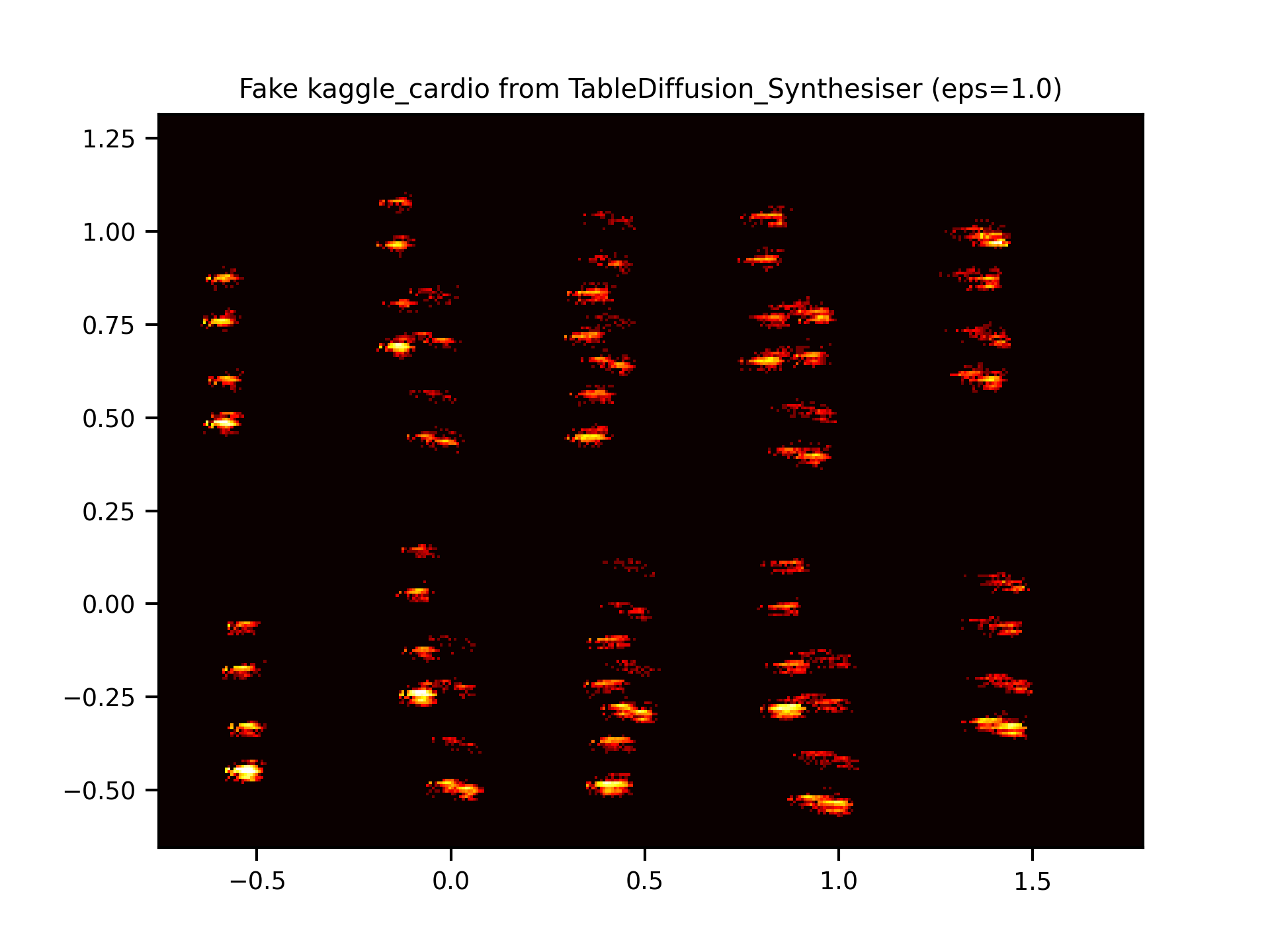}
\label{fig:kaggle_cardio_pca_TableDiffusion_Synthesiser_1.0_1}}

\vspace{-1ex}

\subfloat{\includegraphics[trim=0.3\linewidth{} 0.25\linewidth{} 0.25\linewidth{} 0.12\linewidth{},clip,width=0.6\linewidth]{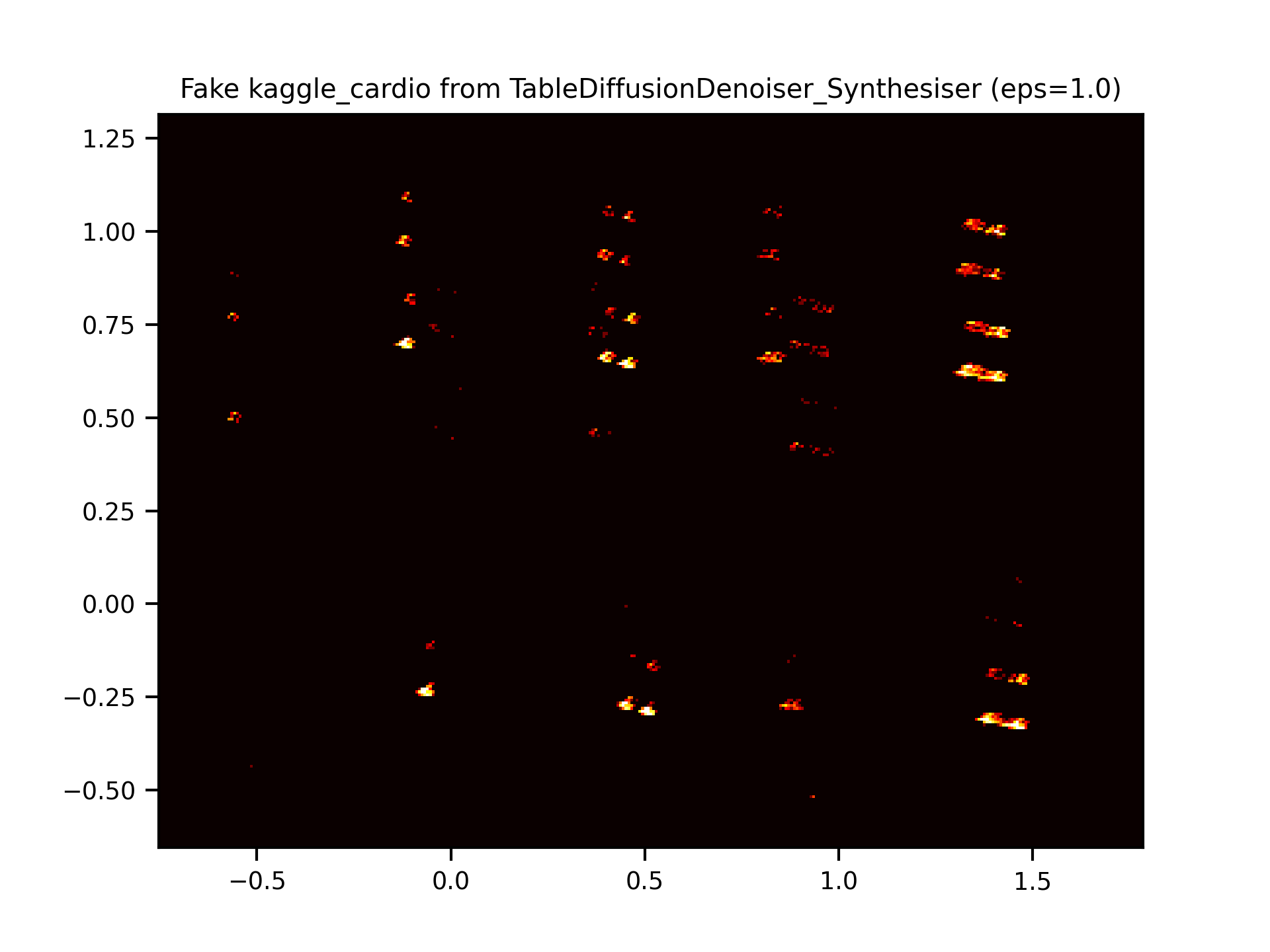}
\label{fig:kaggle_cardio_pca_TableDiffusionDenoiser_Synthesiser_1.0_1}}

\vspace{-1ex}

\subfloat{\includegraphics[trim=0.3\linewidth{} 0.25\linewidth{} 0.25\linewidth{} 0.12\linewidth{},clip,width=0.6\linewidth]{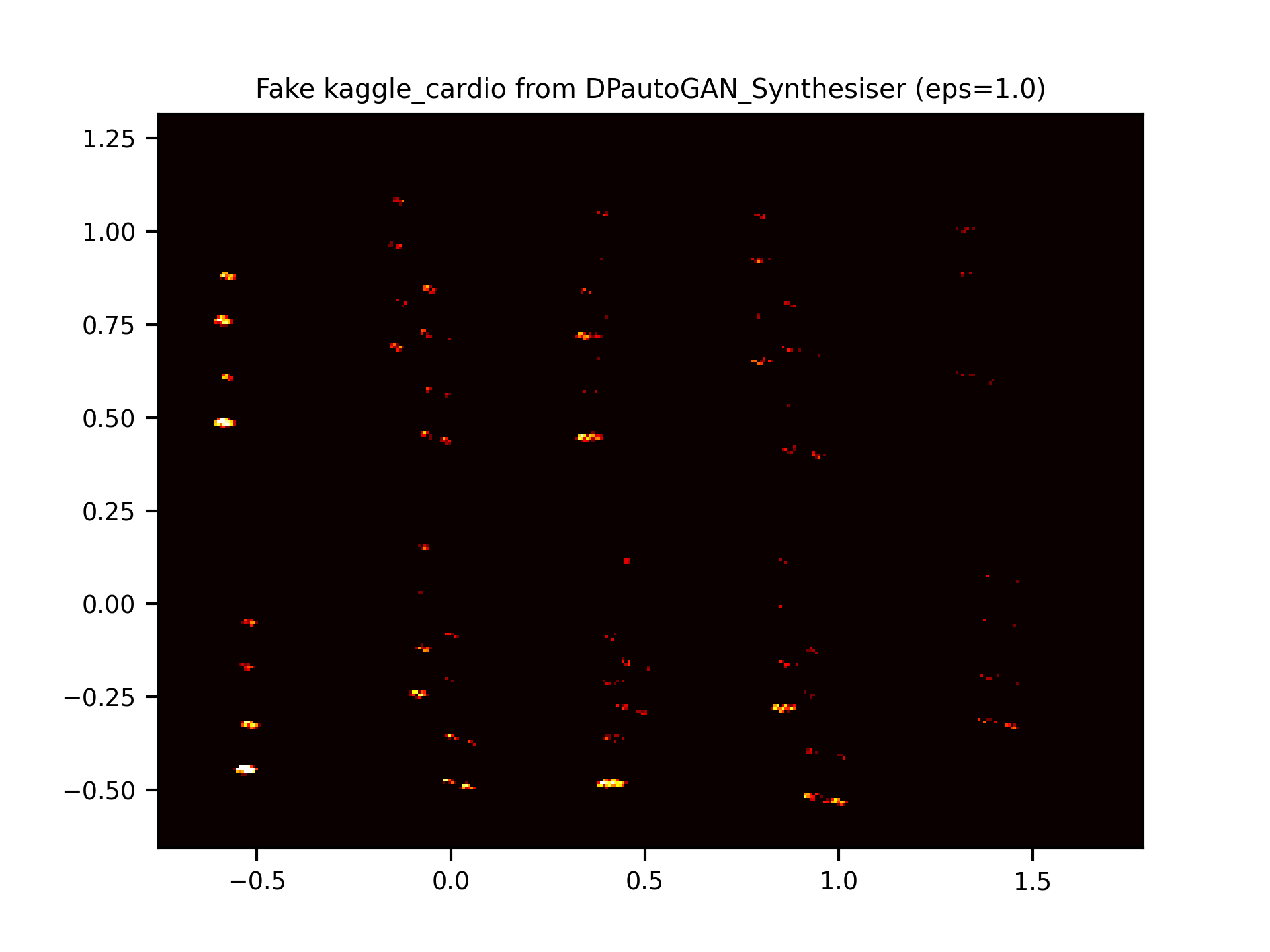}
\label{fig:kaggle_cardio_pca_DPautoGAN_Synthesiser_1.0_1}}

\vspace{-1ex}

\subfloat{\includegraphics[trim=0.3\linewidth{} 0.25\linewidth{} 0.25\linewidth{} 0.12\linewidth{},clip,width=0.6\linewidth]{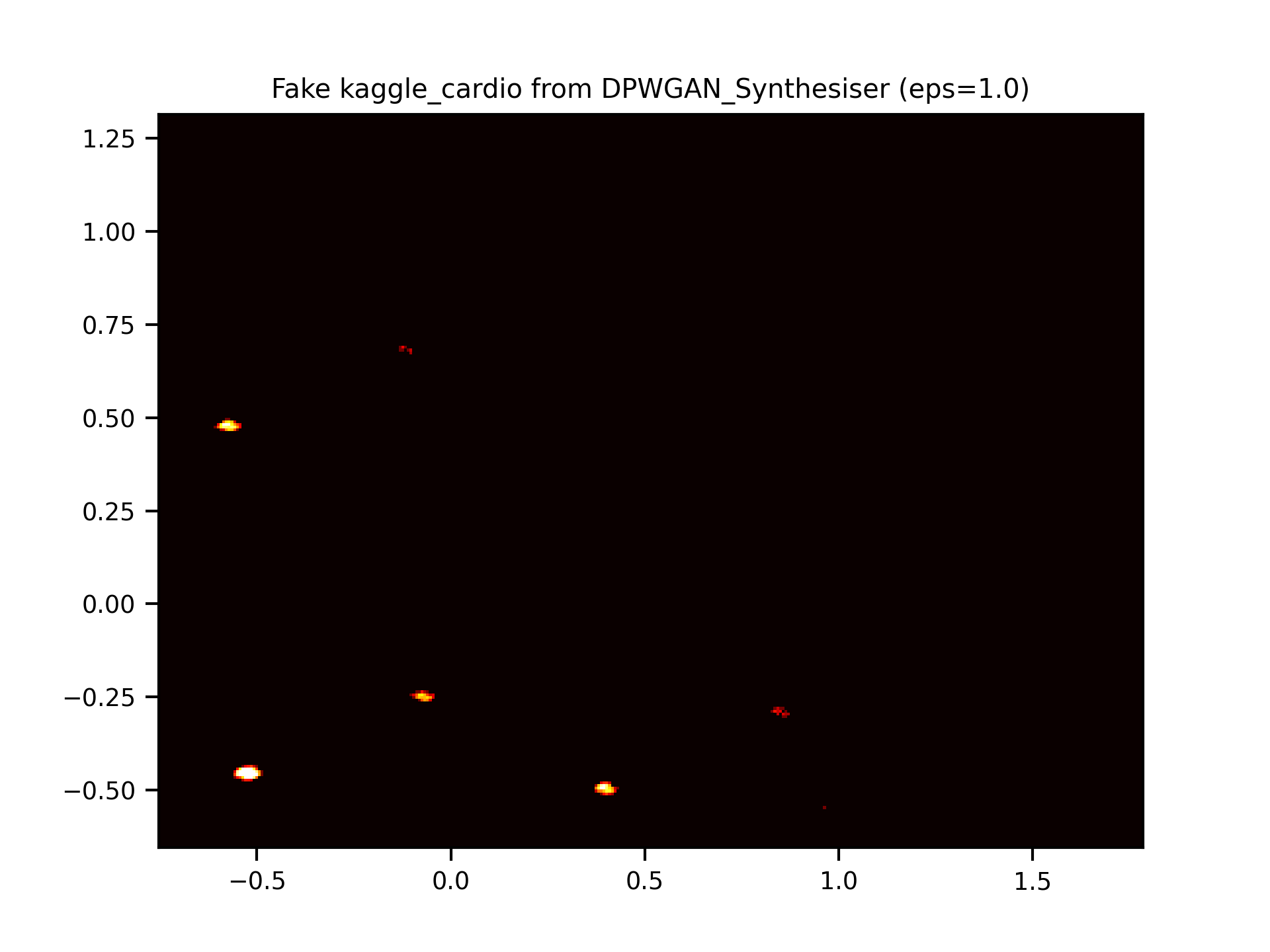}
\label{fig:kaggle_cardio_pca_DPWGAN_Synthesiser_1.0_1}}

\vspace{-1ex}

\subfloat{\includegraphics[trim=0.3\linewidth{} 0.25\linewidth{} 0.25\linewidth{} 0.12\linewidth{},clip,width=0.6\linewidth]{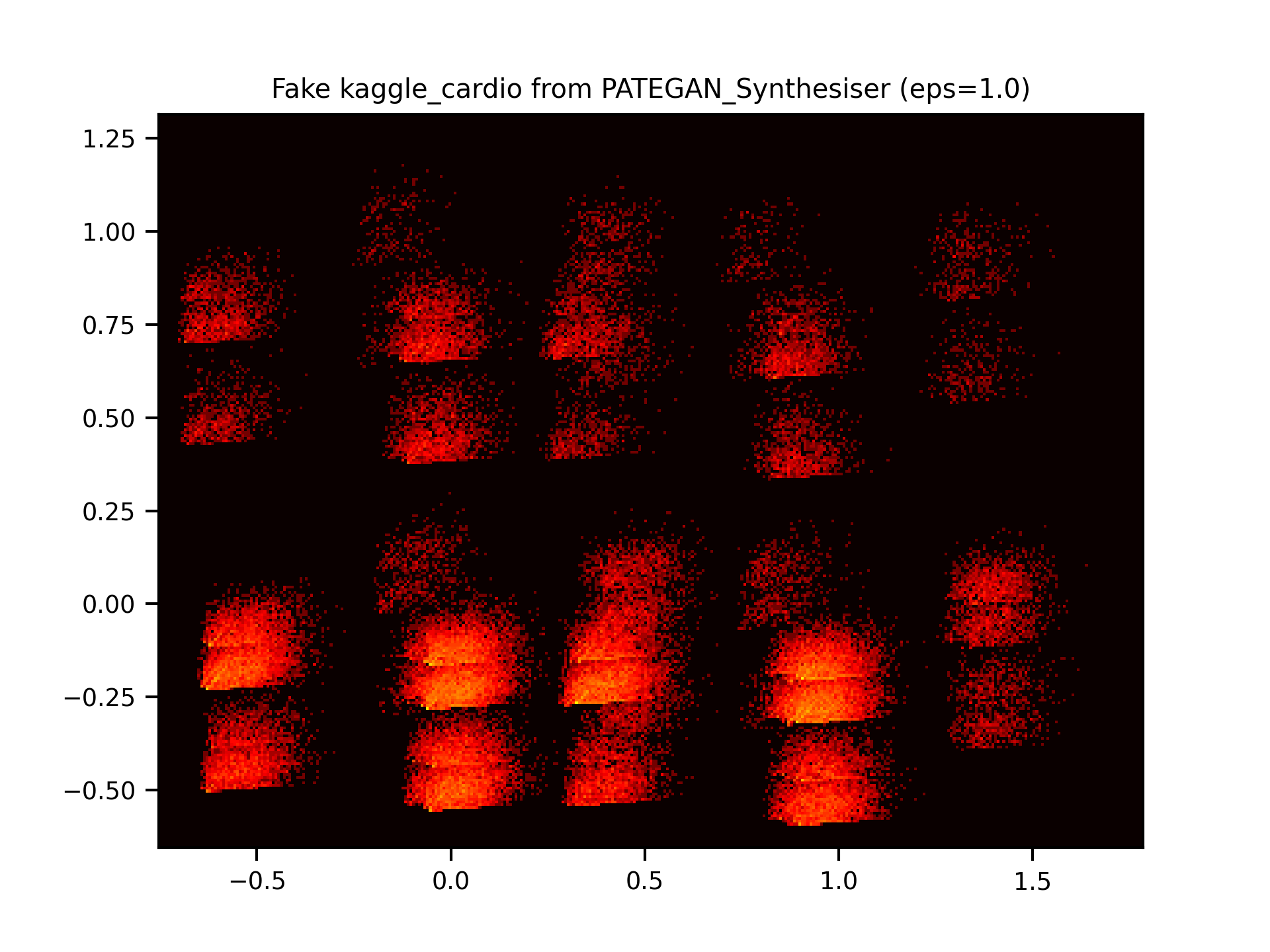}
\label{fig:kaggle_cardio_pca_PATEGAN_Synthesiser_1.0_1}}
\end{minipage}
\hfill
\begin{minipage}[t]{.40\linewidth}
\raggedright 
\subfloat{\includegraphics[trim=0.3\linewidth{} 0.25\linewidth{} 0.25\linewidth{} 0.12\linewidth{},clip,width=0.6\linewidth]{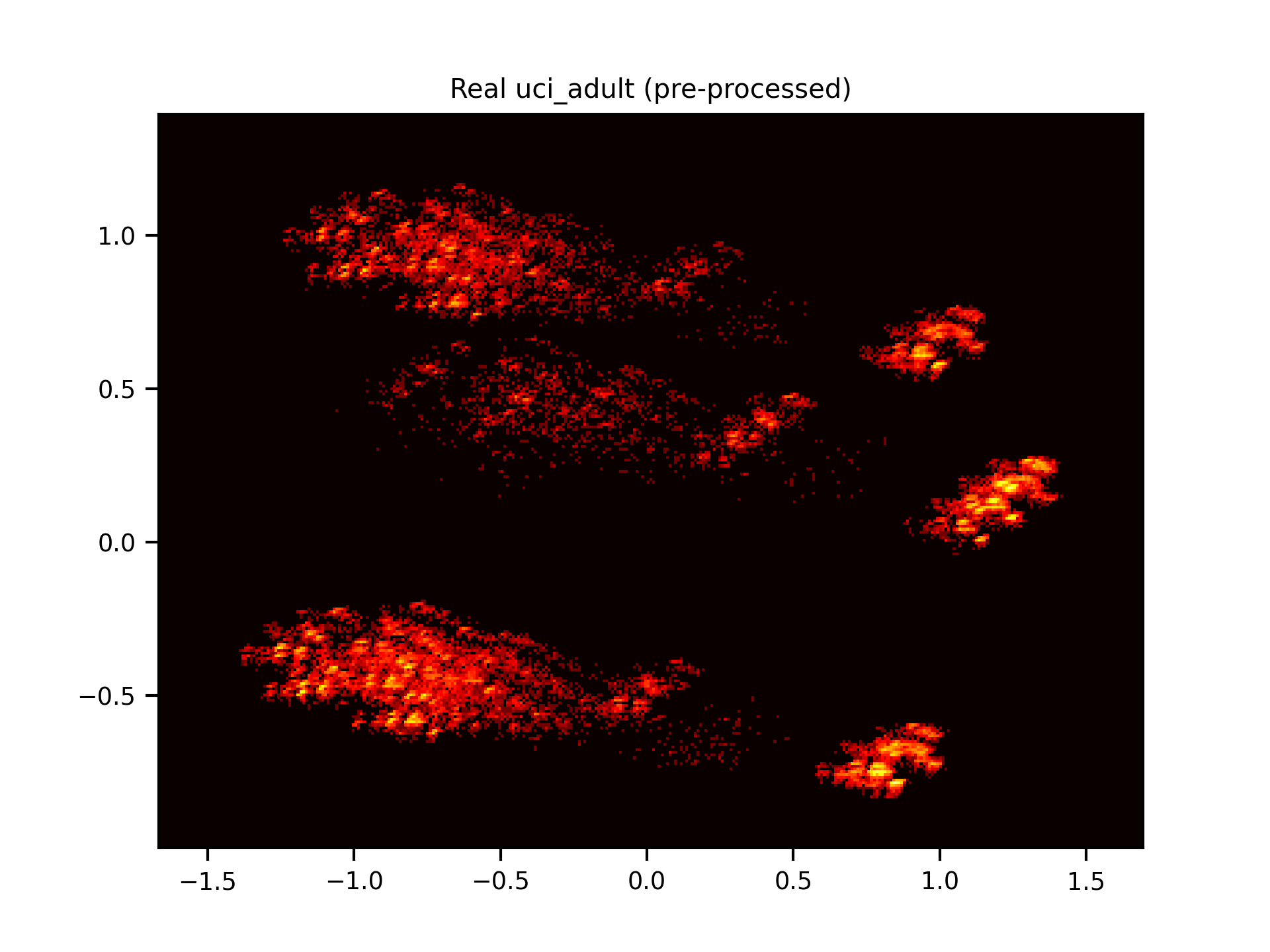}
\label{fig:uci_adult_pca_real}}

\vspace{-1ex}

\subfloat{\includegraphics[trim=0.3\linewidth{} 0.25\linewidth{} 0.25\linewidth{} 0.12\linewidth{},clip,width=0.6\linewidth]{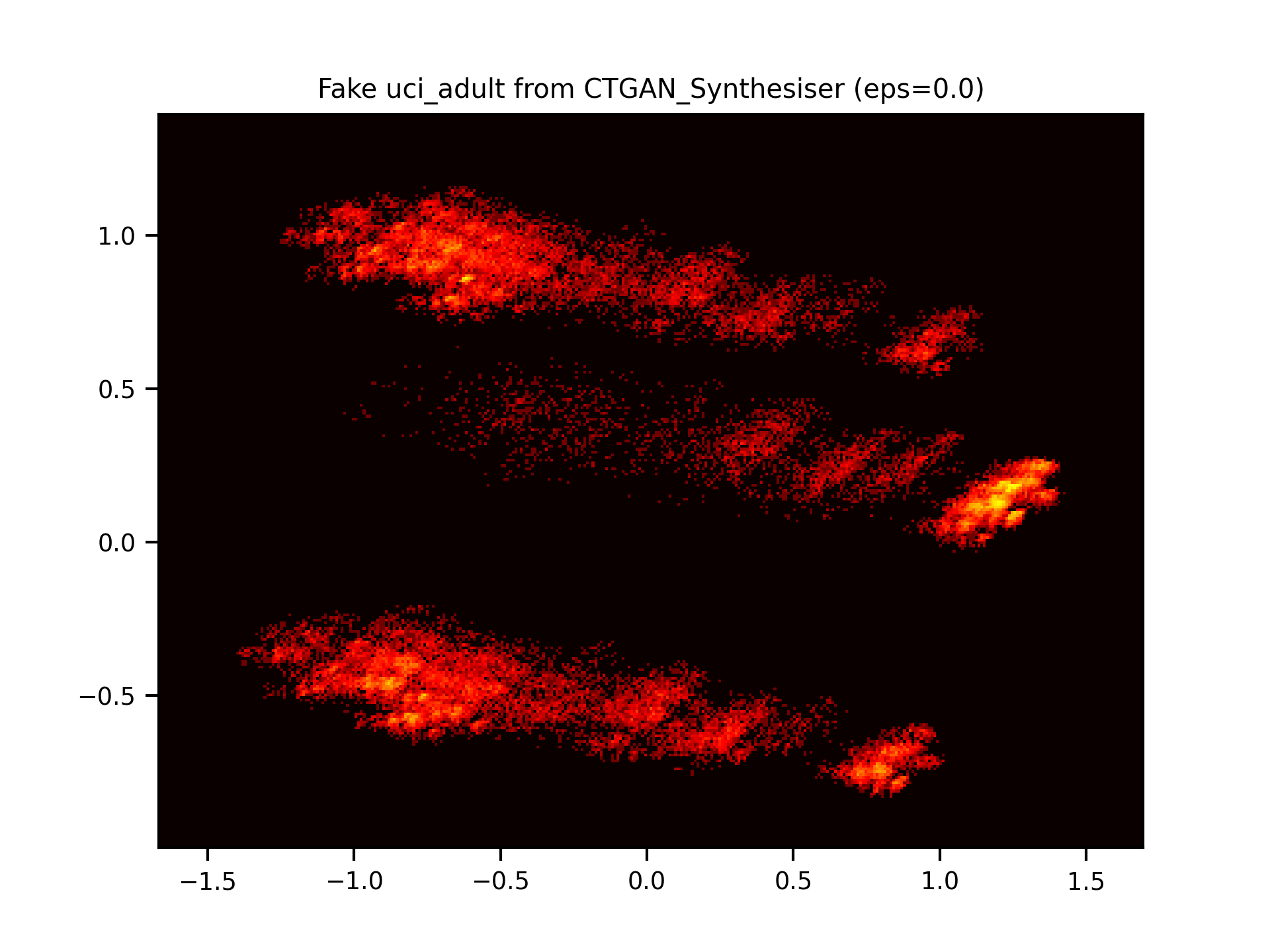}
\label{fig:uci_adult_pca_CTGAN_Synthesiser_0.0_1}}

\vspace{-1ex}

\subfloat{\includegraphics[trim=0.3\linewidth{} 0.25\linewidth{} 0.25\linewidth{} 0.12\linewidth{},clip,width=0.6\linewidth]{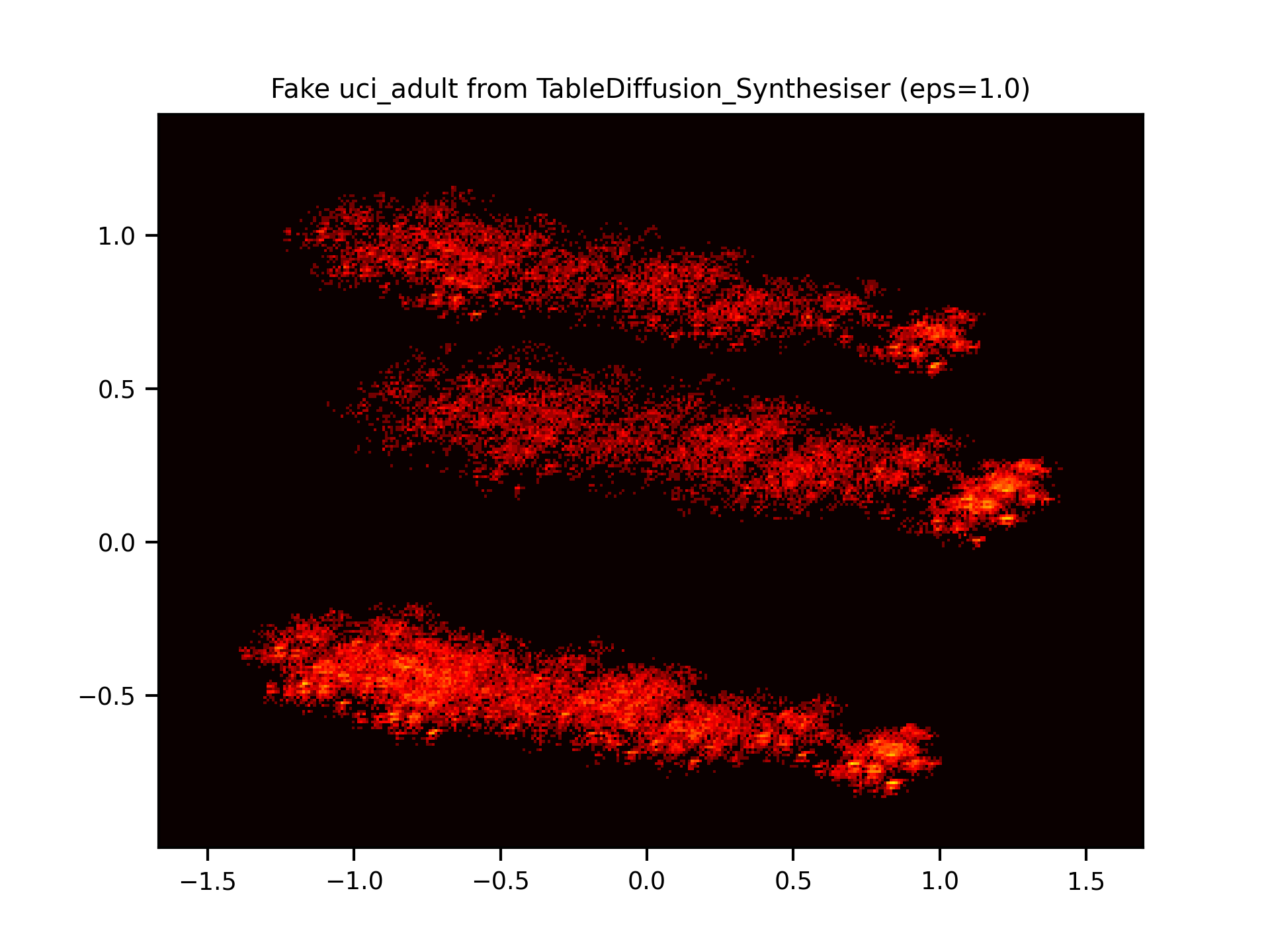}
\label{fig:uci_adult_pca_TableDiffusion_Synthesiser_1.0_1}}

\vspace{-1ex}

\subfloat{\includegraphics[trim=0.3\linewidth{} 0.25\linewidth{} 0.25\linewidth{} 0.12\linewidth{},clip,width=0.6\linewidth]{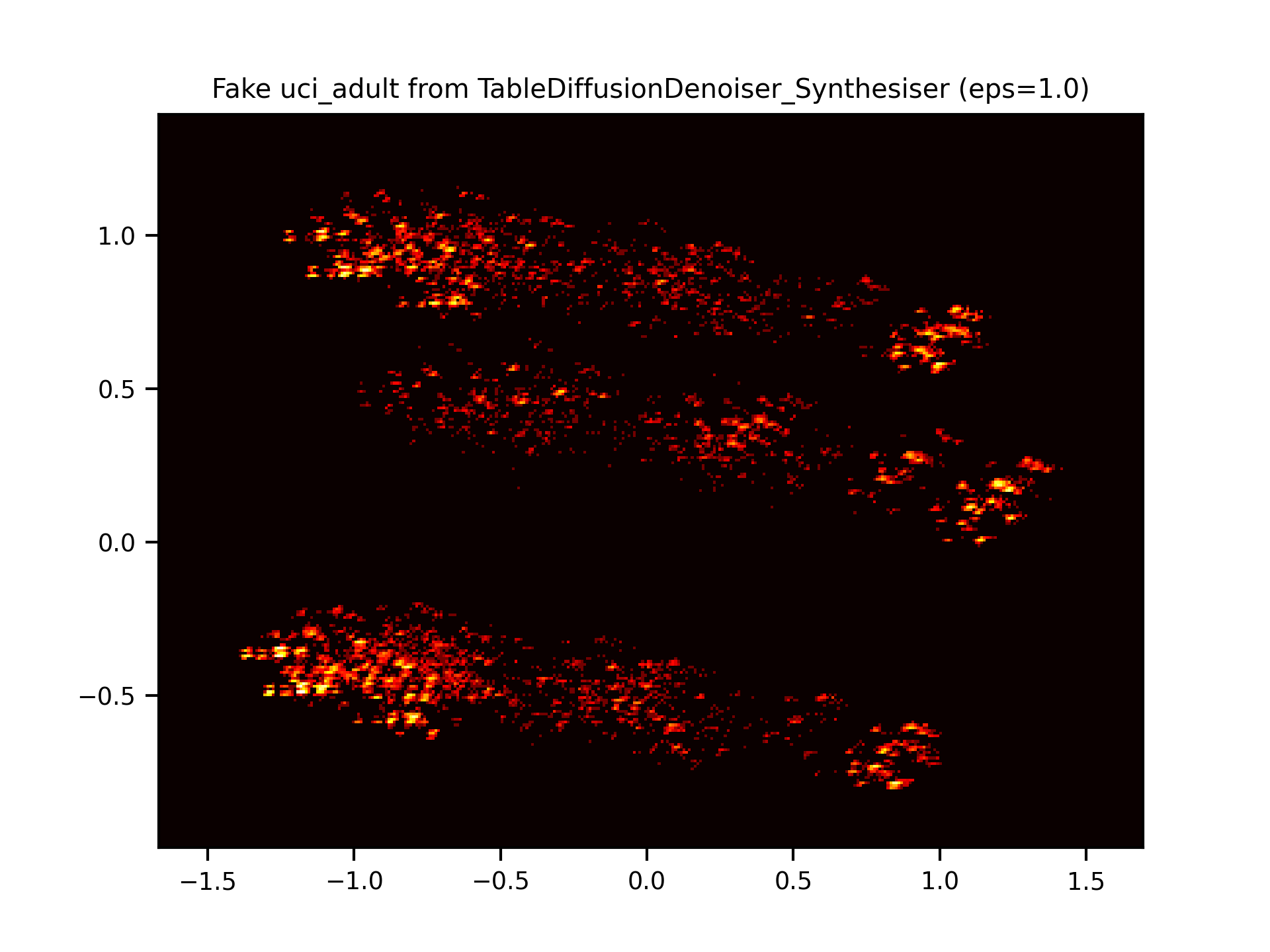}
\label{fig:uci_adult_pca_TableDiffusionDenoiser_Synthesiser_1.0_1}}

\vspace{-1ex}

\subfloat{\includegraphics[trim=0.3\linewidth{} 0.25\linewidth{} 0.25\linewidth{} 0.12\linewidth{},clip,width=0.6\linewidth]{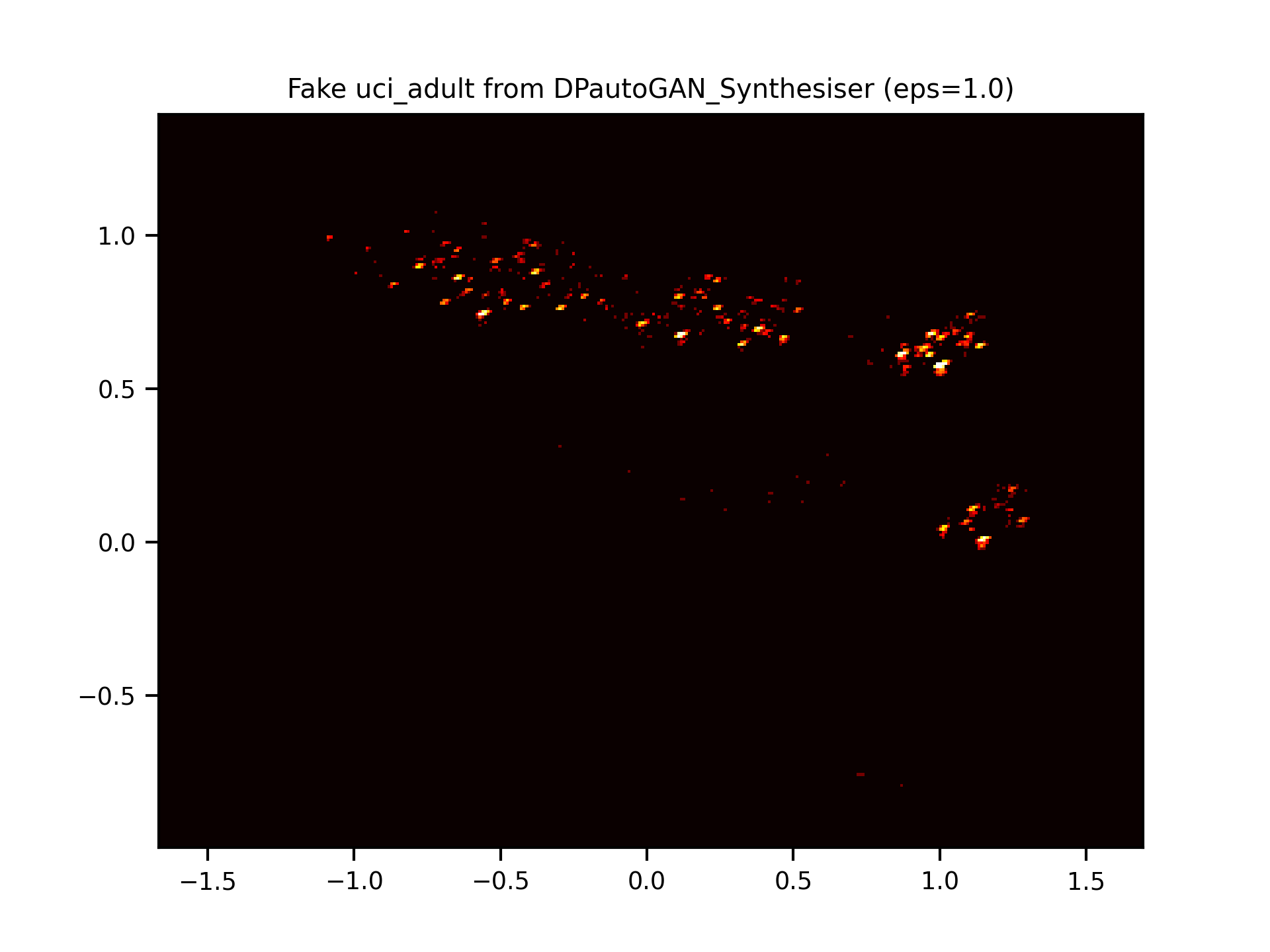}
\label{fig:uci_adult_pca_DPautoGAN_Synthesiser_1.0_1}}

\vspace{-1ex}

\subfloat{\includegraphics[trim=0.3\linewidth{} 0.25\linewidth{} 0.25\linewidth{} 0.12\linewidth{},clip,width=0.6\linewidth]{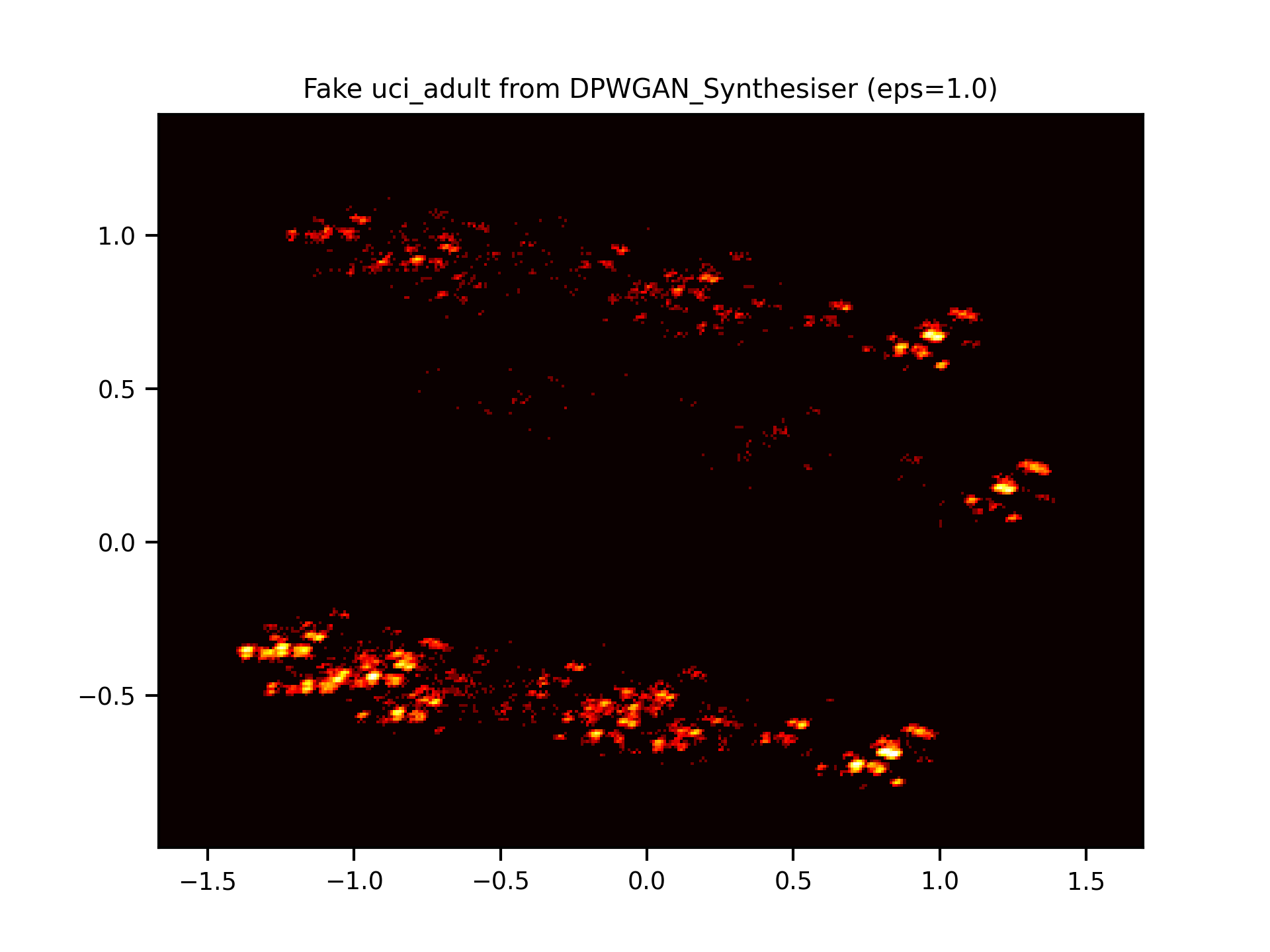}
\label{fig:uci_adult_pca_DPWGAN_Synthesiser_1.0_1}}

\vspace{-1ex}

\subfloat{\includegraphics[trim=0.3\linewidth{} 0.25\linewidth{} 0.25\linewidth{} 0.12\linewidth{},clip,width=0.6\linewidth]{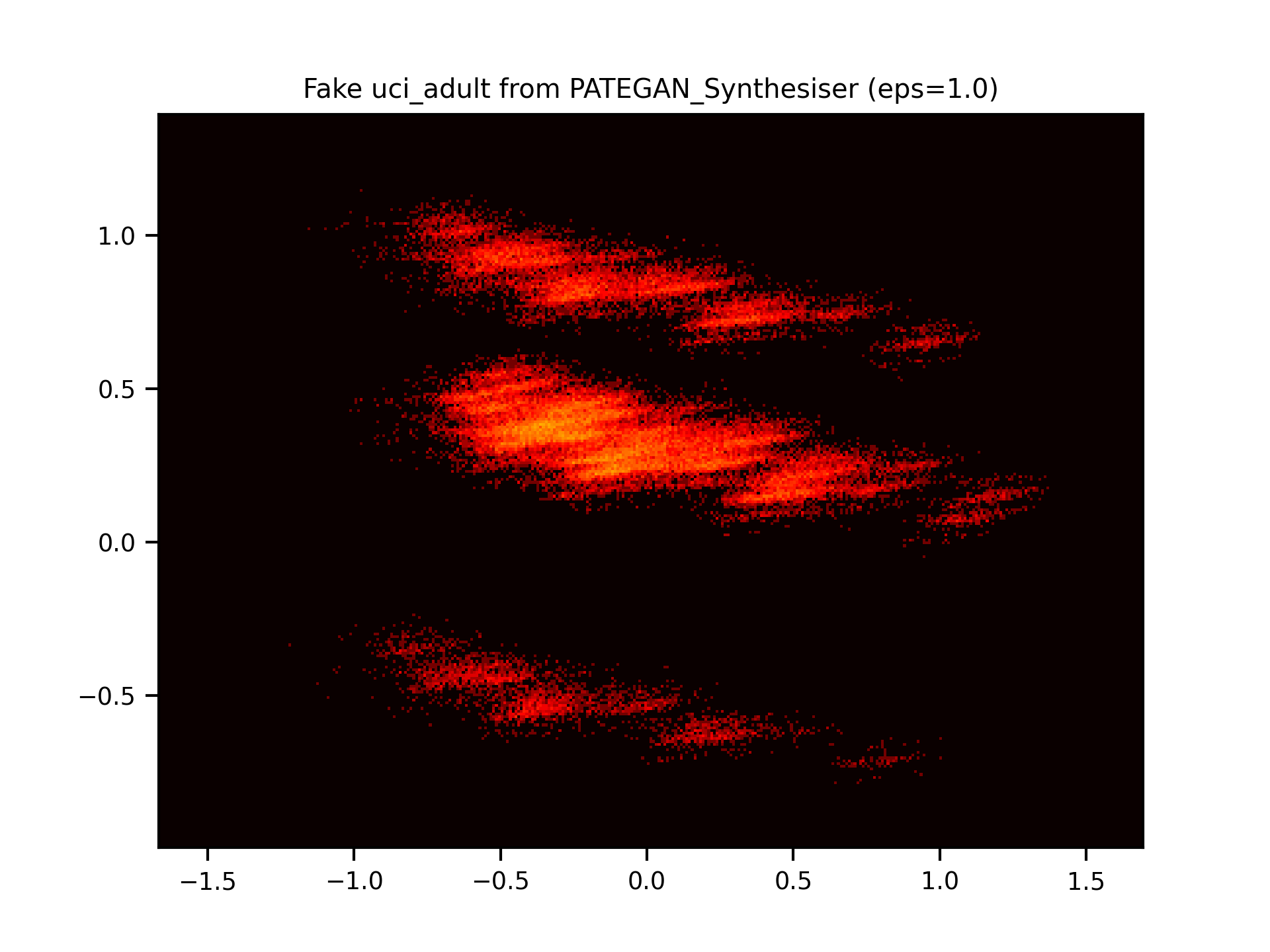}
\label{fig:uci_adult_pca_PATEGAN_Synthesiser_1.0_1}}
\end{minipage}

\caption{Projection heatmaps of real (top row) and synthetic datasets for Kaggle Cardio (left column) and UCI Adult (right column) datasets. The second row shows an unprivatised CTGAN benchmark for reference. Rows 3 through 7 are from the GAN and Diffusion models, trained under $(1, 10^{-5})$-DP. Darker regions have a lower density of points, whilst lighter regions indicate higher density. The plot axes are the first and second principal components of the transformer and normalised real datasets. The same projection eigenvalues and eigenvectors are shared across all subfigures in each column, allowing direct comparisons.}
\label{fig:projection_heatmaps}
\end{figure}

To get a visual intuition of multimodal distributions, we preprocessed the real and synthetic datasets, then rendered a heatmap of samples projected on the axes of the first two principal components of the real datasets (Figure \ref{fig:projection_heatmaps}). We see that the Kaggle Cardio dataset had much more distinct mode clusters than the UCI Adult dataset. This corresponds with the higher (and thus worse) pMSE ratio and marginal distance scores produced by all models on the Kaggle Cardio dataset (Table \ref{tab:metrics_kaggle_cardio_1.0}). When visualised as heatmaps, we see that training generative models under differential privacy exacerbates the blurring effect. At strict privacy levels, the denoising variant of the diffusion models produces much more discrete and scattered representations of the original datasets than the noise-predicting variant, highlighting a propensity for mode collapse. We can also see that the noise-predicting diffusion model produces excellent representations of the original data -- on par with, or exceeding those of the CTGAN baseline, even at strict privacy budgets. The GAN-based models suffer from extreme mode collapse, showing fewer clusters with much higher density, often missing out entire superclusters altogether. Interestingly, PATE-GAN seems to suffer less extreme mode collapse, but much more of the blurring effect around each cluster. This makes sense, given its teacher-student paradigm, and corresponds with its low $\alpha$-precision and $\beta$-recall scores.

\section{Discussion}

The superior performance of our diffusion-based models over GAN-based models across privacy levels and metrics is the most notable result of our work. Not only were the diffusers better at learning the joint and marginal distributions of the original data, but they were also more consistent across runs, showing much lower variance and more stable training loss curves. This suggests that the diffusion approach is vastly more stable and effective than the adversarial approach employed by GANs. This concurs with other findings in the diffusion literature and the recent explosion in generative performance since diffusion models have become widespread. Where GANs rely on training two neural networks in tandem (using two optimisers in the privacy-preserving context) and involve a delicate balancing act to converge, the diffusion approach trains only one network with one optimiser. Additionally, the iterative approach of varying noise levels in diffusion training makes for a more manageable and forgiving learning scenario than the all-or-nothing challenge posed to the Generator module of a GAN.  

Moreover, in the context of differentially-private gradient descent, the diffusion paradigm has a number of advantages over the adversarial one. The gradient clipping and noise addition appear to destabilise GANs easily, but act almost like a regularisation constraint for the diffusion models. DP-SGD, in its essence, uses noise to obfuscate sensitive information. Diffusers are models designed to model added noise as their information bottleneck, making the additional (gradient) noise of DP-SGD a less aggressive perturbation to their training. This also appears to make varying the privacy budget (and thus the added noise) less disruptive to the tuning of diffusion models than to delicately-balanced GANs, making off-the-shelf diffusers a much more reliable starting point for real-world applications.

In almost all cases, the noise-predicting diffusion model outperformed the denoising variant in terms of both raw performance and consistency across runs. This was likely due to the mixed-type tabular modality, which neural networks find very challenging. Whilst the denoiser attempted to reconstruct the original data (in the preprocessed space), the noise predictor simply had to predict multivariate Gaussian noise -- a vastly easier modality for neural networks. This may also explain the lower occurrence of invented modes in the noise predictor, as generating a Gaussian distribution is far less likely to result in numerical instability within the neural network than generating transformed tabular distributions. Additionally, the loss function of the denoiser was the sum of the MSE loss on the continuous features and the normalised KL-divergence loss on the categorical features, whilst the noise predictor only had an MSE loss term. Balancing the continuous and categorical loss terms would involve tuning additional weight parameters and is specific to each dataset. Particularly in the case of high-cardinality categoricals, like those in the UCI Adult dataset, the categorical loss can potentially overwhelm the continuous loss. 

This difference between the two variants of the diffusion model highlights one of the key strengths of the diffusion paradigm. By iteratively predicting the Gaussian noise added to the data instead of directly denoising and reconstructing the data, we are able to leverage the strengths of the neural network in ways that GANs and VAEs cannot, leaving the task of constructing the synthetic data up to simple vector arithmetic. Whilst the denoising variant still outperformed the GAN-based models in most cases -- a testament to the data efficiency and stability of the diffusion paradigm -- our ablation analysis highlights the numerous advantages and versatility of diffusion models.

Crucially, the diffusion approach appears to avoid the biggest issue that plagues GANs across data modalities, mode collapse. While many techniques have been developed to mitigate the issue in GANs, the off-the-shelf performance of the diffusion models still outperformed highly-engineered GAN variants. This is particularly useful for the kinds of high-cardinality categorical features often seen in tabular data.

\chapter{Limitations and future work}\label{ch:limits-and-future}

Whilst all efforts were made in this work to conduct rigorous and representative evaluations, there are many areas to explore in future work. Although the diffusion models outperformed the GAN approaches on all measures, they were simplified proof-of-concept models. There have been an overwhelming number of advances in diffusion techniques in recent research that could be evaluated in the sensitive tabular data setting. A single, fixed cosine noise schedule was used for all the diffusion implementations across all datasets. The evaluation of more sophisticated schedulers, and even DP-optimised schedulers, is an interesting area for future work. 

Our experiments focussed on evaluating synthetic data fidelity, which is more quantifiable and application-agnostic than utility, whilst being strongly correlated. Naturally, our analysis could be extended to also evaluate synthetic data utility on a set of downstream use-cases in specific fields. Moreover, it would be of immense value for future work to formally validate the privacy of our novel diffusion models with various privacy attacks, especially in preparation for real-world deployment.

To allow for direct comparisons, many experimental variables were fixed, such as the neural network architecture, privacy accountant, hyperparameter tuning protocol, and preprocessing techniques. Future work could naturally explore the effects of different configurations and optimisations. However, there were also some unavoidable differences across models which should be considered along with the results of this work: The CTGAN baseline and the diffusion models were exposed to the raw tabular data and internally manage their own pre- and post- processing of the features due to the ways in which they handle continuous versus categorical features, whereas the other models used a shared pre- and post- processing configuration that provides the models a single, continuous latent space. 

To constrain the scope and compute budget, this work fixed the neural network architecture and many of the hyperparameters. The variable hyperparameters were only tuned for general performance, not for specific privacy budgets and datasets. In practice, model performance could increase significantly if tuned for a specific dataset and privacy level.

Whilst two widely-available, challenging, and representative datasets were selected for this study, future work could expand these comparisons to a greater variety of datasets. In particular, the relative model performance under different dataset dimensionality, data type composition, and distributions would be a fascinating direction to explore. Two of the biggest opportunities for future work are the evaluation of different approaches to tabular diffusion models -- with different schedulers, loss functions, and sampling mechanisms -- and reconciling the end-to-end representation learning  approach with diffusion models.

\chapter{Conclusions}\label{ch:conclusions}
The goal of this study was to improve the fidelity-privacy trade-off of generative models, with the goal of synthesising higher-quality tabular datasets without compromising individual privacy. Our research questions (Section \ref{sec:problem-and-research-Qs}) explored:
\begin{enumerate}
    \item \textbf{Increased data efficiency}: Can we make better use of the sensitive training data and thus produce a higher-fidelity synthetic dataset with the same privacy budget?
    \begin{enumerate}
        \item \textbf{Improved tabular representation}: Can we improve the way mixed-type tabular data is represented to the models and thus learn with higher-fidelity under the same privacy budget?
        \item \textbf{Sample augmentation}: Can we augment the sensitive data samples such that the models can learn more at each training step under the same privacy budget?
    \end{enumerate}
    \item \textbf{Increased training efficiency}: Can we implement generative models that produce high-fidelity data in fewer training steps and thus use less privacy budget?
\end{enumerate}

We implemented the first differentially-private diffusion models for tabular data and comprehensively benchmarked them against the state-of-the-art GAN-based models. Our findings demonstrate that diffusion models significantly outperformed GAN-based models across different datasets and privacy levels in terms of fidelity, diversity, and stability. 

By implementing two variants of the tabular diffusion model, we could perform ablation to understand which mechanisms resulted in improved performance relative to the GANs. The noise-predicting diffusion model was extremely successful at replicating the original data's marginal and joint distributions whilst operating under strict privacy constraints. Moreover, the diffusion models avoided the well-documented issues of mode collapse and oversampling of secondary modes, which are common drawbacks of GAN-based models. This suggests that latent diffusion models hold greater promise for practical applications where synthetic data generation must be both high-fidelity and compliant with data privacy regulations. Another key finding is the stable performance of diffusion models in contrast to the variability of GAN-based models. The stability of the diffusion models can be attributed to their training process, which is not only less complex compared to the adversarial nature of GANs but also seems to be inherently more compatible with differentially-private training regimes. 

One of the biggest challenges is learning robust representations of the tabular format that are conducive to neural networks and other ML architectures. We implemented end-to-end tabular data generators (using self-attention and inter-sample attention mechanisms) to learn their own representations of tabular data (RQ1a). Unfortunately, the additional parameters and training steps involved drained the privacy budget quickly and destabilised the inner generative models. However, the use of noise-predicting diffusion models enabled us to bypass the representation challenges of mixed-type tabular data, as outputting Gaussian noise is far more achievable with neural networks.

Overall, the diffusion paradigm is far more data-efficient (RQ1) and training-efficient (RQ2) than the adversarial paradigm. By re-using the same training data multiple times in each batch, augmented by varying degrees of noise, the diffusion models were able to leverage higher data efficiency to produced higher synthetic data fidelity (RQ1b). The smoother gradient updates and increased resilience to sparsity reduced the rate at which the privacy budget was consumed in the diffusion models, making the models more training efficient (RQ2).

This work contributes to the ongoing literature on synthetic data generation under differential privacy. The superiority of noise-predicting diffusion models in terms of performance and stability offers a new avenue for researchers and practitioners who are seeking reliable and privacy-compliant ways of generating synthetic data. Future research should focus on incorporating advanced diffusion techniques, exploring more sophisticated noise schedules, and expanding the analysis to a wider variety of datasets and architectures.

In conclusion, diffusion models are a powerful new tool for generating high-quality synthetic tabular data, even when adhering to differential privacy constraints. It is our hope that these contributions unlock new avenues for research and enable real-world progress on sensitive tabular data.

\bibliography{bibliography}

\appendix

\chapter{Results of end-to-end attention-based models}\label{ch:appendix-attn} 

\begin{table}[htbp]
\centering
\caption{Mean performance (over 10 runs) on Kaggle Cardio under (1.0,$10^{-5}$)-DP.}
\label{tab:attn_metrics_kaggle_cardio}
\begin{tabular}{lllllll}
\hline
                 Model & $\epsilon$ &       pMSE &             MD & $\alpha$-Precision & $\beta$-Recall &          AUPRC \\
\hline
                DP-WGAN &          1 &      15164 &          0.591 &              0.639 &          0.001 &          0.001 \\
             DP-auto-GAN &          1 &      17162 &          0.401 &              0.716 &          0.001 &          0.001 \\
               PATE-GAN &          1 &      15448 &          0.749 &              0.021 &          0.001 &          0.000 \\
        TableDiffusion &          1 &       2461 &          0.467 &    0.881 & 0.286 & 0.253 \\
TableDiffusionDenoiser &          1 &       9543 & 0.331 &              0.599 &          0.032 &          0.019 \\
\hline
           \textbf{AttentionAE} &          1 &      13419 &          0.456 &              0.150 &          0.001 &          0.000 \\
          \textbf{AttentionGAN} &          -- & 12162 &          0.683 &              0.001 &          0.001 &          0.000 \\
          \textbf{AttentionVAE} &          -- &      17493 &          0.653 &              0.067 &          0.001 &          0.000 \\
                 
\hline
		CTGAN (benchmark) &         -- &       1518 &          0.421 &              0.778 &          0.291 &          0.227 \\
\hline
\end{tabular}
\end{table}

\begin{table}[htbp]
\centering
\caption{Mean performance (over 10 runs) on UCI Adult under (1.0,$10^{-5}$)-DP.}
\label{tab:attn_metrics_uci_adult}
\begin{tabular}{lllllll}
\hline
                 Model & $\epsilon$ &         pMSE &             MD & $\alpha$-Precision & $\beta$-Recall &          AUPRC \\
\hline
                DP-WGAN &          1 &         1689 &          0.355 &              0.513 &          0.009 &          0.005 \\
             DP-auto-GAN &          1 &         1991 &          0.291 &              0.742 &          0.001 &          0.001 \\
               PATE-GAN &          1 &         2031 &          0.692 &              0.007 &          0.001 &          0.000 \\
        TableDiffusion &          1 & 590 &          0.122 &              0.667 & 0.170 &          0.115 \\
TableDiffusionDenoiser &          1 &         1088 & 0.089 &     0.833 &          0.161 & 0.134 \\
\hline
           \textbf{AttentionAE} &          1 &         1716 &          0.318 &              0.748 &          0.001 &          0.001 \\
          \textbf{AttentionGAN} &          -- &          829 &          0.671 &              0.001 &          0.001 &          0.000 \\
          \textbf{AttentionVAE} &          -- &         2078 &          0.777 &              0.001 &          0.001 &          0.000 \\
                 
\hline
		CTGAN (benchmark) &         -- &          353 &          0.242 &              0.733 &          0.143 &          0.106 \\
\hline
\end{tabular}
\end{table}

\begin{figure}
    \centering
    \includegraphics[width=0.95\textwidth]{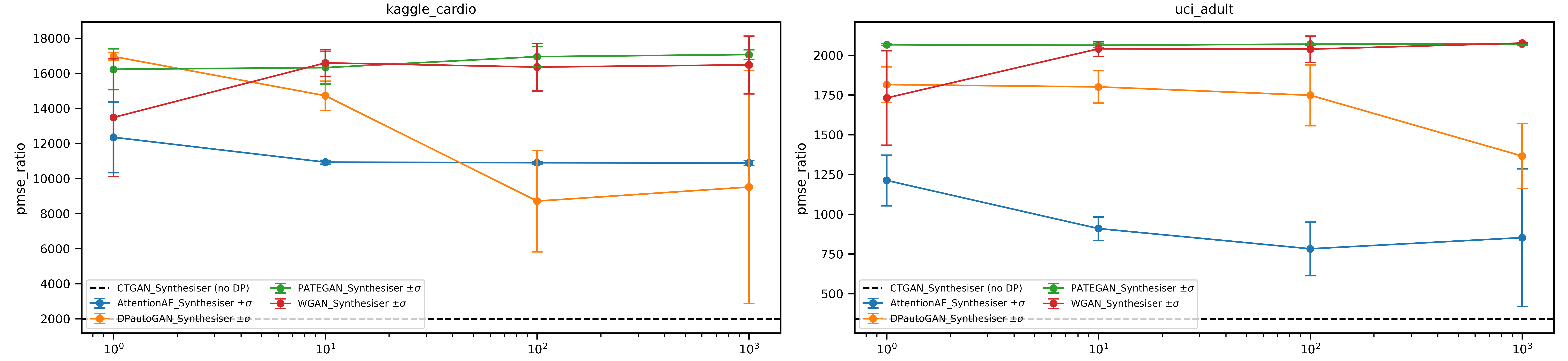}
    \caption{Comparison of pMSE ratio (lower is better) of AttentionAE reconstructions against other models across datasets and privacy budgets ($\epsilon$). Points show the mean performance over repeated runs and error bars show $\pm 1\sigma$ of standard deviation. Dashed line shows mean CTGAN baseline score over 10 runs with no privatisation.}
    \label{fig:exp_220221_121750_results_pmse}
\end{figure}

\begin{figure}
    \centering
    \includegraphics[width=0.95\textwidth]{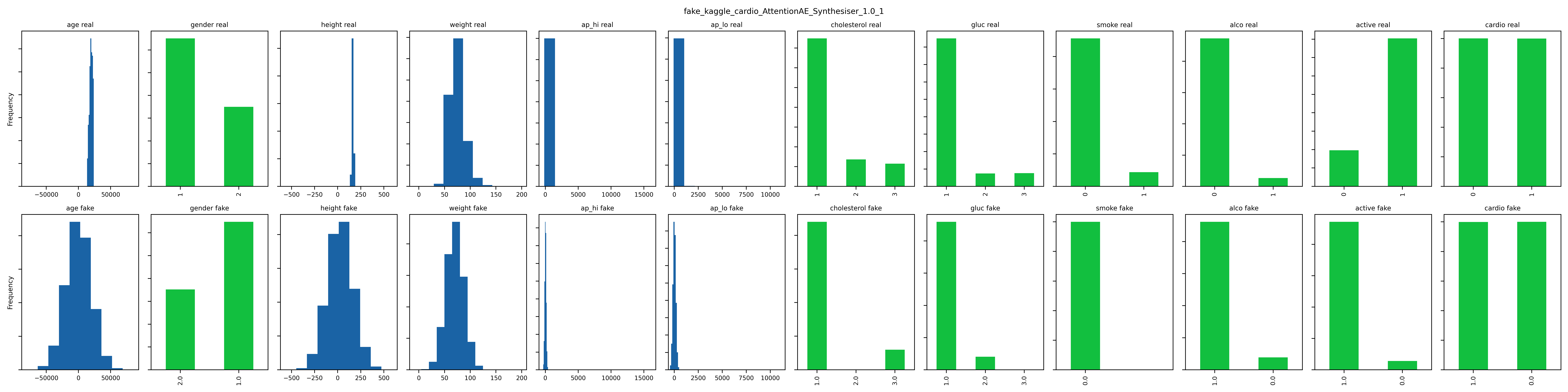}
    \caption{Comparison of real (top row) and reconstructed (bottom row) marginal distributions of AttentionAE on the Kaggle Cardio dataset. Categorical features are coloured green and continuous features are coloured blue. For easy comparison, all subfigures are histograms sharing the same vertical axes ($y \in [0,1]$). Real and synthetic features in the same column share the same horizontal axes.}
    \label{fig:kaggle_cardio_marginals_AttentionAE_Synthesiser_1.0_1}
\end{figure}
\begin{figure}
    \centering
    \includegraphics[width=0.95\textwidth]{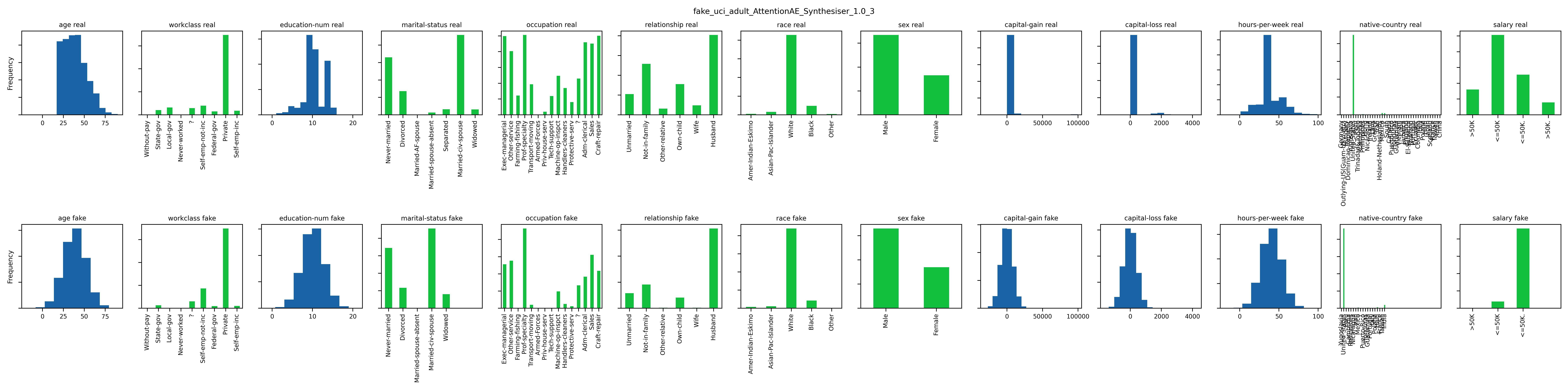}
    \caption{Comparison of real (top row) and reconstructed (bottom row) marginal distributions of AttentionAE on the UCI Adult dataset. Categorical features are coloured green and continuous features are coloured blue. For easy comparison, all subfigures are histograms sharing the same vertical axes ($y \in [0,1]$). Real and synthetic features in the same column share the same horizontal axes.}
    \label{fig:uci_adult_marginals_AttentionAE_Synthesiser_1.0_3}
\end{figure}

\begin{figure}[ht]
\centering

\begin{minipage}[t]{.45\linewidth}
\raggedleft 
\subfloat{\includegraphics[trim=0.3\linewidth{} 0.25\linewidth{} 0.25\linewidth{} 0.12\linewidth{},clip,width=0.6\linewidth]{figs/kaggle_cardio_pca_real.png}
\label{fig:kaggle_cardio_pca_real_appendix}}

\vspace{-1ex}

\subfloat{\includegraphics[trim=0.3\linewidth{} 0.25\linewidth{} 0.25\linewidth{} 0.12\linewidth{},clip,width=0.6\linewidth]{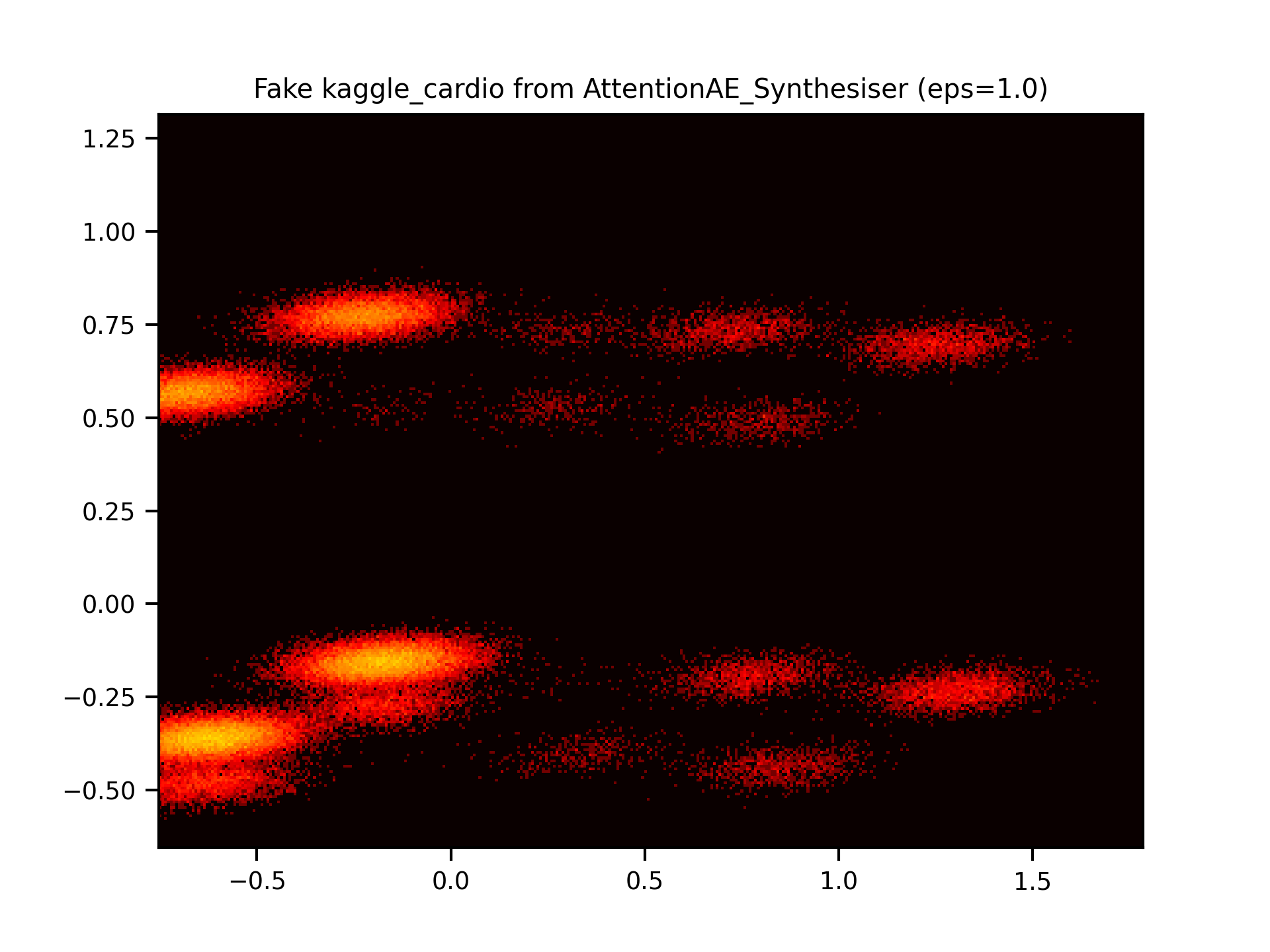}
\label{fig:kaggle_cardio_pca_AttentionAE_Synthesiser_1.0_1}}

\end{minipage}
\hfill
\begin{minipage}[t]{.45\linewidth}
\raggedright 
\subfloat{\includegraphics[trim=0.3\linewidth{} 0.25\linewidth{} 0.25\linewidth{} 0.12\linewidth{},clip,width=0.6\linewidth]{figs/uci_adult_pca_real.png}
\label{fig:uci_adult_pca_real_appendix}}

\vspace{-1ex}

\subfloat{\includegraphics[trim=0.3\linewidth{} 0.25\linewidth{} 0.25\linewidth{} 0.12\linewidth{},clip,width=0.6\linewidth]{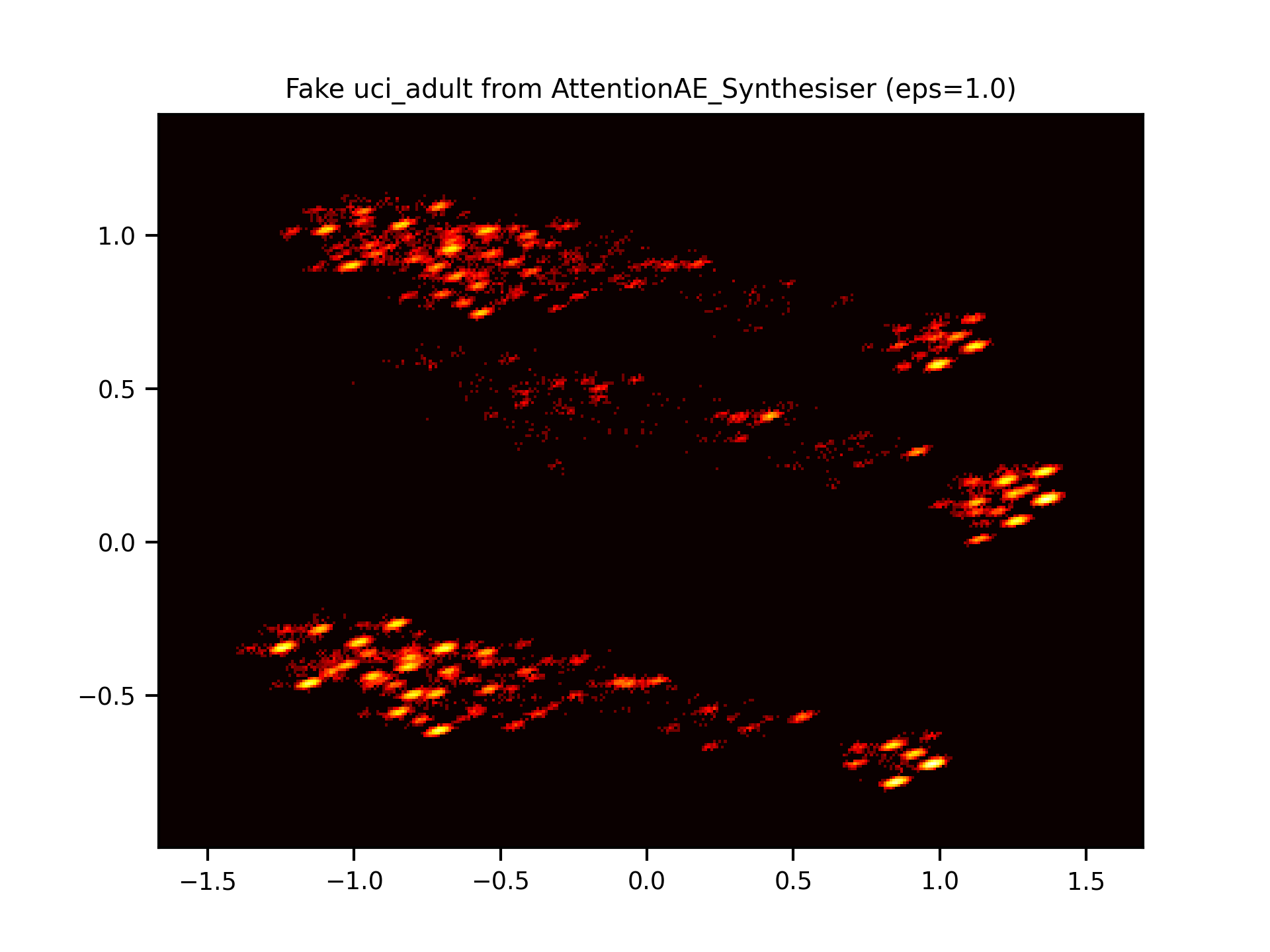}
\label{fig:uci_adult_pca_AttentionAE_Synthesiser_1.0_5}}

\end{minipage}

\caption{Projection heatmaps of real (top row) and AttentionAE-reconstructed datasets for Kaggle Cardio (left column) and UCI Adult (right column) datasets. The second row shows the reconstructed dataset from AttentionAE, trained under $(1, 10^{-5})$-DP. Darker regions have a lower density of points, whilst lighter regions indicate higher density. The plot axes are the first and second principal components of the transformer and normalised real datasets. The same projection eigenvalues and eigenvectors are shared across all subfigures in each column, allowing direct comparisons.}
\label{fig:projection_heatmaps_attentionAE}
\end{figure}

\end{document}